\documentclass{article}
\usepackage{iclr2026_conference,times}

\usepackage{amsmath,amsfonts,bm}

\def\eqref#1{equation~\ref{#1}}

\def\1{\bm{1}}

\DeclareMathAlphabet{\mathsfit}{\encodingdefault}{\sfdefault}{m}{sl}
\SetMathAlphabet{\mathsfit}{bold}{\encodingdefault}{\sfdefault}{bx}{n}

\usepackage{multirow}
\usepackage[dvipsnames]{xcolor}
\usepackage{booktabs}
\usepackage{array}
\usepackage{amssymb}
\usepackage{color}
\usepackage{colortbl}
\usepackage{subcaption}
\usepackage{caption}
\usepackage{graphicx}
\usepackage{mdframed}
\usepackage{pifont}
\usepackage{tikz}
\usepackage{comment}
\usepackage{pgfplots}
\usepackage{comment}
\pgfplotsset{compat=1.16}

\newcommand{\numberofsystems}{eleven}

\usepackage[frozencache,cachedir=_minted-main]{minted}
\fvset{fontsize=\scriptsize,xleftmargin=10pt,numbers=left,numbersep=5pt}

\newcommand{\code}[1]{\texttt{#1}}
\newmdenv[
  backgroundcolor=gray!5,
  linecolor=black,
  roundcorner=6pt,
  skipabove=4pt,
  skipbelow=0pt,
  innertopmargin=8pt,
  innerbottommargin=8pt,
  innerleftmargin=5pt,
  innerrightmargin=5pt,
  font=\ttfamily\scriptsize,
]{promptbox}

\newenvironment{promptboxwithheader}[1]{%
  \begin{mdframed}[
    backgroundcolor=gray!5,
    linecolor=black,
    roundcorner=6pt,
    skipabove=8pt,
    skipbelow=0pt,
    innertopmargin=0pt,
    innerbottommargin=4pt,
    innerleftmargin=5pt,
    innerrightmargin=5pt,
    frametitlefont=\bfseries\small\color{white},
    frametitlebackgroundcolor=black!80,
    frametitle={#1},
    frametitleaboveskip=3pt,
    frametitlebelowskip=3pt,
    frametitlerule=true,
    frametitlerulewidth=0pt,
    font=\ttfamily\scriptsize,
  ]%
  \vspace{4pt}%
  \selectfont
}{
  \vspace{2pt}%
  \small
  \end{mdframed}%
}

\newenvironment{longlisting}{\captionsetup{type=figure}}{}

\definecolor{deepskyblue}{rgb}{0.0, 0.75, 1.0}

\newenvironment{plsreview}
  {\begingroup}
  {\endgroup}
\newcommand{\plsreviewinline}[1]{{#1}}
\newcommand{\plsreviewtable}{}

\newcommand{\bench}{\textsc{SysMoBench}}
\newcommand{\tla}{TLA$^+$}

\newcommand{\bla}{\color{black}}
\newcommand{\gra}{\color{gray}}

\definecolor{darkgreen}{rgb}{0.0,0.5,0.0}
\newcommand{\promptgreen}{\color{darkgreen}}

\definecolor{Gray2}{gray}{0.75}
\definecolor{LightGray2}{gray}{0.85}
\definecolor{VeryLightGray2}{gray}{0.95}
\definecolor{blue-violet}{rgb}{0.54, 0.17,0.89}

\definecolor{bittersweet}{rgb}{1.0, 0.44,0.37}

\definecolor{chocolate}{rgb}{0.82, 0.41,0.12}

\newcommand{\checkmarksymbol}{\textcolor{ForestGreen}{\ \ding{51}}}
\newcommand{\wrongsymbol}{\textcolor{red}{\ \ding{55}}}

\newcommand{\excellentcell}{\cellcolor{ForestGreen!7}}
\newcommand{\goodcell}{\cellcolor{ForestGreen!4}}
\newcommand{\faircell}{\cellcolor{ForestGreen!1}}

\newcommand{\para}[1]{\smallskip\noindent {\bf #1} }

\newenvironment{packed_itemize}{
\begin{list}{\labelitemi}{\leftmargin=1.0em}
 \setlength{\itemsep}{2.5pt}
 \setlength{\parskip}{2pt}
 \setlength{\parsep}{0pt}
 \setlength{\headsep}{0pt}
 \setlength{\topskip}{0pt}
 \setlength{\topmargin}{0pt}
 \setlength{\topsep}{0pt}
 \setlength{\partopsep}{0pt}
}{\end{list}}

\newenvironment{packed_enumerate}{
\begin{enumerate}
 \setlength{\itemsep}{1pt}
 \setlength{\parskip}{0pt}
 \setlength{\parsep}{0pt}
 \setlength{\topsep}{0pt}
 \setlength{\partopsep}{0pt}
}{\end{enumerate}}

\definecolor{midnightblue}{rgb}{0.1, 0.1, 0.44}
\newcommand{\systemlink}[2]{\href{#1}{\textcolor{midnightblue}{#2}}}
\newcommand{\buglink}[2]{\href{#1}{\textcolor{midnightblue}{#2}}}

\usepackage{hyperref}
\usepackage{url}
\usepackage{graphicx}

\title{\bench{}: Evaluating AI on Formally\\ Modeling Complex Real-World Systems}

\author{Qian Cheng${}^{\spadesuit}$
  \quad Ruize Tang${}^{\heartsuit}$
  \quad Emilie Ma${}^{\clubsuit}$
  \quad Finn Hackett${}^{\clubsuit}$
  \quad Peiyang He${}^{\spadesuit}$
  \\
  {\bf Yiming Su}${}^{\diamondsuit}$
  \quad {\bf Ivan Beschastnikh}${}^{\clubsuit}$
  \quad {\bf Yu Huang}${}^{\spadesuit}$
  \quad {\bf Xiaoxing Ma}${}^{\spadesuit}$
  \quad {\bf Tianyin Xu}${}^{\diamondsuit}$
  \\
  ${}^{\spadesuit} $Nanjing University, \{cq, peiyang\_he\}@smail.nju.edu.cn, {\{xxm, yuhuang\}@nju.edu.cn}\\
  ${}^{\heartsuit} $Microsoft Research Asia, {ruizetang@microsoft.com}\\
  ${}^{\clubsuit} $University of British Columbia, {contact@emilie.ma}, {\{fhackett, bestchai\}@cs.ubc.ca}\\
  ${}^{\diamondsuit} $University of Illinois Urbana-Champaign, \{yiming34, tyxu\}@illinois.edu\\
}

\iclrfinalcopy
\begin{document}

\maketitle

\begin{abstract}
Formal models are essential to specifying large, complex computer systems and verifying their correctness,
    but are notoriously expensive to write and maintain.
Recent advances in generative AI
    show promise in generating certain forms of specifications.
However, existing work mostly targets small code, not complete systems.
It is unclear whether 
    AI can deal with realistic system artifacts, as this requires abstracting their complex behavioral properties into formal models.
We present \bench{}, a benchmark that evaluates AI's
    ability to formally model large, complex systems.
We focus on concurrent and distributed systems, which are keystones 
    of today's critical computing infrastructures, encompassing operating systems and cloud infrastructure.
We use \tla{}, the {\it de facto} specification language for concurrent and distributed systems,
    though the benchmark can be extended to other specification languages.
We address the primary challenge of evaluating AI-generated models by 
    automating metrics like
    syntactic and runtime correctness,
    conformance to system code,
    and invariant correctness.
\bench{} currently includes \numberofsystems{} diverse system artifacts: the Raft implementation
    of Etcd and Redis, \plsreviewinline{the leader election of ZooKeeper}, the Spinlock, Mutex, and \plsreviewinline{Ringbuffer} in Asterinas OS, etc., with more being added.
\bench{} enables us to understand the capabilities and limitations of today's LLMs 
    and agents, putting tools in this area on a firm footing and opening up promising new research directions.
\end{abstract}

\section{Introduction}
\label{sec:intro}

Formal models are essential to specifying computer systems and 
    reasoning about their correctness.
They provide a mathematical foundation 
    to document and verify the {\it design} of complex systems,
    such as distributed protocols and concurrent algorithms~\citep{lamportbook,tasiran2003,newcombe2015,hackett2023}.
Recently, formal models are used to describe system {\it implementations}---system code that 
    runs on user devices and in production environments.
Such models, which we refer to as {\it system models},
    enable verification of system code via comprehensive testing and model checking~\citep{bornholt2021lightweight,tang2024sandtable,ouyang:eurosys:25,atc25-tang}.
For example, system models of Apache ZooKeeper (a distributed coordination system) were used 
    to detect deep bugs that violate system safety and verify their fixes~\citep{ouyang:eurosys:25}.

However, system models are notoriously expensive to write and maintain.
Different from protocols and algorithms, system code contains low-level details, is more complex,
    and constantly evolves.
Hence, synthesis of system models is an open challenge (e.g.,~\citet{tla-challenge}).

Recent advances in generative AI, represented by large language models (LLMs) and agentic techniques,
    show promise in generating function-level specifications, in the form of
    pre- and post-conditions~\citep{rego2025,cao2025acl,xie2025,chakraborty2025formalspeccpp,ma2024specgen}.
It indicates that AI techniques can capture certain behaviors of software programs.
However, it is unclear whether AI could effectively model a complex system,
    which requires altogether different capabilities
    than the synthesis of pre- and post-conditions of a function.
Modeling a system requires the AI to understand the system design (e.g., the underlying protocols and algorithms),
    reasoning about safety and liveness under unexpected faults and external events,
    and abstracting system behavior into an executable program.
It is unclear to what extent AI has such capabilities.

\begin{figure}
    \centering
    \includegraphics[width=0.975\textwidth]{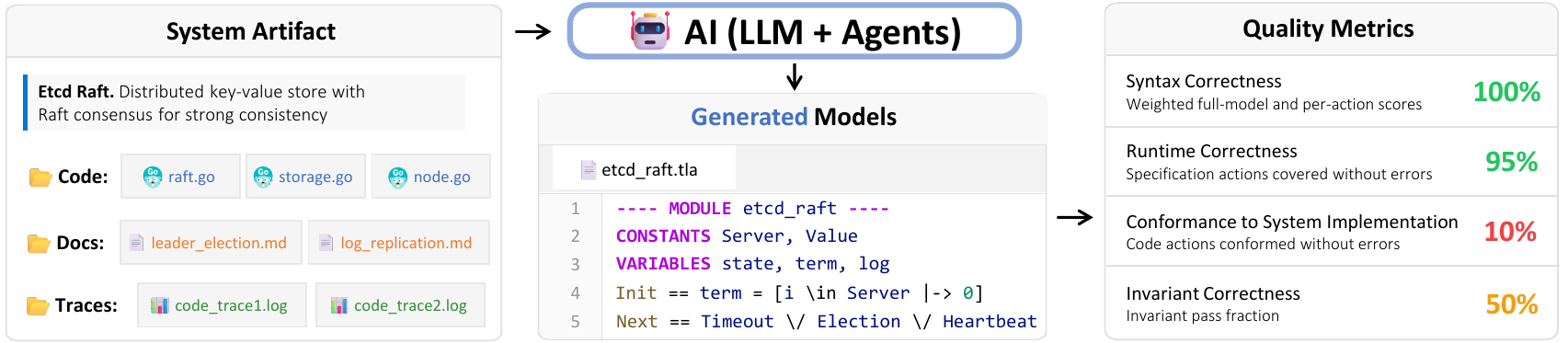}
     \vspace{5pt}
    \caption{\bench{} sources its tasks from real-world systems (e.g., Etcd Raft in the figure).
    It automatically evaluates the system models in \tla{} generated by AI with different metrics.}
    \vspace{-10pt}
    \label{fig:agent-arch}
\end{figure}

In this paper, we present \bench{}, a benchmark to evaluate AI's
    ability to formally model complex systems.
We target all forms of generative AI, including LLMs and agentic techniques.
We focus on {\it concurrent and distributed systems},
    which are especially difficult to model.
They also underpin today's critical computing infrastructure, which includes operating systems and cloud computing.
We focus on \tla{}, the {\it de facto} formal specification language for concurrent and distributed systems \plsreviewinline{(\S\ref{sec:background})}.
\plsreviewinline{\bench{} can be easily extended to support other specification languages
    such as Alloy~\citep{jackson2012alloybook}, PAT~\citep{sun2009pat},
    P~\citep{desai2013p}, and SPIN~\citep{holzmann1997model}.
We have added the support for Alloy and PAT
    (see Appendix~\ref{appendix:extensibility}).}

The key challenge of \bench{} is to {\it automatically} evaluate
    AI-generated models---how can we tell if a system model is of high quality?
We did not find any directly applicable metrics in use by existing work on \tla{} specification generation.
For example, \citet{cao2025acl} only check if the generated \tla{} specification
    can be run by the TLC model checker~\citep{yu99tlc}. But, successfully running TLC 
    is not an indicator of whether the model correctly describes the system.
One approach is to evaluate AI-generated pre-/post-conditions~\citep{rego2025,ma2024speceval} against
    human-written reference specifications. 
However, such a comparison can be brittle, and 
    real-world systems rarely have such low-level specifications.
Writing 
        a system model remains a highly challenging expert task that requires months \plsreviewinline{to years} of effort.

A key contribution of our benchmark is %
    quality metrics that can be \emph{automatically} checked. %
\plsreviewinline{These metrics reflect the fundamental requirements of a formal system model
    for use cases like formal verification~\citep{lamportbook} and model-driven testing~\citep{clarke2018model}.}
\begin{packed_itemize}
    \item {\bf Syntax correctness.} We statically check whether the generated system model uses valid \tla{} syntax using the
        SANY Syntactic Analyzer.
    \item {\bf Runtime correctness.} We check how much 
        of the generated \tla{} can be executed
        using the TLC model checker~\citep{yu99tlc}, which is a proxy for logical self-consistency.
    \item {\bf Conformance.} We measure whether the model conforms to the system
        implementation via trace validation~\citep{cirstea2024trace,atc25-tang,hackett2025tracelink}.
    \item {\bf Invariant correctness.} We model-check the system model against 
        system-specific invariants
        that reflect the system's safety and liveness properties.
\end{packed_itemize}

\bench{} currently includes \numberofsystems{} real-world artifacts, including 
    distributed systems like Etcd, Redis, \plsreviewinline{and ZooKeeper}, and concurrent systems like spinlock, mutex, \plsreviewinline{and ringbuffer} from Asterinas OS.
We also include system artifacts synthesized by PGo~\citep{hackett2023pgo} to evaluate
    AI's ability to comprehend \emph{generated} system code.
More system artifacts are actively being added.

\bench{} enables us to understand the capabilities and limitations of AI 
    in \plsreviewinline{using \tla{}} to model real-world systems
    by evaluating different agent designs with various AI models. 
State-of-the-art LLMs show good performance in modeling small system artifacts such as 
    a spinlock implementation.
On the other hand, these LLMs show limited ability in comprehending 
    and abstracting large, complex systems such as a Raft implementation~\citep{ongaro2014search}.
Overall, we believe that \bench{} can spur innovative AI approaches in the context of
    formal system models, similar to the role of SWE-bench~\citep{jimenez2024swebench} in software engineering.

Here is a snapshot of \bench{}: {\footnotesize \url{https://anonymous.4open.science/r/SysMoBench-BA9F/}}.

\newpage
\section{Background}
\label{sec:background}

\bench{} focuses on formal models written in \tla{}~\citep{lamportbook}, 
\plsreviewinline{which is the {\it de facto} formal specification language for modeling distributed and concurrent systems in practice.
The choice of \tla{} is made from a practical standpoint, not a language standpoint (\bench{} supports other specification
    languages; Appendix~\ref{appendix:extensibility}).
\tla{} is widely used by software companies like Amazon,
    Microsoft, Nvidia, Google, Oracle, etc (see~\citet{tla-industrial-use})
    to check and verify critical infrastructure systems such as distributed consensus systems (e.g., Etcd and ZooKeeper),
    confidential consortium frameworks~\citep{howard2025ccf},
    databases (e.g., CosmosDB and MongoDB),
    OS kernel synchronization~\citep{atc25-tang}, 
    and cache coherence~\citep{beers2008pre}.}

A \tla{} model specifies system behaviors as a collection of state variables, 
    an initial predicate that defines their initial values, 
    a next-state relation that determines state transitions, 
    and temporal properties that specify correctness requirements.
The next-state relation is expressed as multiple actions, each describing an atomic state update of all variables.
\plsreviewinline{\tla{} is built upon the Temporal Logic of Actions (TLA), which {\it includes} and {\it extends} standard linear temporal logic (LTL)~\citep{pnueli1977temporal},
    providing a rigorous mathematical foundation for reasoning about system behavior over time.}
\tla{} models can be verified using explicit-state 
    model checking via TLC~\citep{yu99tlc}, symbolic model checking via Apalache~\citep{konnov2019apalache}, and deductive verification via the \tla{} Proof System~\citep{chaudhuri2010tlaps}.
In \bench{},
    we primarily use TLC, the most widely used \tla{} tool that systematically explores all reachable states of a system model 
    to ensure that properties hold over the entire state space.
\plsreviewinline{These characteristics make \tla{} particularly well-suited for modeling complex concurrent and distributed systems.}

\begin{figure}[htbp]
    \vspace{5pt}
    \begin{subfigure}[t]{0.45\linewidth} 
\begin{minted}{rust}
pub struct SpinLock<T> { lock: AtomicBool }
pub struct SpinLockGuard<T, R: SpinLock<T>> {
  guard: R,
}
impl<T> SpinLock<T> {
  pub const fn new(val: T) -> Self {
    SpinLock { lock: AtomicBool::new(false) }
  }
  pub fn lock(&self) -> SpinLockGuard<T> {
    self.acquire_lock();
    SpinLockGuard { guard: self }
  }
  fn acquire_lock(&self) {
    while !self.try_acquire_lock() {}
  }
  fn try_acquire_lock(&self) -> bool {
    self.lock.compare_exchange(false, true)
      .is_ok()
  }
  fn release_lock(&self) { // on guard drop
    self.lock.store(false);
  }
}
\end{minted}
        \label{fig:spin-code}
    \end{subfigure}%
    \hspace{1.5pt}%
    \begin{minipage}[t]{1pt}
        {\gra\rule{0.5pt}{6cm}}
    \end{minipage}
    \hspace{0.6em}%
    \begin{subfigure}[t]{0.53\linewidth} 
        \input{code/spin-spec.tex}
        \label{fig:spin-spec}
    \end{subfigure}
    \vspace{7.5pt}
    \caption{Simplified code that implements a spinlock in Asterinas (left) 
        and an AI-generated \tla{} model (right).
        A spinlock represents the simplest system in \bench{}.}
    \vspace{5pt}
    \label{fig:spin-code-spec}
\end{figure}

Figure~\ref{fig:spin-code-spec} shows simplified code that implements a spinlock in the Asterinas
    operating system~\citep{peng2025asterinas} and 
    the corresponding \tla{} model that describes the code.
The \tla{} model is generated by the AI agent we evaluate in \S\ref{sec:eval}.
The model defines constants such as \code{\small Threads} (line 1) 
    and system-state variables such as \code{\small lock\_state} and \code{\small pc} (line 2).
The initial state \code{\small Init} (line 3) assigns initial values to all variables.
Three actions are defined (lines 5--19)---\code{\small StartLock}, \code{\small Acquire}, and \code{\small Unlock}---corresponding to the code logic,
    where \code{\small Acquire} combines the logic of \code{\small acquire\_lock} and \code{\small try\_acquire\_lock}.
Each action is enabled by certain conditions, e.g., \code{\small StartLock} is enabled when a thread's \code{\small pc} is ``idle'';
it then assigns next-state values to all variables.

To model the spinlock implementation, the AI must first understand
the behavior of each function. Next, it must decide how to represent
the system. This involves introducing variables, such as auxiliary
ones like \code{\small pc}, and defining atomic actions that preserve
concurrency semantics. Finally, the AI must specify correctness
properties. For example, mutual exclusion (line 22) requires that in
every state, at most one thread can be in the ``locked'' state.

Note that \bench{} concerns formal models of system implementations, or
    {\it system specifications} in the \tla{}
    and formal method literature.
As a specification,
    a system model enables verification 
    of system code, but does not necessarily capture requirements of the design~\citep{stoica2024specificationsmissinglinkmaking}.
\bench{} does not target other forms of specifications, such as 
    formal proofs~\citep{chen2025}
    or function-level pre- and post-conditions~\citep{ma2024specgen}.

\section{\bench{}}
\label{sec:benchmark}

\bench{} is a benchmark that uses real-world distributed and concurrent system
    artifacts to evaluate AI's ability to formally model systems.
Table~\ref{tab:systems} lists the systems that have been integrated in \bench;
we are actively adding more system artifacts (\S\ref{sec:ext}).

\begin{table}[htbp]
    \centering
    \footnotesize
    \caption{System artifacts that have been integrated in the \bench{};
        ``\tla{} LoC'' refers to the AI-generated \tla{} models presented 
        in our evaluation results (\S\ref{sec:eval}).}
    \label{tab:systems}
    \begin{tabular}{lllccc}
    \toprule
        System & Type & Desc. & Source Lang. & Source LoC & \tla{} LoC \\ 
    \midrule
        \systemlink{https://github.com/asterinas/asterinas}{Asterinas Spinlock} & Concurrent  & Synchronization   & Rust & 213   & 151\\
        \systemlink{https://github.com/asterinas/asterinas}{Asterinas Mutex}    & Concurrent  & Synchronization   & Rust & 186   & 219\\
        \systemlink{https://github.com/asterinas/asterinas}{Asterinas Rwmutex}  & Concurrent  & Synchronization   & Rust & 395   & 250\\
        \systemlink{https://github.com/asterinas/asterinas}{Asterinas Ringbuffer} & Concurrent  & Data Structure   & Rust & 615   & 123\\
        \systemlink{https://github.com/etcd-io/raft}{Etcd Raft}                 & Distributed & Consensus (Raft)  & Go   & 2,159 & 385 \\
        \systemlink{https://github.com/RedisLabs/redisraft}{Redis Raft}         & Distributed & Consensus (Raft)  & C    & 2,394 & 349 \\
        \systemlink{https://github.com/xline-kv/Xline}{Xline CURP}              & Distributed & Replication (CURP)  & Rust & 4,064 & 100 \\
        \systemlink{https://github.com/apache/zookeeper}{ZooKeeper FLE}              & Distributed & Leader Election  & Java & 5,360 & 141 \\
        \systemlink{https://github.com/UBC-NSS/pgo}{PGo dqueue}                 & Distributed & Distributed Queue & Go   & 175   & 75 \\
        
        \systemlink{https://github.com/UBC-NSS/pgo}{PGo locksvc}                                                       & Distributed & Lock Server       & Go  & 281   & 93 \\
        \systemlink{https://github.com/UBC-NSS/pgo}{PGo raftkvs}                                                     & Distributed & Consensus (Raft)  & Go  & 3,163 & 508 \\
        \bottomrule
    \end{tabular}
\end{table}

\subsection{Task Formulation}
\label{sec:task}

A \bench{} task is to generate a system model for a given system artifact (Table~\ref{tab:systems}).
\bench{} does not concern how the system model is generated. It
    can be generated by prompting LLMs directly, with few-shot learning, 
    or with agentic techniques that invoke external tools (we evaluate both in \S\ref{sec:eval}).
Since system artifacts in \bench{} are real-world system projects,
    one can feed various data sources to the LLMs/agents, such as 
    source code, documents, and runtime traces. 
The task mirrors real-world modeling workflows of human engineers.

Each task specifies the granularities at which
    to model the target system's 
    essential behavioral properties and state transitions.
\begin{plsreview}
The required level of granularity is defined based on target use cases;
    our current use case is model-checking based system verification---we require
    the same level of detail as in prior work on verification and bug finding.
The model must include core actions that interact with other components,
    while excluding implementation details unrelated to system behavior.
We evaluate behavioral conformance rather than structural equivalence,
    allowing fine-grained modeling of core actions as long as they preserve semantic obligations needed for verification.
To make requirements concrete,
    each task lists core actions that must be modeled and actions that should be excluded.
Take Spinlock as an example (Figure~\ref{fig:spin-code-spec}):
    the requirements are specified as follows:
\end{plsreview}
\begin{promptbox}
    \scriptsize
    \ttfamily
    \textbf{Mandatory core actions that must be modeled:}
    \vspace{-7.5pt}
    \begin{packed_itemize}
        \item The model must specify lock() and unlock() actions.
        \item Atomic compare\_exchange operation on the lock variable.
        \item Spinning when the lock is contended.
    \end{packed_itemize}

    \textbf{Actions that should be excluded from the model:}
    \vspace{-7.5pt}
    \begin{packed_itemize}
        \item RAII guard implementation details.
        \item Non-core details (e.g., debug formatting and trait implementation).
    \end{packed_itemize}
\end{promptbox}
Besides, the task also requires generating a TLC configuration as a part of the model.

\subsection{Metrics and Their Measurement}
\label{sec:metrics}

Key contributions of \bench{} are to 
    (1) define metrics that can 
    fairly measure the quality of AI-generated \tla{} models, and 
    (2) design practical techniques to automate metric measurements.
\bench{} does not rely on human evaluation
    which is slow and hard to scale, especially for complex real-world systems.
We do not consider LLM-as-a-judge approaches, as we find these unreliable
    and difficult to interpret.

\begin{figure}
\centering
\vspace{-2pt}
\includegraphics[width=0.9\textwidth]{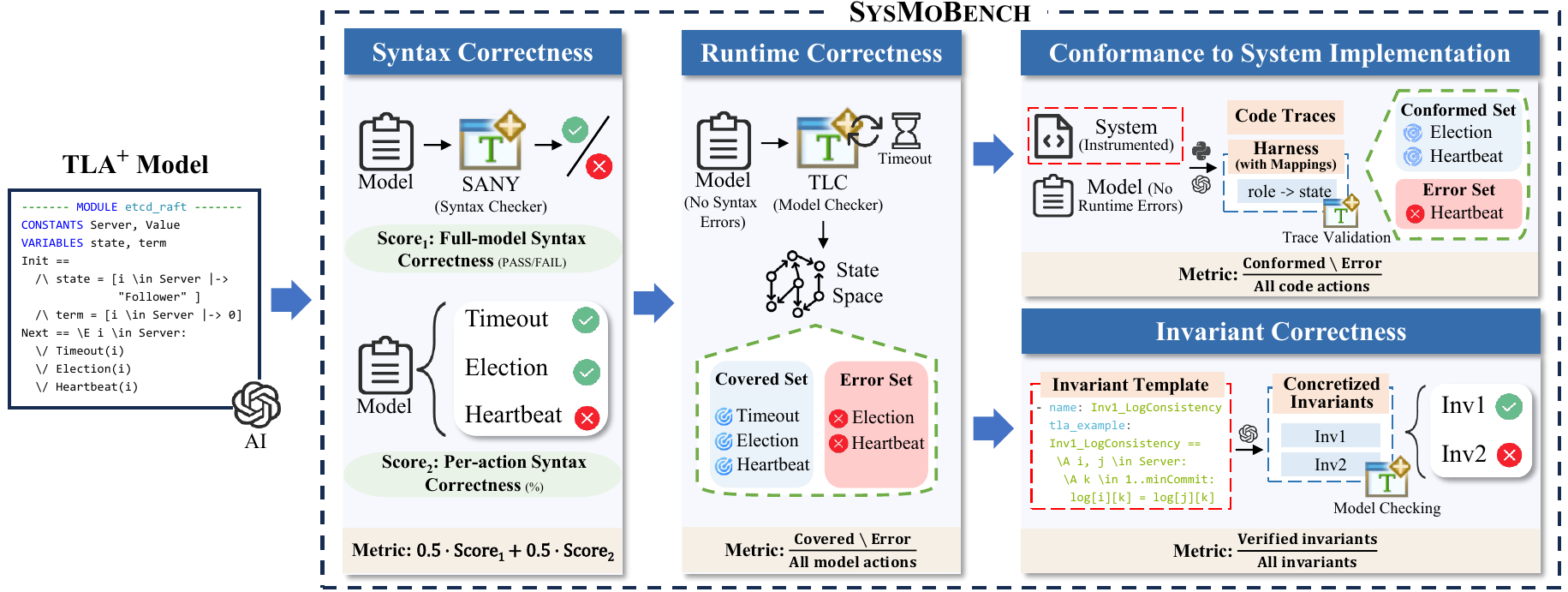}
\caption{Metrics and evaluation workflow of \bench{}. The red dashed boxes denote inputs provided by the system artifact: instrumented system for code traces and required invariants.}
\label{fig:metrics}
\end{figure}

\bench{} includes four metrics that evaluate a \tla{} model
    on syntax (\S\ref{sec:syntax}), runtime correctness (\S\ref{sec:error}), 
    conformance to system code (\S\ref{sec:conformance}),
    and invariant correctness (\S\ref{sec:inv}).
The metrics are not independent, e.g., a model with syntax errors cannot be evaluated for other metrics.
An executable model is evaluated for 
    both conformance and invariant correctness.
We design {\it partial scoring} schemes for every metric 
    and normalize results to 
    percentage values, making them easy to interpret.
Figure~\ref{fig:metrics} illustrates the metrics and the evaluation workflow.

\vspace{-3pt}
\subsubsection{Syntax Correctness}
\label{sec:syntax}
\vspace{-3pt}

\bench{} uses the \tla{} SANY Syntactic Analyzer~\citep{lamportbook} to check the syntax
    of the \tla{} models against
    \tla{} grammar rules, operator usage, module structure, etc.
Note that SANY checks the entire model specification.
If the model specification passes the SANY checks, it earns a full score.
However, many AI-generated models fail the SANY check;
    therefore, we need fine-grained analysis for partial scoring.

\bench{} offers per-action analysis for partial scoring by checking how many generated actions
    are erroneous (and failed SANY).
It encapsulates each action in the model
    into a per-action model by adding necessary dependencies (e.g., constant declarations, 
    variable definitions, etc.). It then uses SANY to check the \plsreviewinline{syntax of} per-action model
    \plsreviewinline{(only syntax correction is concerned in this step; no equivalence check).}
A partial score $S$ represents the percentage of correct actions $n_c$ among the total actions $n_t$,
    i.e., $S = \frac{n_c}{n_t}$.
\plsreviewinline{Here, $n_c$ is determined by running SANY on each per-action module and counting those that pass without syntax errors,
    while $n_t$ is obtained by counting all action definitions in the original model.}
Note that the per-action checks do not account for inter-action dependencies: 
    a model that passes all the per-action checks can still fail.
We use a weighted scoring scheme that gives equal weights to per-action correctness 
    and inter-action correctness. A model that passes the overall SANY check 
    earns 100\%, while only passing all per-action checks earns 50\%.
Only system models with 100\% syntax scores will be evaluated for other metrics, because
    models with syntax errors cannot be compiled or executed (which is required by other metrics).

\vspace{-3pt}
\subsubsection{Runtime Correctness}
\label{sec:error}
\vspace{-3pt}

For a syntactically correct system model,
    \bench{} next
    evaluates if the model can be executed correctly.
To do so, \bench{} performs bounded model checking and simulation using TLC, 
    and then observes covered actions and runtime errors (if any) \plsreviewinline{by parsing TLC's coverage report and error output}.
This model checking and simulation explores the state space
    without any invariant checking (see \S\ref{sec:inv}).
During this state space exploration, \bench{} records all covered actions and the actions with runtime errors.

We define a metric $M_r$ that represents the coverage of actions without runtime errors:
    $M_r = \frac{n_r}{n_t}$, where $n_r$ is the number of covered actions
    that did not report errors during state exploration, and
    $n_t$ is the total number of actions in the model.

Models with no runtime errors can then 
    be executed to explore state space.
Only such models are
    evaluated for conformance and invariant correctness.

\vspace{-3pt}
\subsubsection{Conformance to System implementation}
\label{sec:conformance}
\vspace{-3pt}

For an executable model, \bench{} evaluates its conformance to the behavior of the system implementation
    using trace validation~\citep{cirstea2024trace}.
Trace validation checks whether a trace of the system execution 
    corresponds to a path in the model's state space.
\bench{} supports trace validation mechanisms used by different systems~\citep{atc25-tang,cirstea2024trace,hackett2025tracelink}.

Specifically, to collect execution traces, system code is instrumented with logging statements.
The instrumentation granularity matches the granularity requirements of the task.
If the AI-generated model is coarser than the trace logs, conformance checking could fail; 
    otherwise, we use missing-event inference techniques~\citep{atc25-tang} to account for uninstrumented actions.

The key challenge of automatic conformance checking of any AI-generated models is 
    to correctly map the elements in the model to elements in the system execution log. This is because AI
    will often use names that differ from those in system code.
We solve this problem by using a coding LLM (e.g., Claude-Sonnet-4) to 
    (1) extract constants, variables, and actions from the input model
    and (2) map them to the corresponding elements specified in the task requirement (\S\ref{sec:task}).

The use of LLMs for automatic mapping of elements in the model and code may raise 
    reliability concerns.
In our experience, state-of-the-art LLMs accomplish the mapping task reliably (\S\ref{sec:setup}).
This is because (1) the mapping task is simple and well-defined;
    (2) the generated models are derived from the system artifacts and thus 
        largely follow the naming conventions of the system;
    and, (3) our trace validation technique~\citep{atc25-tang} can tolerate a certain level of missing variables or actions
        though we have not found such cases so far.
A similar use of LLMs for mapping is adopted in~\citet{tlaibench} (discussed in \S\ref{sec:related}).

During trace validation across all traces,
\bench{} keeps track of code actions that are covered and those code actions that trigger errors.
\plsreviewinline{Specifically, \bench{} feeds the trace to TLC along with the model,
    and records whether TLC successfully validates the trace.
    If validation fails, \bench{} identifies where the mismatch occurred by analyzing TLC's trace validation output.}
To measure conformance, we define $M_c$ as the coverage of code actions without conformance errors: $M_c = \frac{n_c}{n_t}$,
where $n_c$ is the number of code actions that were covered during validation with no errors, and $n_t$ is the total number of actions in the instrumented code.
We use instrumented code actions instead of model actions because this provides a stable, implementation-grounded granularity that is consistent across different AI-generated models.

\vspace{-3pt}
\subsubsection{Invariant Correctness}
\label{sec:inv}
\vspace{-3pt}

\bench{} also evaluates whether AI-generated models always satisfy invariants that 
    describe the expected safety and liveness properties of the system.
In principle, if a system model fully conforms to code, 
    violations of these invariants would indicate bugs in system code;
    in practice, few AI-generated models achieved fine-grained conformance.
\plsreviewinline{Nevertheless, AI-generated models have demonstrated practical utility by successfully reproducing 
    known bugs from previous system versions (Appendix~\ref{appendix:bug-reproduction}).}
Table~\ref{tab:spinlock_inv} lists the invariants for the spinlock code in Figure~\ref{fig:spin-code-spec}.
These invariants are part of the benchmark defined by the task (\S\ref{sec:ext}).

\begin{table}[htbp]
    \centering
    \caption{Example spinlock invariants}
    \label{tab:spinlock_inv}
    \begin{tabular}{lll}
    \toprule
    {Invariants} & {Description} & {Type} \\ 
    \midrule
    {Mutual exclusion} & At most one process can be in the critical section at any time & Safety  \\
    {Lock consistency} & The lock state accurately reflects critical section occupancy & Safety \\
    {No deadlock} & Not all threads can be stuck spinning simultaneously & Safety  \\
    {Guard lifecycle} & Every thread eventually releases the lock it acquires & Liveness  \\
    {Eventual release} & The system eventually reaches a state where all threads are idle & Liveness \\ 
    \bottomrule
    \end{tabular}
\end{table}

\bench{} addresses a similar challenge as in \S\ref{sec:conformance}: it needs to automatically map 
    the actions, variables, and data structures in the system model to those expressed in the invariants.
For this, the invariants in \bench{} are templates that 
    contain a description of the property, formal definitions, and example \tla{} invariants. 
We then use an LLM to translate these templates into model-specific invariants that can be 
    checked against the system model.
For example, the following template defines the mutual exclusion invariant in Table~\ref{tab:spinlock_inv}:
\begin{promptbox}
    \scriptsize
- name:"MutualExclusion"\\
\hspace*{9pt}type:"safety"\\
\hspace*{9pt}natural\_language:"Only one thread can access a shared resource at a time"\\
\hspace*{9pt}formal\_description:"No more than one thread in the critical section"\\
\hspace*{9pt}tla\_example:\texttt{\footnotesize MutualExclusion}
    \texttt{\footnotesize ==}
    \texttt{\footnotesize Cardinality(\{t}
    \texttt{\footnotesize \textbackslash{}in} 
    \texttt{\footnotesize Threads:}\texttt{\footnotesize pc[t]}
    \texttt{\footnotesize =}
    \texttt{\footnotesize "in\_cs"\})} 
     \texttt{\footnotesize <=} 
    \texttt{\footnotesize 1}
\end{promptbox}
\bench{} prompts the LLM with both the invariant template and the system model
    and asks it to concretize the template using the model.
This mapping is highly structured: the output substitutes 
    the template's variables and constants with those in the model.
For example, the mutual exclusion invariant,
    \texttt{\footnotesize Cardinality(\{t} 
    \texttt{\footnotesize \textbackslash{}in} 
    \texttt{\footnotesize Threads:} 
    \texttt{\footnotesize status[t]} 
    \texttt{\footnotesize =} 
    \texttt{\footnotesize "locked"\})} 
    \texttt{\footnotesize <=} 
    \texttt{\footnotesize 1},
    is a concretization of the template by replacing \texttt{\footnotesize pc} with \texttt{\footnotesize status} and \texttt{\footnotesize in\_cs} with \texttt{\footnotesize locked}.
We evaluate the reliability of this LLM-assisted concretization in \S\ref{sec:setup}.

The invariants are used by TLC during model checking,
  and \bench{} observes whether each invariant is violated.
\plsreviewinline{Specifically, for each invariant, \bench{} creates a separate model with that invariant
    and runs TLC independently.
    This allows \bench{} to record whether each invariant is violated.} %
We define a metric $M_i$ that represents the \textit{fraction of invariants passed}, denoted as $M_i = \frac{n_i}{n_t}$,
where $n_i$ is the number of invariants that hold across the explored state space,
and $n_t$ is the total number of invariants defined for the model.
Models with a higher $M_i$ are of higher quality.
When combined with runtime and conformance coverage metrics, a higher $M_i$
increases confidence in the correctness of the input specification.

\vspace{-5pt}
\subsection{Adding New Systems and Specification Languages to \bench{}}
\label{sec:ext}
\vspace{-5pt}

\bench{} provides an extensible framework to add more real-world system artifacts.
To add a new artifact to \bench{}, one needs to 
    (1) prepare the system artifact (e.g., source code and documents);
    (2) create a new task that specifies the abstractions and components to model (\S\ref{sec:task}); 
    (3) develop invariant template (\S\ref{sec:inv}) that specifies correctness properties (safety and liveness);
    and (4) provide harness for trace validation by instrumenting system code.
In our experience, the effort to add a new system artifact to \bench{} is manageable.
For example, adding Etcd Raft took one \bench{} author four days;
an Xline CURP contributor with no experience of \bench{} added the system to \bench{} in four days.
Most of the effort is spent on instrumenting the system to collect execution logs for trace validation
    in order to measure conformance.
Unlike other benchmarks (\S\ref{sec:related}),
    \bench{} does not require writing reference models;
    in fact, we hope that some of the AI-generated models can eventually be adopted by real-world system projects.

\begin{plsreview}
\bench{} is extensible to formal specification languages other than \tla{}.
We extended \bench{} to support Alloy~\citep{jackson2012alloybook} and PAT~\citep{sun2009pat}, 
demonstrating its generality.
Details of these extensions and preliminary evaluation are presented in Appendix~\ref{appendix:extensibility}.
The results show that while our framework is extensible, \tla{} remains the practical 
    choice and can benefit from AI-driven techniques (existing LLMs are less familiar with Alloy and PAT).
\end{plsreview}

\vspace{-5pt}
\section{Evaluation Setup}
\label{sec:setup}
\vspace{-5pt}

To evaluate AI's system modeling abilities,
we use three agents powered by LLMs.
\begin{packed_itemize}
    \item {\bf Basic Modeling Agent.} This agent reflects the LLM's raw modeling abilities.
    The agent prompts an LLM with the source code of the system and the task requirement (\S\ref{sec:task}).
    The detailed prompts are documented in Appendix~\ref{sec:model_agent}.
    \item {\bf Code Translation Agent.} This agent
    uses an LLM to \emph{translate} system code into an equivalent \tla{} form.
    The agent translates code statement by statement (from the source language to \tla{}),
    and then organizes the control flows of the translated statements into a \tla{} model.
    The agent reflects the capabilities of LLM-based code translation.
    We adopt the implementation of~\citet{specula} as our code translation agent. 
    \item {\bf Trace Learning Agent.} This agent does not use code as input, but tries to 
        learn the system model from system traces.
    It prompts LLMs with the traces to infer the system model (see Appendix~\ref{sec:trace_agent}).
        This agent reflects the capability of automata learning~\citep{Biermann72} with LLMs.
\end{packed_itemize}
We follow HumanEval~\citep{chen2021evaluatinglargelanguagemodels} to run each agent five times 
    and evaluate the best output model.
The agents can enhance the model with feedback loops (three iterations are allowed) if the generated model
    cannot pass compilation or has runtime errors.
    No human intervention is allowed.

We use four different LLMs to power the three agents: Claude-Sonnet-4 (20250514), GPT-5 (20250807), Gemini-2.5-Pro (20250617), and DeepSeek-R1 (20250528).
We run the SANY Syntactic Analyzer, TLC model checker, and system code (for conformance checking)
    on a server with dual AMD EPYC 7642 48-Core Processors and 256GB RAM running Ubuntu 22.04.

\textbf{Robustness of LLM-assisted Components.}
\bench{} uses LLM-assisted techniques to map elements in an AI-generated \tla{} model to those in the system logs (\S\ref{sec:conformance}),
    and to concretize invariant templates (\S\ref{sec:inv}).
We inspected the LLM-assisted mapping and concretization, and found the results to be correct. 
We also conducted an experiment using the ``gold model'' for Etcd Raft and Asterinas spinlock, which are known to be correct.
    We created 10 models (5 for each system) by changing the names of the variables and actions and tweaking the model's granularity.
The gold models achieve a perfect score on all metrics, empirically validating the quality of our metrics.

\begin{plsreview}
\textbf{Training Data Contamination.}
One may be concerned about the fairness of \bench{} because it uses open-source projects where the
    system code likely already appears in LLM training data.
In fact, it is intended to have system code in LLM training data.
The design mirrors how human engineers write formal models: 
    they first learn system code before writing formal models.
Our goal is to leverage LLMs to write effective \tla{} models for important, safety-critical software systems,
    which requires LLMs to have internalized knowledge of these systems.
Note that this is different from coding benchmarks in that we ask LLMs/agents to write existing code.

Second, few system artifacts in \bench{} have \tla{} system models in their open-source repositories. 
The \tla{} models of Asterinas Spinlock/Mutex/Rwmutex are never released. Redis Raft and Xline CURP do not have any TLA+ models. 
Etcd Raft and PGo systems do have TLA+ models in the repositories. However, those models are for protocols, 
    not for system code.
Our goal is to use AI to write TLA+ models for all important,
    safety-critical software systems in the wild.
\end{plsreview}

\vspace{-8pt}
\section{Results}
\label{sec:eval}
\vspace{-8pt}

We present evaluation results for the \emph{basic modeling agent} and the \emph{code translation agent} 
    on Asterinas Spinlock and Etcd Raft (Table~\ref{tab:merged-results}).
Appendix~\ref{appendix:results} contains the complete results for all systems in \bench{}.
We omit the results of the trace learning agent (which fails to pass runtime checks).

\begin{table}
    \centering
    \caption{Evaluation results of two AI agents on two representative system artifacts. 
        \checkmarksymbol {} and \wrongsymbol {} mark whether the model is evaluated in the next phase of measurements (see Figure~\ref{fig:metrics}).}
    \label{tab:merged-results}
    \footnotesize
    \begin{subtable}{\textwidth}
        \centering
        \caption{Asterinas Spinlock}
        \label{tab:spinlock-result}
        \vspace{-2pt}
        \begin{tabular}{llcccc}
            \toprule
        \bf Agent & \bf LLM & \multicolumn{1}{l}{\bf Syntax} & \multicolumn{1}{l}{\bf Runtime} & \multicolumn{1}{l}{\bf Conformance} & \multicolumn{1}{l}{\bf Invariant} \\ \midrule
        \multirow{4}{*}{\begin{tabular}[c]{@{}l@{}}Basic Modeling\end{tabular}} & Claude-Sonnet-4 & \excellentcell{}100.00\%\checkmarksymbol{} & \excellentcell{}100.00\%\checkmarksymbol{} & \excellentcell{}100.00\% & \excellentcell{}100.00\% \\
        & GPT-5 & \excellentcell{}100.00\%\checkmarksymbol{} & \excellentcell{}100.00\%\checkmarksymbol{} & \excellentcell{}80.00\% & \excellentcell{}100.00\% \\
        & Gemini-2.5-Pro & \excellentcell{}100.00\%\checkmarksymbol{} & \excellentcell{}100.00\%\checkmarksymbol{} & \excellentcell{}80.00\% & \excellentcell{}85.71\% \\
        & DeepSeek-R1 & \excellentcell{}100.00\%\checkmarksymbol{} & \excellentcell{}100.00\%\checkmarksymbol{} & \excellentcell{}80.00\% & \excellentcell{}100.00\% \\ \midrule
\multirow{4}{*}{Code Translation} & Claude-Sonnet-4 & \excellentcell{}100.00\%\checkmarksymbol{} & \excellentcell{}100.00\%\checkmarksymbol{} & \excellentcell{}100.00\% & \excellentcell{}100.00\% \\
        & GPT-5 & \excellentcell{}100.00\%\checkmarksymbol{} & \excellentcell{}100.00\%\checkmarksymbol{} & \excellentcell{}100.00\% & \excellentcell{}85.71\% \\
        & Gemini-2.5-Pro & \excellentcell{}100.00\%\checkmarksymbol{} & \excellentcell{}100.00\%\checkmarksymbol{} & \excellentcell{}100.00\% & \excellentcell{}100.00\% \\
        & DeepSeek-R1 & \excellentcell{}100.00\%\checkmarksymbol{} & \excellentcell{}100.00\%\checkmarksymbol{} & \excellentcell{}100.00\% & \excellentcell{}100.00\% \\ \bottomrule
        \end{tabular}
    \end{subtable}

    \vspace{4pt} %

    \begin{subtable}{\textwidth}
        \centering
        \caption{Etcd Raft}
        \label{tab:etcd-result}
        \vspace{-2pt}
        \begin{tabular}{llcccc}
            \toprule
            \bf Agent & \bf LLM & \multicolumn{1}{l}{\bf Syntax} & \multicolumn{1}{l}{\bf Runtime} & \multicolumn{1}{l}{\bf Conformance} & \multicolumn{1}{l}{\bf Invariant} \\ \midrule
\multirow{4}{*}{\begin{tabular}[c]{@{}l@{}}Basic Modeling\end{tabular}} & Claude-Sonnet-4 & \excellentcell{}100.00\%\checkmarksymbol{} & \faircell{}25.00\%\checkmarksymbol{} & \faircell{}7.69\% & \goodcell{}69.23\% \\
         & GPT-5 & \faircell{}47.87\%\wrongsymbol{} & - & - & - \\
         & Gemini-2.5-Pro & \goodcell{}50.00\%\wrongsymbol{} & - & - & - \\
         & DeepSeek-R1 & \goodcell{}50.00\%\wrongsymbol{} & - & - & - \\ \midrule
\multirow{4}{*}{Code Translation} & Claude-Sonnet-4 & \excellentcell{}100.00\%\checkmarksymbol{} & \goodcell{}66.67\%\checkmarksymbol{} & \faircell{}15.38\% & \excellentcell{}92.31\% \\
         & GPT-5 & \excellentcell{}100.00\%\checkmarksymbol{} & \faircell{}20.00\%\wrongsymbol{} & - & - \\
         & Gemini-2.5-Pro & \faircell{}44.44\%\wrongsymbol{} & - & - & - \\
         & DeepSeek-R1 & \excellentcell{}100.00\%\checkmarksymbol{} & 0.00\%\wrongsymbol{} & - & - \\ \bottomrule
        \end{tabular}
    \end{subtable}
    \vspace{-10pt}
\end{table}

{\bf Modeling Capability.}
We focus on the results of the basic modeling agent.
The basic modeling agent can generate high-quality \tla{} models for Spinlock, 
    which is among the simplest artifacts in \bench{} (Table~\ref{tab:systems}),
    showing certain levels of 
    modeling capability.
However, for larger and more complex systems such as the distributed protocol implementations,
    the basic modeling agent performs poorly.
For Etcd Raft, only with Claude-Sonnet-4, the modeling agent reaches 
    the conformance and invariant checking,
    and scores are low.
Clearly, the complexity and size of Etcd Raft exceed the modeling ability of the LLMs and agents.

For Etcd Raft, the basic modeling agents struggle with
    (1) code verbosity, (2) protocol complexity, and (3) abstraction.
Etcd Raft has much more code than Spinlock, 
    with low-level utilities (e.g., for debugging) and implementation-specific comments, 
    which often cause agents to lose focus on essential system logic.
Moreover, the Raft protocol~\citep{ongaro2014search} has more complex logic than a spinlock
    in terms of ordering and intricate conditions of state transitions.
Both (1) and (2) make Etcd Raft significantly more challenging for LLMs to comprehend the system artifact.
For (3), Etcd Raft presents significant abstraction challenges: concepts like 
    distributed logs require nested data structures, demanding LLMs to precisely 
    express them using \tla{} language constructs.

The basic modeling agents also perform poorly on PGo systems (Appendix~\ref{appendix:results}),
indicating limited LLM ability to comprehend machine-generated systems.
Code in PGo-generated systems is a mix of compiler-generated patterns 
    and a runtime library (Appendix~\ref{sec:pgo}).
The generated code is repetitive, and, while it borrows some variable names from the source specification, 
    intermediate variables have synthetic, non-significant names, 
    which provide few semantic clues to an LLM (or a human reader).

\begin{figure}
    \centering
    \vspace{5pt}
    \includegraphics[width=\textwidth]{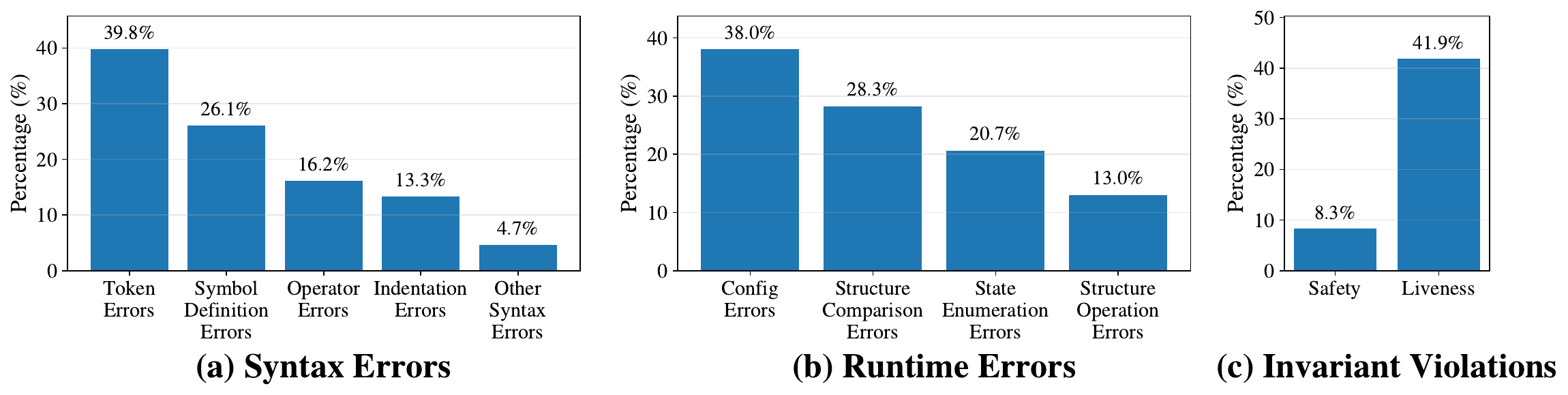}
    \caption{LLM error attribution regarding the \bench{} metrics in the basic modeling agent.
    The conformance metric is omitted as it has a single attribution.}
    \vspace{-5pt}
    \label{fig:error_comparison}
\end{figure}

{\bf Analysis on Agents.} For complex systems like Etcd Raft, the code translation agent outperforms the basic modeling agent.
We believe this is due to the powerful translation abilities of LLMs~\citep{yang2024exploring}.
Specifically, the code translation agent leverages symbolic control-flow analysis 
    to synthesize a \tla{} model rigorously.
The translation approach also prevents LLMs from hallucinating logic
    by adhering to system code. %
These results indicate that %
    leveraging LLMs' code translation abilities can assist in model generation.
Finally, we observed that LLMs would sometimes imitate classic \tla{} models from their training set,
    missing important system-specific content.

{\bf Analysis on Invariants.} 
For invariants, LLMs violate very different types of invariants---only 8.3\% of safety properties are violated 
    while 41.9\% of liveness properties were violated (Figure~\ref{fig:error_comparison}c).
This indicates the limited ability of LLMs in temporal reasoning.
\begin{plsreview}
To understand the nature of these violations,
    we conducted a fine-grained analysis categorizing them by root causes (Appendix~\ref{appendix:liveness-analysis}).
We find that while fairness assumption violations (e.g., missing or incorrectly specified fairness assumptions) 
    are a significant issue across systems,
    logical and structural errors tend to manifest earlier and block progress
    before fairness-related issues emerge.
\end{plsreview}

{\bf Analysis on LLMs.} 
We observe that LLMs constantly introduce syntax errors (Figure~\ref{fig:error_comparison}a), 
    especially
    GPT-5, Gemini-2.5-Pro, and DeepSeek-R1.
For example,
DeepSeek-R1 often misuses mathematical symbols (e.g., $\cap$, $\forall$) instead of ASCII \tla{} operators.
Gemini-2.5-Pro and GPT-5 often mix \tla{} syntax with those of other programming languages like Python.
LLMs also misuse operators with incorrect parameters and produce malformed indentation.
In terms of runtime errors (Figure~\ref{fig:error_comparison}b),
    LLMs frequently generate inconsistent TLC configurations,
    such as missing constants or mismatched declarations.
Misunderstanding of \tla{} data structures is also a common error, e.g., 
    comparing incompatible types or applying invalid operations (e.g., set operations on records).

\begin{plsreview}
Among all evaluated LLMs, Claude-Sonnet-4 in general outperforms others in most metrics across evaluated 
    system artifacts.
Since only syntax-valid models can proceed to subsequent evaluation phases,
    Claude-Sonnet-4's ability to generate syntactically correct \tla{} models provides an initial advantage.
However, Claude-Sonnet-4's strength extends beyond syntax correctness.
\bench{} decomposes the evaluation into four distinct metrics that separate syntactic correctness from reasoning about system behavior.
As shown in the Appendix~\ref{appendix:results}, 
    for models that successfully pass syntax checks, 
    Claude-Sonnet-4 generally still achieves higher scores on runtime, conformance, and invariant metrics compared to other LLMs.
\end{plsreview}

\begin{plsreview}
{\bf Qualitative Assessment.}
We conducted qualitative evaluation to assess AI-generated system model quality and utility
    in terms of bug finding (Appendix~\ref{appendix:qualitative}).
Comparing with human-written \tla{} models, AI-generated \tla{} models differ in structure and completeness
    but they capture essential system behaviors.
Despite these limitations, AI-generated models have successfully reproduced known bugs in five systems,
    demonstrating their practical utility for partial correctness checking.
\end{plsreview}

\section{Related Work}
\label{sec:related}

\bench{} is the first framework that evaluates AI on formally modeling 
    real-world systems.

\plsreviewinline{{\bf Benchmarks for Formal Specifications.}}
There are several benchmarks for evaluating AI (including LLMs and AI agents)
    on generating function-level pre-/post-conditions
    and loop invariants~\citep{rego2025,xie2025,cao2025acl,chakraborty2025formalspeccpp,ma2024specgen,wen2024enchanting}.
Those benchmarks typically use small programs, such as sample programs in VeriFast that implement data structures~\citep{rego2025} and LeetCode programs~\citep{ma2024specgen}.
There also exist benchmarks on proof generation for deductive software verification~\citep{yang2024autoverus}
    and on verified code generation~\citep{thakur2025clever,ye2025verina}.
None of these benchmarks target complex real-world computing systems as in \bench{}.
Fundamentally, those benchmarks evaluate AI's abilities of code comprehension and specification, 
    not system modeling.
\plsreviewinline{Similarly, PAT-Agent~\citep{zuo2025patagent} and Alloy-APR~\citep{alhanahnah2025empirical} 
    target smaller tasks such as puzzles and repairing injected errors (see Appendix~\ref{appendix:comparison}).}
As AI for code is becoming mature, the next step is capturing how AI can benefit
    practical verification of real-world systems. We developed \bench{} with this motivation in mind.
The arguably most related benchmark is~\citet{tlaibench} which evaluates 
    AI-generated \tla{} specifications.
Tasks in TLAiBench are primarily logic puzzles, not real-world systems.
TLAiBench is useful for evaluating AI's ability in \emph{using} the \tla{} language,
   not system comprehension or modeling.
Hence, TLAiBench and related benchmarks such as~\citet{cao2025acl} only measure 
    the syntax and runtime correctness of the \tla{} specifications.
\citet{li2025osvbench} develop a benchmark for inference of system calls of Hyperkernel;
    however, the benchmark does not consider distributed systems, concurrency, and assumes a ground-truth specification.

Our evaluation aims to establish a baseline using simple, straightforward agents
    to reflect the status quo of today's generative AI technologies.
More advanced agents, especially those equipped with domain-specific knowledge
    and specialized techniques such as~\citet{bhatia2024verified,wang2025skeleton}, 
    can be developed to improve the quality of AI-generated models.

\begin{plsreview}
{\bf General AI Benchmarks.}
\bench{} differs from general AI reasoning benchmarks such as MMLU~\citep{hendryckstest2021}, ARC~\citep{clark2018think}, and HELM~\citep{liang2022holistic}.
These benchmarks evaluate generic reasoning, knowledge, and problem-solving capabilities across diverse domains,
    while \bench{} focuses on the specific task of formally modeling large, complex software systems as
    a foundation of formal system verification.
\bench{} also differs from benchmarks targeting AI agent safety such as
    Agent-SafetyBench~\citep{zhang2024agent}. %
It currently targets traditional distributed and concurrent systems that are implemented in system code 
    without neural components.
The formal system modeling tasks evaluated by \bench{} are not covered by existing benchmarks such as EvalScope~\citep{evalscope}.
\end{plsreview}

\section{Concluding Remarks}
\label{sec:conclusion}

This paper presents \bench{}, a new benchmark for evaluating generative AI in formally
    modeling real-world computing systems.
\bench{} pushed us to articulate the criteria
    of formal system models
    and to develop metrics that can be collected automatically.
We find that modern AI, despite showing strong abilities in coding and bug fixing,
    is still limited in comprehending, abstracting, and specifying 
    large, complex systems.
We hope to use \bench{} as a vehicle to advance AI technologies 
    towards software system intelligence, rather than code intelligence.

We are actively adding new system artifacts to \bench{} and improving 
    the benchmark's usability. We encourage others
    to contribute their system artifacts to \bench{}.
We are also exploring ways to measure the maintainability of 
    AI-generated system models and considering ways to include
    human evaluation as part of \bench{}.

\section*{Ethics Statement}

We strictly obeyed the principles outlined in the ICLR Code of Ethics, 
    and carefully examined potential ethical concerns, 
    including potential impacts on human subjects, practices to data set releases, potentially harmful insights, methodologies and applications, potential conflicts of interest and sponsorship, discrimination/bias/fairness concerns, privacy and security issues, legal compliance, and research integrity issues.
We do not identify any potential risks.
In fact, we believe that the work, together with its artifacts (e.g., the \tla{} models) 
    will have positive impacts on the correctness
    of real-world computing systems and infrastructures.

\section*{Reproducibility Statement}

We have made faithful efforts to ensure the reproducibility of our work.
We have provided the details of our work in the paper and its appendix, 
    including the prompts, implementations, and complete results.
We have open-sourced all the research artifacts described in this paper,
    and created an anonymous snapshot at \url{https://anonymous.4open.science/r/SysMoBench-BA9F/}
    for the paper review,
    which documents how to use and extend different parts of the benchmark.
We expect that readers can easily reproduce our results reported in the paper.
We also maintain an active forum to assist with reproduction problems 
    and questions on how to use and build on \bench{}.

\bibliography{main}
\bibliographystyle{iclr2026_conference}

\newpage
\appendix
\section{Alternative Metrics}
\label{appendix:metrics}

No metric is perfect.
Besides the core metrics presented in the paper,
\bench{} also measures complementary metrics that provide 
    different measures of the system model quality.

\subsection{Runtime Pass Rate}

\bench{} repeatedly runs an agent to generate multiple \tla{} system models 
    and evaluates whether each system model passes the runtime checks.
The runtime pass rate is defined as $M_{ar} = \frac{n_{ar}}{n_{at}}$, 
    where $n_{ar}$ is the number of \tla{} models that passed runtime checking, 
    and $n_{at}$ is the total number of generated \tla{} models.
This metric complements the system model's action-level coverage metric $M_r$ (see \S\ref{sec:error}), 
    as it reflects the agent's ability and reliability to produce fully executable \tla{} models.
Note that a high $M_{ar}$ does not necessarily mean most actions in the \tla{} 
    models are correct.
Even if some actions may contain runtime errors but are never executed during execution, 
    the \tla{} model can still pass runtime checking.
Conversely, a low $M_{ar}$ may result from a small number of frequently failing actions 
    rather than errors affecting many actions.

\subsection{Conformance Pass Rate}

\bench{} repeatedly executes the system code to generate multiple code traces 
    and checks which traces fully pass conformance checking.
The conformance pass rate is defined as $M_{ac} = \frac{n_{ac}}{n_{at}}$, 
    where $n_{ac}$ is the number of traces that passed conformance checking, 
    and $n_{at}$ is the total number of traces generated.
This metric complements the code action-level coverage metric $M_c$ (see \S\ref{sec:conformance})
    and provides a coarse-grained empirical measure of the \tla{} model's 
    overall alignment with observed system behavior.
As with runtime correctness, a low $M_{ac}$ does not necessarily indicate that most actions are unconformed,
while a high $M_{ac}$ generally suggests better overall system model quality, 
    given sufficiently diverse traces.

\begin{plsreview}
\section{Extensibility to Other Specification Languages}
\label{appendix:extensibility}

\bench{} is general to specification languages beyond \tla{}.
To demonstrate its extensibility, we extended \bench{} to support Alloy~\citep{jackson2012alloybook} and PAT~\citep{sun2009pat}.

\subsection{Supporting PAT and Alloy}

{\bf PAT.}
PAT (Process Analysis Toolkit) is a formal verification framework for concurrent and real-time systems.
Supporting PAT in \bench{} is straightforward because PAT's tooling provides a workflow similar to \tla{}.
We leverage PAT's parser for syntax checking, its simulator for runtime evaluation, and its assertion mechanism with model checking for invariant validation.
Conformance is evaluated using PAT's native trace refinement checker.
We implement adaptors to translate our concrete system traces into the PAT format, which are then validated against the PAT models.

{\bf Alloy.}
Alloy is a declarative specification language based on first-order relational logic.
For Alloy support, evaluating syntax, runtime, and invariant correctness is straightforward using the Alloy Analyzer tool.
Since Alloy does not provide a built-in notion of ``action'' as in \tla{}, we adapt the runtime metric by computing the proportion of variables and fields that become instantiated during bounded execution. 
This metric is analogous to the action-trigger coverage in \tla{}, and it indicates whether a model executes normally and whether certain branches are unreachable. 
For the conformance metric, we express a concrete system trace into Alloy facts, which are global constraints over a bounded sequence of states and must hold in all generated instances, for trace validation.

\subsection{Evaluation Results}
\label{appendix:extensibility-results}

We evaluated the basic modeling agent with four LLMs (Claude-Sonnet-4, GPT-5, Gemini-2.5-Pro, and DeepSeek-R1) on generating Alloy and PAT models for the Spinlock system, with three attempts per LLM.
Table~\ref{tab:extensibility-results} shows the results.
For both PAT and Alloy, the four evaluation metrics (syntax, runtime, conformance, and invariant) remain applicable.
However, due to limitations of current tools, syntax checking for PAT and Alloy does not yet support partial scoring as in \tla{}.

\begin{table}[h]
\centering
\small
\caption{\plsreviewinline{Preliminary results of Alloy and PAT support on Asterinas Spinlock using the basic modeling agent (3 attempts per LLM).}}
\label{tab:extensibility-results}
\plsreviewtable{}
\begin{tabular}{llcccc}
\toprule
\textbf{Language} & \textbf{LLM} & \textbf{Syntax} & \textbf{Runtime} & \textbf{Conformance} & \textbf{Invariant} \\
\midrule
\multirow{4}{*}{Alloy}
& Claude-Sonnet-4 & 0.00\% & 0.00\% & 0.00\% & 0.00\% \\
& GPT-5 & 100.00\% & 0.00\% & 0.00\% & 0.00\% \\
& Gemini-2.5-Pro & 0.00\% & 0.00\% & 0.00\% & 0.00\% \\
& DeepSeek-R1 & 0.00\% & 0.00\% & 0.00\% & 0.00\% \\
\midrule
\multirow{4}{*}{PAT}
& Claude-Sonnet-4 & 0.00\% & 0.00\% & 0.00\% & 0.00\% \\
& GPT-5 & 0.00\% & 0.00\% & 0.00\% & 0.00\% \\
& Gemini-2.5-Pro & 0.00\% & 0.00\% & 0.00\% & 0.00\% \\
& DeepSeek-R1 & 0.00\% & 0.00\% & 0.00\% & 0.00\% \\
\bottomrule
\end{tabular}
\end{table}

The AI-generated Alloy and PAT models are poor compared to \tla{} models.
For Alloy, only GPT-5 was able to generate a model that passes the syntax correctness check after multiple attempts, but the generated model scored 0\% on runtime correctness. %
For PAT, none of the evaluated LLMs demonstrated familiarity with the PAT syntax---all generated PAT models failed syntax checks.

Our analysis reveals that current LLMs are unfamiliar with the syntax of Alloy and PAT.
In practice, nearly all generated models failed at the parsing or type-checking stage.
For PAT, LLMs frequently produced syntax borrowed from other languages such as C, Promela, or PRISM. 
For example, channels were often declared using PRISM-style range expressions (e.g., \texttt{channel acquire:\{0..2\};}) which PAT does not support. 
We also observed the introduction of keywords and type annotations that do not exist in PAT, such as adding explicit types (\texttt{int}) after variables or using \texttt{chan} instead of PAT's actual channel declaration syntax.
For Alloy, we observed similarly systematic breakdowns.
A common pattern was referencing signatures (types) that were never declared in the model, such as using \texttt{Time} in module imports without defining what \texttt{Time} is.
The models also wrote constraints that mixed incompatible language features, which Alloy does not accept.

We believe that the weak model capabilities using PAT and Alloy are primarily because Alloy and PAT are much less popular than \tla{} in real-world systems.
Consequently, LLMs are not extensively trained on these languages, resulting in poor generation quality.
These results justify the use of \tla{} as the specification language of choice for \bench{}.

\subsection{Comparison with Related Work}
\label{appendix:comparison}

PAT-Agent~\citep{zuo2025patagent} and Alloy-APR~\citep{alhanahnah2025empirical} also evaluate AI's ability to work with formal models using PAT and Alloy, reporting promising results on their benchmarks.
However, their tasks and complexity differ fundamentally from \bench{}.
Table~\ref{tab:task-summary} summarizes the task of each work.

\begin{table}[h]
\centering
\small
\caption{\plsreviewinline{Summary of the tasks of each work.}}
\plsreviewtable{}
\label{tab:task-summary}
\begin{tabular}{ll}
\toprule
\textbf{Work} & \textbf{Task} \\
\midrule
\bench{} &
Generating formal models for real-world software systems from their source code. \\
PAT-Agent &
Generating formal models from natural language descriptions. \\
Alloy-APR &
Repairing an existing model with injected errors. \\
\bottomrule
\end{tabular}
\end{table}

Because these tasks are inherently different, it is difficult to compare their complexity directly. Instead, we compare the complexity of the generated formal models as shown in Table~\ref{tab:benchmark-complexity}.

\begin{table}[ht]
\centering
\caption{\plsreviewinline{Complexity comparison across benchmarks measured by lines of code of formal models.}}
\label{tab:benchmark-complexity}
\plsreviewtable{}
\small
\begin{tabular}{lrrrl}
\toprule
\textbf{Benchmark} & \textbf{Smallest} & \textbf{Largest} & \textbf{Median} & \textbf{Task Type} \\
\midrule
\bench{} & 75 & 508 & 219 & Generation \\
PAT-Agent & 16 & 142 & 45 & Generation \\
Alloy-APR (ARepair) & 15 & 99 & 50 & Repair \\
Alloy-APR (Alloy4Fun) & 1 & 234 & 21 & Repair \\
\bottomrule
\end{tabular}
\end{table}

Compared with the generation task in PAT-Agent, most models we expect LLMs/agents to generate in \bench{} are larger than the largest models in the PAT-Agent paper.
PAT-Agent's tasks are small samples such as river-crossing puzzles and restaurant workflows, not real-world software systems.
Alloy-APR's tasks are similar, which come from ARepair and Alloy4Fun; neither of them uses real-world system artifacts.

To further validate our understanding, we reproduced the results of Alloy-APR and PAT-Agent using the same LLMs evaluated in \bench{}.
For Alloy-APR, we used the official artifact on the ARepair benchmark.
Table~\ref{tab:alloy-apr-reproduction} shows the results.

\begin{table}[h]
\centering
\caption{\plsreviewinline{Reproduction of Alloy-APR results on ARepair benchmark with LLMs used in \bench{}.}}
\label{tab:alloy-apr-reproduction}
\plsreviewtable{}
\small
\begin{tabular}{lrr}
\toprule
\textbf{Model} & \textbf{Correct Items} & \textbf{Success Rate} \\
\midrule
Claude-Sonnet-4 & 38 / 38 & 100.0\% \\
GPT-5 & 30 / 38 & 78.9\% \\
Gemini-2.5-Pro & 14 / 38 & 36.8\% \\
DeepSeek-R1 & 5 / 38 & 13.2\% \\
\midrule
Best result in Alloy-APR & 28 / 38 & 73.7\% \\
\bottomrule
\end{tabular}
\end{table}

Our reproduction results show that Claude-Sonnet-4 and GPT-5 outperform the best results reported in the Alloy-APR paper.
This suggests that existing LLMs can solve these repair tasks effectively--the high scores in the paper are largely due to the fact that the task itself is relatively simple.
In contrast, our results show that these LLMs still struggle to generate syntax-correct Alloy models from complex system code in \bench{} (see Table~\ref{tab:extensibility-results}).

For PAT-Agent, we ran the NoPlanning workflow using the LLMs evaluated in \bench{}.
This workflow is similar to our Basic Modeling Agent: it calls the LLM to generate a PAT model and then iteratively fixes errors.
Table~\ref{tab:pat-agent-reproduction} shows the results.

\begin{table}[h]
\centering
\caption{\plsreviewinline{Reproduction of PAT-Agent results using NoPlanning workflow with LLMs from \bench{}. CSR: Compilation Success Rate, FPR: Full Pass Rate, APR: Average Pass Rate.}}
\label{tab:pat-agent-reproduction}
\plsreviewtable{}
\small
\begin{tabular}{lrrr}
\toprule
\textbf{Model} & \textbf{CSR} & \textbf{FPR} & \textbf{APR} \\
\midrule
Claude-Sonnet-4 & 84.6\% & 80.8\% & 87.3\% \\
GPT-5 & 84.6\% & 69.2\% & 76.4\% \\
Gemini-2.5-Pro & 84.6\% & 65.4\% & 74.8\% \\
DeepSeek-R1 & 57.7\% & 50.0\% & 54.9\% \\
\bottomrule
\end{tabular}
\vspace{-10pt}
\end{table}

The results are consistent with the original paper's findings.
Similar to Alloy-APR, current LLMs can solve these relatively simple tasks to a reasonable extent (e.g., Claude-Sonnet-4 achieves 87.3\% APR).
However, their ability to generate formal models for real-world software systems is much weaker, as evidenced by our results (see Table~\ref{tab:extensibility-results}).

These results suggest that existing benchmarks such as PAT-Agent and Alloy-APR mostly exercise simplified modeling tasks.
In contrast, \bench{} targets formal models derived from real system code, where current LLMs often fail to produce even syntax-correct specifications (see Table~\ref{tab:extensibility-results}).

\end{plsreview}

\begin{plsreview}
\section{Qualitative Evaluation of AI-Generated Models}
\label{appendix:qualitative}

Beyond the automated quantitative metrics, we performed qualitative evaluation to assess 
    the practical utility of AI-generated models for human engineers.
We evaluated AI-generated models in two aspects: 
    (1) comparison with human-written ground-truth models from the community, and 
    (2) their ability to reproduce known bugs in system code.

\subsection{Comparison with Human-Written Models}

To assess the quality of AI-generated models, 
    two human experts evaluated models produced by two different agents:
    the basic modeling agent and the code translation agent.
Each expert compared AI-generated \tla{} models against human-written models for nine of the systems in \bench{}.
The experts identified ten main types of differences between AI-generated and human models
    (Tables~\ref{tab:human-comparison} and~\ref{tab:human-comparison-translation} summarize their occurrence across systems and LLMs):

\begin{packed_enumerate}
    \item Unnecessary \texttt{EXTENDS}/\texttt{INSTANCE} statements
    \item Topics present in the human model but missing in AI models
    \item Topics introduced by AI but absent in the human model
    \item Fewer comments compared to human models
    \item Properties present in human models but not in AI models
    \item Different fairness assumptions compared to human models
    \item Longer composite actions in AI models
    \item Overly complex or random fairness conditions
    \item Overspecialization with hard-coded values instead of parameters
\end{packed_enumerate}

\begin{table}[h]
\caption{\plsreviewinline{Types of differences between AI-generated models (produced by the basic modeling agent) and human-written models. Numbers refer to the types listed above.}}
\label{tab:human-comparison}
\centering
\plsreviewtable{}
\small
\begin{tabular}{lcccc}
\toprule
\textbf{System} & \textbf{Claude-Sonnet-4} & \textbf{GPT-5} & \textbf{Gemini-2.5-Pro} & \textbf{DeepSeek-R1} \\
\midrule
Asterinas Spin & 1, 3 & 1, 5, 8 & 1, 5, 8 & 1, 5, 8 \\
Asterinas Mutex & 1, 5 & 1, 5 & 1, 5, 9 & 5 \\
Asterinas Rwmutex & 1, 5 & 1, 5 & 1, 5 & 1, 5 \\
Etcd Raft & 1, 2, 4, 5, 7 & 1, 2, 4, 5 & 1, 2, 4, 5, 7 & 1, 2, 4, 5, 7 \\
Redis Raft & 1, 4 & 1, 4, 7 & 1, 4, 7 & 1, 4 \\
Xline CURP & 1, 2, 4, 5, 6 & 1, 2, 3, 4, 5, 6 & 1, 2, 4, 5, 6 & 1, 2, 4, 5, 6 \\
PGo dqueue & 1, 5 & 1, 5 & 1, 5 & 1, 5, 7 \\
PGo locksvc & 1, 5 & 1, 5 & 1, 5 & 5, 6, 7, 8 \\
PGo raftkvs & 1, 5, 7 & 1, 5, 7, 8 & 1, 5, 8 & 1, 5 \\
\bottomrule
\end{tabular}
\end{table}

\begin{table}[h]
\caption{\plsreviewinline{Types of differences between AI-generated models (produced by the code translation agent) and human-written models. Numbers refer to the types listed above.}}
\label{tab:human-comparison-translation}
\centering
\small
\plsreviewtable{}
\begin{tabular}{lcccc}
\toprule
\textbf{System} & \textbf{Claude-Sonnet-4} & \textbf{GPT-5} & \textbf{Gemini-2.5-Pro} & \textbf{DeepSeek-R1} \\
\midrule
Asterinas Spin & 1, 2, 3, 5, 6 & 1, 3, 5, 6, 8 & 1, 2, 3, 5, 6 & 1, 2, 3, 5, 6, 8 \\
Asterinas Mutex & 1, 2, 5, 8 & 1, 2, 5 & 1, 2, 5 & 1, 2, 5, 8 \\
Asterinas Rwmutex & 1, 5, 6 & 1, 3, 5, 6, 8 & 1, 3, 5, 6 & 1, 5, 6, 8 \\
Etcd Raft & 1, 2, 4, 5, 6, 8 & 1, 2, 4, 5, 6 & 1, 2, 4, 5, 6, 7 & 1, 2, 4, 5, 6, 7 \\
Redis Raft & 1, 3, 4, 5 & 1, 3, 4, 5 & 1, 3, 4, 5 & 1, 2, 4, 5, 6 \\
Xline CURP & 1, 4, 5, 6 & 4, 5, 6 & 1, 2, 4, 5, 6 & 1, 4, 5, 6 \\
PGo dqueue & 1, 2, 3, 5, 6 & 1, 5, 6 & 1, 5, 6 & 1, 3, 5, 6 \\
PGo locksvc & 1, 5 & 1, 5, 8 & 1, 5 & 1, 5, 6, 8 \\
PGo raftkvs & 1, 5, 7 & 1, 5, 7, 8 & 1, 5, 6, 7, 8 & 1, 5, 6, 7 \\
\bottomrule
\end{tabular}
\end{table}

We group and discuss these differences below.

{\bf Prompt-induced patterns (types 1, 4, 5).}
For both agents, many AI models include unnecessary \texttt{EXTENDS} / \texttt{INSTANCE} statements (type 1), 
    lack comments (type 4), and omit certain properties (type 5).
These patterns largely result from our prompting and evaluation design.
The prompt requires including common libraries to avoid syntax errors; 
    this does not harm correctness or the evaluation of AI's modeling capability, 
    as human experts also sometimes copy-paste \texttt{EXTENDS} with unnecessary dependencies.
Missing comments and properties are expected, 
    as \bench{} focuses on state/action modeling and does not require comment or property generation.

{\bf Model utility (types 2, 3, 6, 8, 9).}
AI-generated models may miss certain variables or actions (type 2) or include extra details (type 3),
    especially when there is a significant difference in abstraction levels between the human-written models and AI-generated models.
For instance, the code translation agent tends to produce more concrete specifications compared to human-written ones,
    leading to more frequent occurrences of type 2 (missing topics) and type 3 (extra topics).
Fairness definitions (types 6 and 8) of AI-generated models often differ from human models or are overly technical or random, 
    which can affect liveness checking.
There are also isolated cases of overspecialization (type 9).
Overall, these differences show that AI models capture the general structure 
    but may vary in completeness, fairness, and abstraction compared to human models.

{\bf Readability and documentation (types 4, 7).}
For both agents, the issue of fewer comments (type 4) is due to our prompt design; 
    when we removed the instruction not to generate comments, 
    AI models produced reasonably long comments that are easy to read.
Some models also contain long composite actions (type 7) or use unconventional ordering of structure 
    (e.g., \texttt{TypeOK} checks placed unusually).
Nevertheless, AI-generated models from both agents remain generally readable, 
    with meaningful variable and action names and understandable structure.

\subsection{Bug Reproduction}
\label{appendix:bug-reproduction}

AI-generated models can be practically useful for partial correctness checking.
Without any hints about specific bugs, AI-generated models successfully reproduced several hard-to-find bugs 
    across multiple systems in \bench{}.
Table~\ref{tab:bug-reproduction} lists these reproduced bugs with links to the corresponding issue reports or pull requests.

\begin{table}[h]
\caption{\plsreviewinline{Bugs successfully reproduced by AI-generated models.}}
\label{tab:bug-reproduction}
\centering
\plsreviewtable{}
\footnotesize
\begin{tabular}{ll}
\toprule
\textbf{Bug ID} & \textbf{Description} \\
\midrule
\buglink{https://github.com/etcd-io/etcd/pull/10998}{Etcd Raft \#10998} & Learners cannot vote during promotion causing election failure \\
\buglink{https://github.com/xline-kv/Xline/issues/402}{Xline CURP \#402} & Cluster will enter a frozen state after multiple crashes and recoveries \\
\buglink{https://github.com/RedisLabs/redisraft/issues/19}{Redis Raft \#19} & Stale reads under process pauses, violating linearizability \\
\buglink{https://github.com/asterinas/asterinas/pull/1279}{Asterinas Mutex \#1279} & Failed \code{try\_lock} incorrectly unlocks mutex breaking mutual exclusion \\
\buglink{https://github.com/asterinas/asterinas/issues/1303}{Asterinas Rwmutex \#1303} & Lost wakeup when upgradeable reader releases lock \\
\bottomrule
\end{tabular}
\end{table}

These bugs were discovered by having AI agents generate \tla{} models from earlier versions of the system code
    and then using model checking to identify the issues.
\end{plsreview}

\section{Examples of AI-generated \tla{} Models}
\label{appendix:examples}

We present two AI-generated system models.
Figures~\ref{fig:app-spin-claude} and~\ref{fig:app-spin-claude-cfg}
    show the \tla{} model and its corresponding TLC configuration
    for Asterinas Spinlock generated by the basic modeling agent with Claude-Sonnet-4.
Figures~\ref{fig:app-etcd-specula} and~\ref{fig:app-etcd-specula-cfg}
    show the \tla{} model and its corresponding TLC configuration
    for Etcd Raft generated by the code translation agent with Claude-Sonnet-4.

\vspace{10pt}

\begin{longlisting}
    \begin{minted}[escapeinside=\%\%,breaklines,breakanywhere,breakautoindent]{text}
%\PYG{o}{----}% %\PYG{k}{MODULE}% spin %\PYG{o}{----}%

%\PYG{k}{EXTENDS}% TLC, Sequences, SequencesExt, Naturals, FiniteSets, Bags

%\PYG{k}{CONSTANTS}% Threads

%\PYG{k}{VARIABLES}%
    lock_state,
    thread_state,
    guards

Vars %\PYG{o}{==}% %\PYG{o}{<}%%\PYG{o}{<}%lock_state, thread_state, guards%\PYG{o}{>}%%\PYG{o}{>}%

TypeOK %\PYG{o}{==}%
    %\PYG{o}{/\symbol{92}}% lock_state %\PYG{o}{\symbol{92}in}% BOOLEAN
    %\PYG{o}{/\symbol{92}}% thread_state %\PYG{o}{\symbol{92}in}% %\PYG{o}{[}%Threads %\PYG{o}{->}% %\PYG{o}{\{}%%\PYG{l+s}{"idle"}%, %\PYG{l+s}{"spinning"}%, %\PYG{l+s}{"locked"}%%\PYG{o}{\}}%%\PYG{o}{]}%
    %\PYG{o}{/\symbol{92}}% guards %\PYG{o}{\symbol{92}subset}%eq Threads

Init %\PYG{o}{==}%
    %\PYG{o}{/\symbol{92}}% lock_state %\PYG{o}{=}% %\PYG{k}{FALSE}%
    %\PYG{o}{/\symbol{92}}% thread_state %\PYG{o}{=}% %\PYG{o}{[}%t %\PYG{o}{\symbol{92}in}% Threads %\PYG{o}{|->}% %\PYG{l+s}{"idle"}%%\PYG{o}{]}%
    %\PYG{o}{/\symbol{92}}% guards %\PYG{o}{=}% %\PYG{o}{\{}%%\PYG{o}{\}}%

%\PYG{n+nf}{TryAcquire}%(t) %\PYG{o}{==}%
    %\PYG{o}{/\symbol{92}}% thread_state%\PYG{o}{[}%t%\PYG{o}{]}% %\PYG{o}{\symbol{92}in}% %\PYG{o}{\{}%%\PYG{l+s}{"idle"}%, %\PYG{l+s}{"spinning"}%%\PYG{o}{\}}%
    %\PYG{o}{/\symbol{92}}% lock_state %\PYG{o}{=}% %\PYG{k}{FALSE}%
    %\PYG{o}{/\symbol{92}}% lock_state%\PYG{o}{'}% %\PYG{o}{=}% %\PYG{k}{TRUE}%
    %\PYG{o}{/\symbol{92}}% thread_state%\PYG{o}{'}% %\PYG{o}{=}% %\PYG{o}{[}%thread_state %\PYG{k}{EXCEPT}% %\PYG{o}{![}%t%\PYG{o}{]}% %\PYG{o}{=}% %\PYG{l+s}{"locked"}%%\PYG{o}{]}%
    %\PYG{o}{/\symbol{92}}% guards%\PYG{o}{'}% %\PYG{o}{=}% guards %\PYG{o}{\symbol{92}cup}% %\PYG{o}{\{}%t%\PYG{o}{\}}%

%\PYG{n+nf}{StartSpin}%(t) %\PYG{o}{==}%
    %\PYG{o}{/\symbol{92}}% thread_state%\PYG{o}{[}%t%\PYG{o}{]}% %\PYG{o}{=}% %\PYG{l+s}{"idle"}%
    %\PYG{o}{/\symbol{92}}% lock_state %\PYG{o}{=}% %\PYG{k}{TRUE}%
    %\PYG{o}{/\symbol{92}}% thread_state%\PYG{o}{'}% %\PYG{o}{=}% %\PYG{o}{[}%thread_state %\PYG{k}{EXCEPT}% %\PYG{o}{![}%t%\PYG{o}{]}% %\PYG{o}{=}% %\PYG{l+s}{"spinning"}%%\PYG{o}{]}%
    %\PYG{o}{/\symbol{92}}% %\PYG{k}{UNCHANGED}% %\PYG{o}{<}%%\PYG{o}{<}%lock_state, guards%\PYG{o}{>}%%\PYG{o}{>}%

%\PYG{n+nf}{SpinLoop}%(t) %\PYG{o}{==}%
    %\PYG{o}{/\symbol{92}}% thread_state%\PYG{o}{[}%t%\PYG{o}{]}% %\PYG{o}{=}% %\PYG{l+s}{"spinning"}%
    %\PYG{o}{/\symbol{92}}% lock_state %\PYG{o}{=}% %\PYG{k}{TRUE}%
    %\PYG{o}{/\symbol{92}}% %\PYG{k}{UNCHANGED}% %\PYG{o}{<}%%\PYG{o}{<}%lock_state, thread_state, guards%\PYG{o}{>}%%\PYG{o}{>}%

%\PYG{n+nf}{SpinAcquire}%(t) %\PYG{o}{==}%
    %\PYG{o}{/\symbol{92}}% thread_state%\PYG{o}{[}%t%\PYG{o}{]}% %\PYG{o}{=}% %\PYG{l+s}{"spinning"}%
    %\PYG{o}{/\symbol{92}}% lock_state %\PYG{o}{=}% %\PYG{k}{FALSE}%
    %\PYG{o}{/\symbol{92}}% lock_state%\PYG{o}{'}% %\PYG{o}{=}% %\PYG{k}{TRUE}%
    %\PYG{o}{/\symbol{92}}% thread_state%\PYG{o}{'}% %\PYG{o}{=}% %\PYG{o}{[}%thread_state %\PYG{k}{EXCEPT}% %\PYG{o}{![}%t%\PYG{o}{]}% %\PYG{o}{=}% %\PYG{l+s}{"locked"}%%\PYG{o}{]}%
    %\PYG{o}{/\symbol{92}}% guards%\PYG{o}{'}% %\PYG{o}{=}% guards %\PYG{o}{\symbol{92}cup}% %\PYG{o}{\{}%t%\PYG{o}{\}}%

%\PYG{n+nf}{TryLock}%(t) %\PYG{o}{==}%
    %\PYG{o}{/\symbol{92}}% thread_state%\PYG{o}{[}%t%\PYG{o}{]}% %\PYG{o}{=}% %\PYG{l+s}{"idle"}%
    %\PYG{o}{/\symbol{92}}% %\PYG{k}{IF}% lock_state %\PYG{o}{=}% %\PYG{k}{FALSE}%
       %\PYG{k}{THEN}% %\PYG{o}{/\symbol{92}}% lock_state%\PYG{o}{'}% %\PYG{o}{=}% %\PYG{k}{TRUE}%
            %\PYG{o}{/\symbol{92}}% thread_state%\PYG{o}{'}% %\PYG{o}{=}% %\PYG{o}{[}%thread_state %\PYG{k}{EXCEPT}% %\PYG{o}{![}%t%\PYG{o}{]}% %\PYG{o}{=}% %\PYG{l+s}{"locked"}%%\PYG{o}{]}%
            %\PYG{o}{/\symbol{92}}% guards%\PYG{o}{'}% %\PYG{o}{=}% guards %\PYG{o}{\symbol{92}cup}% %\PYG{o}{\{}%t%\PYG{o}{\}}%
       %\PYG{k}{ELSE}% %\PYG{o}{/\symbol{92}}% %\PYG{k}{UNCHANGED}% %\PYG{o}{<}%%\PYG{o}{<}%lock_state, thread_state, guards%\PYG{o}{>}%%\PYG{o}{>}%

%\PYG{n+nf}{Unlock}%(t) %\PYG{o}{==}%
    %\PYG{o}{/\symbol{92}}% thread_state%\PYG{o}{[}%t%\PYG{o}{]}% %\PYG{o}{=}% %\PYG{l+s}{"locked"}%
    %\PYG{o}{/\symbol{92}}% t %\PYG{o}{\symbol{92}in}% guards
    %\PYG{o}{/\symbol{92}}% lock_state%\PYG{o}{'}% %\PYG{o}{=}% %\PYG{k}{FALSE}%
    %\PYG{o}{/\symbol{92}}% thread_state%\PYG{o}{'}% %\PYG{o}{=}% %\PYG{o}{[}%thread_state %\PYG{k}{EXCEPT}% %\PYG{o}{![}%t%\PYG{o}{]}% %\PYG{o}{=}% %\PYG{l+s}{"idle"}%%\PYG{o}{]}%
    %\PYG{o}{/\symbol{92}}% guards%\PYG{o}{'}% %\PYG{o}{=}% guards %\PYG{o}{\symbol{92} }%%\PYG{o}{\{}%t%\PYG{o}{\}}%

Next %\PYG{o}{==}%
    %\PYG{o}{\symbol{92}E}% t %\PYG{o}{\symbol{92}in}% Threads:
        %\PYG{o}{\symbol{92}/}% %\PYG{n+nf}{TryAcquire}%(t)
        %\PYG{o}{\symbol{92}/}% %\PYG{n+nf}{StartSpin}%(t)
        %\PYG{o}{\symbol{92}/}% %\PYG{n+nf}{SpinLoop}%(t)
        %\PYG{o}{\symbol{92}/}% %\PYG{n+nf}{SpinAcquire}%(t)
        %\PYG{o}{\symbol{92}/}% %\PYG{n+nf}{TryLock}%(t)
        %\PYG{o}{\symbol{92}/}% %\PYG{n+nf}{Unlock}%(t)

Spec %\PYG{o}{==}% Init %\PYG{o}{/\symbol{92}}% %\PYG{o}{[}%%\PYG{o}{]}%%\PYG{o}{[}%Next%\PYG{o}{]}%_Vars %\PYG{o}{/\symbol{92}}% %\PYG{o}{\symbol{92}A}% t %\PYG{o}{\symbol{92}in}% Threads: %\PYG{n+nf}{WF\_Vars}%(%\PYG{n+nf}{Unlock}%(t))

%\PYG{o}{==}%%\PYG{o}{==}%
\end{minted}

    \caption{Asterinas Spinlock \tla{} model generated by the basic modeling agent with Claude-Sonnet-4.}
    \vspace{15pt}
    \label{fig:app-spin-claude}
\end{longlisting}

\begin{longlisting}
    \begin{minted}[escapeinside=\%\%,breaklines,breakanywhere,breakautoindent]{text}
%\PYG{k}{SPECIFICATION}% Spec

%\PYG{k}{CONSTANTS}%
    Threads %\PYG{o}{=}% %\PYG{o}{\{}%t1, t2, t3%\PYG{o}{\}}%
\end{minted}

    \caption{Asterinas Spinlock TLC configuration generated by the basic modeling agent with Claude-Sonnet-4.}
    \vspace{15pt}
    \label{fig:app-spin-claude-cfg}
\end{longlisting}

\begin{longlisting}
    \begin{minted}[escapeinside=\%\%,breaklines,breakanywhere,breakautoindent]{text}
%\PYG{o}{----}% %\PYG{k}{MODULE}% etcdraft %\PYG{o}{----}%
%\PYG{k}{EXTENDS}% TLC, Sequences, SequencesExt, Naturals, FiniteSets, Bags

%\PYG{k}{CONSTANTS}%
    Server,          %\PYG{c+c1}{\symbol{92}* Set of server IDs}%
    Value,           %\PYG{c+c1}{\symbol{92}* Set of possible values for log entries}%
    Nil,             %\PYG{c+c1}{\symbol{92}* Constant representing no value/no leader}%
    Follower,        %\PYG{c+c1}{\symbol{92}* Node state constants}%
    PreCandidate,
    Candidate,
    Leader

%\PYG{k}{VARIABLES}%
    state,           %\PYG{c+c1}{\symbol{92}* state[i] is the state of server i}%
    currentTerm,     %\PYG{c+c1}{\symbol{92}* currentTerm[i] is the current term of server i}%
    votedFor,        %\PYG{c+c1}{\symbol{92}* votedFor[i] is who server i voted for in currentTerm[i]}%
    log,             %\PYG{c+c1}{\symbol{92}* log[i] is the log of server i}%
    commitIndex,     %\PYG{c+c1}{\symbol{92}* commitIndex[i] is the commit index of server i}%
    nextIndex,       %\PYG{c+c1}{\symbol{92}* nextIndex[i][j] is the next log index to send to server j from leader i}%
    matchIndex,      %\PYG{c+c1}{\symbol{92}* matchIndex[i][j] is the highest log index known to be replicated on server j by leader i}%
    messages,        %\PYG{c+c1}{\symbol{92}* Set of messages in transit}%
    electionTimeout, %\PYG{c+c1}{\symbol{92}* electionTimeout[i] tracks election timeout for server i}%
    leader           %\PYG{c+c1}{\symbol{92}* leader[i] is the current leader known to server i}%

vars %\PYG{o}{==}% %\PYG{o}{<}%%\PYG{o}{<}%state, currentTerm, votedFor, log, commitIndex, nextIndex, matchIndex, messages, electionTimeout, leader%\PYG{o}{>}%%\PYG{o}{>}%

%\PYG{c+c1}{\symbol{92}* Message types}%
MsgHup %\PYG{o}{==}% %\PYG{l+s}{"MsgHup"}%
MsgVote %\PYG{o}{==}% %\PYG{l+s}{"MsgVote"}%
MsgVoteResp %\PYG{o}{==}% %\PYG{l+s}{"MsgVoteResp"}%
MsgPreVote %\PYG{o}{==}% %\PYG{l+s}{"MsgPreVote"}%
MsgPreVoteResp %\PYG{o}{==}% %\PYG{l+s}{"MsgPreVoteResp"}%
MsgApp %\PYG{o}{==}% %\PYG{l+s}{"MsgApp"}%
MsgAppResp %\PYG{o}{==}% %\PYG{l+s}{"MsgAppResp"}%
MsgHeartbeat %\PYG{o}{==}% %\PYG{l+s}{"MsgHeartbeat"}%
MsgProp %\PYG{o}{==}% %\PYG{l+s}{"MsgProp"}%

%\PYG{c+c1}{\symbol{92}* Helper functions}%
%\PYG{n+nf}{Min}%(a, b) %\PYG{o}{==}% %\PYG{k}{IF}% a %\PYG{o}{<}% b %\PYG{k}{THEN}% a %\PYG{k}{ELSE}% b
%\PYG{n+nf}{Max}%(a, b) %\PYG{o}{==}% %\PYG{k}{IF}% a %\PYG{o}{>}% b %\PYG{k}{THEN}% a %\PYG{k}{ELSE}% b

%\PYG{n+nf}{LastTerm}%(xlog) %\PYG{o}{==}% %\PYG{k}{IF}% %\PYG{n+nf}{Len}%(xlog) %\PYG{o}{=}% 0 %\PYG{k}{THEN}% 0 %\PYG{k}{ELSE}% xlog%\PYG{o}{[}%%\PYG{n+nf}{Len}%(xlog)%\PYG{o}{]}%.term

%\PYG{n+nf}{Send}%(m) %\PYG{o}{==}% messages%\PYG{o}{'}% %\PYG{o}{=}% messages %\PYG{o}{\symbol{92}cup}% %\PYG{o}{\{}%m%\PYG{o}{\}}%

%\PYG{c+c1}{\symbol{92}* Initial state}%
Init %\PYG{o}{==}%
    %\PYG{o}{/\symbol{92}}% state %\PYG{o}{=}% %\PYG{o}{[}%i %\PYG{o}{\symbol{92}in}% Server %\PYG{o}{|->}% Follower%\PYG{o}{]}%
    %\PYG{o}{/\symbol{92}}% currentTerm %\PYG{o}{=}% %\PYG{o}{[}%i %\PYG{o}{\symbol{92}in}% Server %\PYG{o}{|->}% 0%\PYG{o}{]}%
    %\PYG{o}{/\symbol{92}}% votedFor %\PYG{o}{=}% %\PYG{o}{[}%i %\PYG{o}{\symbol{92}in}% Server %\PYG{o}{|->}% Nil%\PYG{o}{]}%
    %\PYG{o}{/\symbol{92}}% log %\PYG{o}{=}% %\PYG{o}{[}%i %\PYG{o}{\symbol{92}in}% Server %\PYG{o}{|->}% %\PYG{o}{<}%%\PYG{o}{<}%%\PYG{o}{>}%%\PYG{o}{>}%%\PYG{o}{]}%
    %\PYG{o}{/\symbol{92}}% commitIndex %\PYG{o}{=}% %\PYG{o}{[}%i %\PYG{o}{\symbol{92}in}% Server %\PYG{o}{|->}% 0%\PYG{o}{]}%
    %\PYG{o}{/\symbol{92}}% nextIndex %\PYG{o}{=}% %\PYG{o}{[}%i %\PYG{o}{\symbol{92}in}% Server %\PYG{o}{|->}% %\PYG{o}{[}%j %\PYG{o}{\symbol{92}in}% Server %\PYG{o}{|->}% 1%\PYG{o}{]}%%\PYG{o}{]}%
    %\PYG{o}{/\symbol{92}}% matchIndex %\PYG{o}{=}% %\PYG{o}{[}%i %\PYG{o}{\symbol{92}in}% Server %\PYG{o}{|->}% %\PYG{o}{[}%j %\PYG{o}{\symbol{92}in}% Server %\PYG{o}{|->}% 0%\PYG{o}{]}%%\PYG{o}{]}%
    %\PYG{o}{/\symbol{92}}% messages %\PYG{o}{=}% %\PYG{o}{\{}%%\PYG{o}{\}}%
    %\PYG{o}{/\symbol{92}}% electionTimeout %\PYG{o}{=}% %\PYG{o}{[}%i %\PYG{o}{\symbol{92}in}% Server %\PYG{o}{|->}% 0%\PYG{o}{]}%
    %\PYG{o}{/\symbol{92}}% leader %\PYG{o}{=}% %\PYG{o}{[}%i %\PYG{o}{\symbol{92}in}% Server %\PYG{o}{|->}% Nil%\PYG{o}{]}%

%\PYG{c+c1}{\symbol{92}* Election timeout - triggers election}%
%\PYG{n+nf}{Timeout}%(i) %\PYG{o}{==}%
    %\PYG{o}{/\symbol{92}}% state%\PYG{o}{[}%i%\PYG{o}{]}% %\PYG{o}{\symbol{92}in}% %\PYG{o}{\{}%Follower, PreCandidate, Candidate%\PYG{o}{\}}%
    %\PYG{o}{/\symbol{92}}% electionTimeout%\PYG{o}{'}% %\PYG{o}{=}% %\PYG{o}{[}%electionTimeout %\PYG{k}{EXCEPT}% %\PYG{o}{![}%i%\PYG{o}{]}% %\PYG{o}{=}% 0%\PYG{o}{]}%
    %\PYG{o}{/\symbol{92}}% state%\PYG{o}{'}% %\PYG{o}{=}% %\PYG{o}{[}%state %\PYG{k}{EXCEPT}% %\PYG{o}{![}%i%\PYG{o}{]}% %\PYG{o}{=}% %\PYG{k}{IF}% state%\PYG{o}{[}%i%\PYG{o}{]}% %\PYG{o}{=}% Follower %\PYG{k}{THEN}% PreCandidate %\PYG{k}{ELSE}% @%\PYG{o}{]}%
    %\PYG{o}{/\symbol{92}}% %\PYG{k}{IF}% state%\PYG{o}{[}%i%\PYG{o}{]}% %\PYG{o}{=}% Follower
       %\PYG{k}{THEN}% %\PYG{n+nf}{Send}%(%\PYG{o}{[}%type %\PYG{o}{|->}% MsgHup, from %\PYG{o}{|->}% i, to %\PYG{o}{|->}% i, term %\PYG{o}{|->}% currentTerm%\PYG{o}{[}%i%\PYG{o}{]}%%\PYG{o}{]}%)
       %\PYG{k}{ELSE}% messages%\PYG{o}{'}% %\PYG{o}{=}% messages
    %\PYG{o}{/\symbol{92}}% %\PYG{k}{UNCHANGED}% %\PYG{o}{<}%%\PYG{o}{<}%currentTerm, votedFor, log, commitIndex, nextIndex, matchIndex, leader%\PYG{o}{>}%%\PYG{o}{>}%

%\PYG{c+c1}{\symbol{92}* Start prevote campaign}%
%\PYG{n+nf}{StartPreVote}%(i) %\PYG{o}{==}%
    %\PYG{o}{/\symbol{92}}% state%\PYG{o}{[}%i%\PYG{o}{]}% %\PYG{o}{=}% PreCandidate
    %\PYG{o}{/\symbol{92}}% state%\PYG{o}{'}% %\PYG{o}{=}% %\PYG{o}{[}%state %\PYG{k}{EXCEPT}% %\PYG{o}{![}%i%\PYG{o}{]}% %\PYG{o}{=}% PreCandidate%\PYG{o}{]}%
    %\PYG{o}{/\symbol{92}}% %\PYG{k}{LET}% lastIndex %\PYG{o}{==}% %\PYG{n+nf}{Len}%(log%\PYG{o}{[}%i%\PYG{o}{]}%)
           lastTerm %\PYG{o}{==}% %\PYG{n+nf}{LastTerm}%(log%\PYG{o}{[}%i%\PYG{o}{]}%)
       %\PYG{k}{IN}% messages%\PYG{o}{'}% %\PYG{o}{=}% messages %\PYG{o}{\symbol{92}cup}%
            %\PYG{o}{\{}%%\PYG{o}{[}%type %\PYG{o}{|->}% MsgPreVote,
              from %\PYG{o}{|->}% i,
              to %\PYG{o}{|->}% j,
              term %\PYG{o}{|->}% currentTerm%\PYG{o}{[}%i%\PYG{o}{]}% + 1,
              lastLogIndex %\PYG{o}{|->}% lastIndex,
              lastLogTerm %\PYG{o}{|->}% lastTerm%\PYG{o}{]}% : j %\PYG{o}{\symbol{92}in}% Server %\PYG{o}{\symbol{92} }%%\PYG{o}{\{}%i%\PYG{o}{\}}%%\PYG{o}{\}}%
    %\PYG{o}{/\symbol{92}}% %\PYG{k}{UNCHANGED}% %\PYG{o}{<}%%\PYG{o}{<}%currentTerm, votedFor, log, commitIndex, nextIndex, matchIndex, electionTimeout, leader%\PYG{o}{>}%%\PYG{o}{>}%

%\PYG{c+c1}{\symbol{92}* Handle prevote response}%
%\PYG{n+nf}{HandlePreVoteResponse}%(i, m) %\PYG{o}{==}%
    %\PYG{o}{/\symbol{92}}% m.type %\PYG{o}{=}% MsgPreVoteResp
    %\PYG{o}{/\symbol{92}}% m.to %\PYG{o}{=}% i
    %\PYG{o}{/\symbol{92}}% state%\PYG{o}{[}%i%\PYG{o}{]}% %\PYG{o}{=}% PreCandidate
    %\PYG{o}{/\symbol{92}}% m.term %\PYG{o}{=}% currentTerm%\PYG{o}{[}%i%\PYG{o}{]}% + 1
    %\PYG{o}{/\symbol{92}}% %\PYG{k}{LET}% votes %\PYG{o}{==}% %\PYG{o}{\{}%msg %\PYG{o}{\symbol{92}in}% messages : msg.type %\PYG{o}{=}% MsgPreVoteResp %\PYG{o}{/\symbol{92}}%
                                       msg.to %\PYG{o}{=}% i %\PYG{o}{/\symbol{92}}%
                                       msg.term %\PYG{o}{=}% currentTerm%\PYG{o}{[}%i%\PYG{o}{]}% + 1 %\PYG{o}{/\symbol{92}}%
                                       msg.voteGranted %\PYG{o}{=}% %\PYG{k}{TRUE}%%\PYG{o}{\}}%
           voteCount %\PYG{o}{==}% %\PYG{n+nf}{Cardinality}%(%\PYG{o}{\{}%msg.from : msg %\PYG{o}{\symbol{92}in}% votes%\PYG{o}{\}}%) + 1  %\PYG{c+c1}{\symbol{92}* +1 for self}%
       %\PYG{k}{IN}% %\PYG{k}{IF}% voteCount %\PYG{o}{>}% %\PYG{n+nf}{Cardinality}%(Server) %\PYG{o}{\symbol{92}div}% 2
          %\PYG{k}{THEN}% %\PYG{o}{/\symbol{92}}% state%\PYG{o}{'}% %\PYG{o}{=}% %\PYG{o}{[}%state %\PYG{k}{EXCEPT}% %\PYG{o}{![}%i%\PYG{o}{]}% %\PYG{o}{=}% Candidate%\PYG{o}{]}%
               %\PYG{o}{/\symbol{92}}% currentTerm%\PYG{o}{'}% %\PYG{o}{=}% %\PYG{o}{[}%currentTerm %\PYG{k}{EXCEPT}% %\PYG{o}{![}%i%\PYG{o}{]}% %\PYG{o}{=}% currentTerm%\PYG{o}{[}%i%\PYG{o}{]}% + 1%\PYG{o}{]}%
               %\PYG{o}{/\symbol{92}}% votedFor%\PYG{o}{'}% %\PYG{o}{=}% %\PYG{o}{[}%votedFor %\PYG{k}{EXCEPT}% %\PYG{o}{![}%i%\PYG{o}{]}% %\PYG{o}{=}% i%\PYG{o}{]}%
               %\PYG{o}{/\symbol{92}}% %\PYG{k}{LET}% lastIndex %\PYG{o}{==}% %\PYG{n+nf}{Len}%(log%\PYG{o}{[}%i%\PYG{o}{]}%)
                      lastTerm %\PYG{o}{==}% %\PYG{n+nf}{LastTerm}%(log%\PYG{o}{[}%i%\PYG{o}{]}%)
                  %\PYG{k}{IN}% messages%\PYG{o}{'}% %\PYG{o}{=}% (messages %\PYG{o}{\symbol{92} }%%\PYG{o}{\{}%m%\PYG{o}{\}}%) %\PYG{o}{\symbol{92}cup}%
                       %\PYG{o}{\{}%%\PYG{o}{[}%type %\PYG{o}{|->}% MsgVote,
                         from %\PYG{o}{|->}% i,
                         to %\PYG{o}{|->}% j,
                         term %\PYG{o}{|->}% currentTerm%\PYG{o}{[}%i%\PYG{o}{]}% + 1,
                         lastLogIndex %\PYG{o}{|->}% lastIndex,
                         lastLogTerm %\PYG{o}{|->}% lastTerm%\PYG{o}{]}% : j %\PYG{o}{\symbol{92}in}% Server %\PYG{o}{\symbol{92} }%%\PYG{o}{\{}%i%\PYG{o}{\}}%%\PYG{o}{\}}%
               %\PYG{o}{/\symbol{92}}% %\PYG{k}{UNCHANGED}% %\PYG{o}{<}%%\PYG{o}{<}%log, commitIndex, nextIndex, matchIndex, electionTimeout, leader%\PYG{o}{>}%%\PYG{o}{>}%
          %\PYG{k}{ELSE}% %\PYG{o}{/\symbol{92}}% messages%\PYG{o}{'}% %\PYG{o}{=}% messages %\PYG{o}{\symbol{92} }%%\PYG{o}{\{}%m%\PYG{o}{\}}%
               %\PYG{o}{/\symbol{92}}% %\PYG{k}{UNCHANGED}% %\PYG{o}{<}%%\PYG{o}{<}%state, currentTerm, votedFor, log, commitIndex, nextIndex, matchIndex, electionTimeout, leader%\PYG{o}{>}%%\PYG{o}{>}%

%\PYG{c+c1}{\symbol{92}* Handle vote request}%
%\PYG{n+nf}{HandleVoteRequest}%(i, m) %\PYG{o}{==}%
    %\PYG{o}{/\symbol{92}}% m.type %\PYG{o}{\symbol{92}in}% %\PYG{o}{\{}%MsgVote, MsgPreVote%\PYG{o}{\}}%
    %\PYG{o}{/\symbol{92}}% m.to %\PYG{o}{=}% i
    %\PYG{o}{/\symbol{92}}% %\PYG{k}{LET}% logOk %\PYG{o}{==}% %\PYG{o}{\symbol{92}/}% m.lastLogTerm %\PYG{o}{>}% %\PYG{n+nf}{LastTerm}%(log%\PYG{o}{[}%i%\PYG{o}{]}%)
                    %\PYG{o}{\symbol{92}/}% %\PYG{o}{/\symbol{92}}% m.lastLogTerm %\PYG{o}{=}% %\PYG{n+nf}{LastTerm}%(log%\PYG{o}{[}%i%\PYG{o}{]}%)
                       %\PYG{o}{/\symbol{92}}% m.lastLogIndex %\PYG{o}{>}%%\PYG{o}{=}% %\PYG{n+nf}{Len}%(log%\PYG{o}{[}%i%\PYG{o}{]}%)
           grant %\PYG{o}{==}% %\PYG{o}{/\symbol{92}}% m.term %\PYG{o}{>}%%\PYG{o}{=}% currentTerm%\PYG{o}{[}%i%\PYG{o}{]}%
                   %\PYG{o}{/\symbol{92}}% logOk
                   %\PYG{o}{/\symbol{92}}% %\PYG{k}{IF}% m.type %\PYG{o}{=}% MsgVote
                      %\PYG{k}{THEN}% %\PYG{o}{\symbol{92}/}% votedFor%\PYG{o}{[}%i%\PYG{o}{]}% %\PYG{o}{=}% Nil
                           %\PYG{o}{\symbol{92}/}% votedFor%\PYG{o}{[}%i%\PYG{o}{]}% %\PYG{o}{=}% m.from
                      %\PYG{k}{ELSE}% %\PYG{k}{TRUE}%
       %\PYG{k}{IN}% %\PYG{o}{/\symbol{92}}% %\PYG{k}{IF}% m.type %\PYG{o}{=}% MsgVote %\PYG{o}{/\symbol{92}}% m.term %\PYG{o}{>}% currentTerm%\PYG{o}{[}%i%\PYG{o}{]}%
             %\PYG{k}{THEN}% %\PYG{o}{/\symbol{92}}% state%\PYG{o}{'}% %\PYG{o}{=}% %\PYG{o}{[}%state %\PYG{k}{EXCEPT}% %\PYG{o}{![}%i%\PYG{o}{]}% %\PYG{o}{=}% Follower%\PYG{o}{]}%
                  %\PYG{o}{/\symbol{92}}% currentTerm%\PYG{o}{'}% %\PYG{o}{=}% %\PYG{o}{[}%currentTerm %\PYG{k}{EXCEPT}% %\PYG{o}{![}%i%\PYG{o}{]}% %\PYG{o}{=}% m.term%\PYG{o}{]}%
                  %\PYG{o}{/\symbol{92}}% votedFor%\PYG{o}{'}% %\PYG{o}{=}% %\PYG{o}{[}%votedFor %\PYG{k}{EXCEPT}% %\PYG{o}{![}%i%\PYG{o}{]}% %\PYG{o}{=}% %\PYG{k}{IF}% grant %\PYG{k}{THEN}% m.from %\PYG{k}{ELSE}% Nil%\PYG{o}{]}%
                  %\PYG{o}{/\symbol{92}}% leader%\PYG{o}{'}% %\PYG{o}{=}% %\PYG{o}{[}%leader %\PYG{k}{EXCEPT}% %\PYG{o}{![}%i%\PYG{o}{]}% %\PYG{o}{=}% Nil%\PYG{o}{]}%
             %\PYG{k}{ELSE}% %\PYG{k}{IF}% m.type %\PYG{o}{=}% MsgVote %\PYG{o}{/\symbol{92}}% grant
                  %\PYG{k}{THEN}% %\PYG{o}{/\symbol{92}}% votedFor%\PYG{o}{'}% %\PYG{o}{=}% %\PYG{o}{[}%votedFor %\PYG{k}{EXCEPT}% %\PYG{o}{![}%i%\PYG{o}{]}% %\PYG{o}{=}% m.from%\PYG{o}{]}%
                       %\PYG{o}{/\symbol{92}}% %\PYG{k}{UNCHANGED}% %\PYG{o}{<}%%\PYG{o}{<}%state, currentTerm, leader%\PYG{o}{>}%%\PYG{o}{>}%
                  %\PYG{k}{ELSE}% %\PYG{k}{UNCHANGED}% %\PYG{o}{<}%%\PYG{o}{<}%state, currentTerm, votedFor, leader%\PYG{o}{>}%%\PYG{o}{>}%
          %\PYG{o}{/\symbol{92}}% %\PYG{n+nf}{Send}%(%\PYG{o}{[}%type %\PYG{o}{|->}% %\PYG{k}{IF}% m.type %\PYG{o}{=}% MsgVote %\PYG{k}{THEN}% MsgVoteResp %\PYG{k}{ELSE}% MsgPreVoteResp,
                   from %\PYG{o}{|->}% i,
                   to %\PYG{o}{|->}% m.from,
                   term %\PYG{o}{|->}% %\PYG{k}{IF}% m.type %\PYG{o}{=}% MsgPreVote %\PYG{k}{THEN}% m.term %\PYG{k}{ELSE}% currentTerm%\PYG{o}{'}%%\PYG{o}{[}%i%\PYG{o}{]}%,
                   voteGranted %\PYG{o}{|->}% grant%\PYG{o}{]}%)
          %\PYG{o}{/\symbol{92}}% messages%\PYG{o}{'}% %\PYG{o}{=}% messages %\PYG{o}{\symbol{92} }%%\PYG{o}{\{}%m%\PYG{o}{\}}%
          %\PYG{o}{/\symbol{92}}% %\PYG{k}{UNCHANGED}% %\PYG{o}{<}%%\PYG{o}{<}%log, commitIndex, nextIndex, matchIndex, electionTimeout%\PYG{o}{>}%%\PYG{o}{>}%

%\PYG{c+c1}{\symbol{92}* Handle vote response}%
%\PYG{n+nf}{HandleVoteResponse}%(i, m) %\PYG{o}{==}%
    %\PYG{o}{/\symbol{92}}% m.type %\PYG{o}{=}% MsgVoteResp
    %\PYG{o}{/\symbol{92}}% m.to %\PYG{o}{=}% i
    %\PYG{o}{/\symbol{92}}% state%\PYG{o}{[}%i%\PYG{o}{]}% %\PYG{o}{=}% Candidate
    %\PYG{o}{/\symbol{92}}% m.term %\PYG{o}{=}% currentTerm%\PYG{o}{[}%i%\PYG{o}{]}%
    %\PYG{o}{/\symbol{92}}% %\PYG{k}{LET}% votes %\PYG{o}{==}% %\PYG{o}{\{}%msg %\PYG{o}{\symbol{92}in}% messages : msg.type %\PYG{o}{=}% MsgVoteResp %\PYG{o}{/\symbol{92}}%
                                       msg.to %\PYG{o}{=}% i %\PYG{o}{/\symbol{92}}%
                                       msg.term %\PYG{o}{=}% currentTerm%\PYG{o}{[}%i%\PYG{o}{]}% %\PYG{o}{/\symbol{92}}%
                                       msg.voteGranted %\PYG{o}{=}% %\PYG{k}{TRUE}%%\PYG{o}{\}}%
           voteCount %\PYG{o}{==}% %\PYG{n+nf}{Cardinality}%(%\PYG{o}{\{}%msg.from : msg %\PYG{o}{\symbol{92}in}% votes%\PYG{o}{\}}%) + 1  %\PYG{c+c1}{\symbol{92}* +1 for self vote}%
       %\PYG{k}{IN}% %\PYG{k}{IF}% voteCount %\PYG{o}{>}% %\PYG{n+nf}{Cardinality}%(Server) %\PYG{o}{\symbol{92}div}% 2
          %\PYG{k}{THEN}% %\PYG{o}{/\symbol{92}}% state%\PYG{o}{'}% %\PYG{o}{=}% %\PYG{o}{[}%state %\PYG{k}{EXCEPT}% %\PYG{o}{![}%i%\PYG{o}{]}% %\PYG{o}{=}% Leader%\PYG{o}{]}%
               %\PYG{o}{/\symbol{92}}% leader%\PYG{o}{'}% %\PYG{o}{=}% %\PYG{o}{[}%leader %\PYG{k}{EXCEPT}% %\PYG{o}{![}%i%\PYG{o}{]}% %\PYG{o}{=}% i%\PYG{o}{]}%
               %\PYG{o}{/\symbol{92}}% nextIndex%\PYG{o}{'}% %\PYG{o}{=}% %\PYG{o}{[}%nextIndex %\PYG{k}{EXCEPT}% %\PYG{o}{![}%i%\PYG{o}{]}% %\PYG{o}{=}% %\PYG{o}{[}%j %\PYG{o}{\symbol{92}in}% Server %\PYG{o}{|->}% %\PYG{n+nf}{Len}%(log%\PYG{o}{[}%i%\PYG{o}{]}%) + 1%\PYG{o}{]}%%\PYG{o}{]}%
               %\PYG{o}{/\symbol{92}}% matchIndex%\PYG{o}{'}% %\PYG{o}{=}% %\PYG{o}{[}%matchIndex %\PYG{k}{EXCEPT}% %\PYG{o}{![}%i%\PYG{o}{]}% %\PYG{o}{=}% %\PYG{o}{[}%j %\PYG{o}{\symbol{92}in}% Server %\PYG{o}{|->}% 0%\PYG{o}{]}%%\PYG{o}{]}%
               %\PYG{o}{/\symbol{92}}% %\PYG{k}{LET}% entry %\PYG{o}{==}% %\PYG{o}{[}%term %\PYG{o}{|->}% currentTerm%\PYG{o}{[}%i%\PYG{o}{]}%, value %\PYG{o}{|->}% Nil%\PYG{o}{]}%
                  %\PYG{k}{IN}% %\PYG{o}{/\symbol{92}}% log%\PYG{o}{'}% %\PYG{o}{=}% %\PYG{o}{[}%log %\PYG{k}{EXCEPT}% %\PYG{o}{![}%i%\PYG{o}{]}% %\PYG{o}{=}% %\PYG{n+nf}{Append}%(@, entry)%\PYG{o}{]}%
                     %\PYG{o}{/\symbol{92}}% messages%\PYG{o}{'}% %\PYG{o}{=}% (messages %\PYG{o}{\symbol{92} }%%\PYG{o}{\{}%m%\PYG{o}{\}}%) %\PYG{o}{\symbol{92}cup}%
                          %\PYG{o}{\{}%%\PYG{o}{[}%type %\PYG{o}{|->}% MsgApp,
                            from %\PYG{o}{|->}% i,
                            to %\PYG{o}{|->}% j,
                            term %\PYG{o}{|->}% currentTerm%\PYG{o}{[}%i%\PYG{o}{]}%,
                            prevLogIndex %\PYG{o}{|->}% %\PYG{n+nf}{Len}%(log%\PYG{o}{[}%i%\PYG{o}{]}%),
                            prevLogTerm %\PYG{o}{|->}% %\PYG{n+nf}{LastTerm}%(log%\PYG{o}{[}%i%\PYG{o}{]}%),
                            entries %\PYG{o}{|->}% %\PYG{o}{<}%%\PYG{o}{<}%entry%\PYG{o}{>}%%\PYG{o}{>}%,
                            leaderCommit %\PYG{o}{|->}% commitIndex%\PYG{o}{[}%i%\PYG{o}{]}%%\PYG{o}{]}% : j %\PYG{o}{\symbol{92}in}% Server %\PYG{o}{\symbol{92} }%%\PYG{o}{\{}%i%\PYG{o}{\}}%%\PYG{o}{\}}%
               %\PYG{o}{/\symbol{92}}% %\PYG{k}{UNCHANGED}% %\PYG{o}{<}%%\PYG{o}{<}%currentTerm, votedFor, commitIndex, electionTimeout%\PYG{o}{>}%%\PYG{o}{>}%
          %\PYG{k}{ELSE}% %\PYG{o}{/\symbol{92}}% messages%\PYG{o}{'}% %\PYG{o}{=}% messages %\PYG{o}{\symbol{92} }%%\PYG{o}{\{}%m%\PYG{o}{\}}%
               %\PYG{o}{/\symbol{92}}% %\PYG{k}{UNCHANGED}% %\PYG{o}{<}%%\PYG{o}{<}%state, currentTerm, votedFor, log, commitIndex, nextIndex, matchIndex, electionTimeout, leader%\PYG{o}{>}%%\PYG{o}{>}%

%\PYG{c+c1}{\symbol{92}* Client request (leader only)}%
%\PYG{n+nf}{ClientRequest}%(i, v) %\PYG{o}{==}%
    %\PYG{o}{/\symbol{92}}% state%\PYG{o}{[}%i%\PYG{o}{]}% %\PYG{o}{=}% Leader
    %\PYG{o}{/\symbol{92}}% %\PYG{k}{LET}% entry %\PYG{o}{==}% %\PYG{o}{[}%term %\PYG{o}{|->}% currentTerm%\PYG{o}{[}%i%\PYG{o}{]}%, value %\PYG{o}{|->}% v%\PYG{o}{]}%
           newLog %\PYG{o}{==}% %\PYG{n+nf}{Append}%(log%\PYG{o}{[}%i%\PYG{o}{]}%, entry)
       %\PYG{k}{IN}% %\PYG{o}{/\symbol{92}}% log%\PYG{o}{'}% %\PYG{o}{=}% %\PYG{o}{[}%log %\PYG{k}{EXCEPT}% %\PYG{o}{![}%i%\PYG{o}{]}% %\PYG{o}{=}% newLog%\PYG{o}{]}%
          %\PYG{o}{/\symbol{92}}% messages%\PYG{o}{'}% %\PYG{o}{=}% messages %\PYG{o}{\symbol{92}cup}%
               %\PYG{o}{\{}%%\PYG{o}{[}%type %\PYG{o}{|->}% MsgApp,
                 from %\PYG{o}{|->}% i,
                 to %\PYG{o}{|->}% j,
                 term %\PYG{o}{|->}% currentTerm%\PYG{o}{[}%i%\PYG{o}{]}%,
                 prevLogIndex %\PYG{o}{|->}% %\PYG{n+nf}{Len}%(log%\PYG{o}{[}%i%\PYG{o}{]}%),
                 prevLogTerm %\PYG{o}{|->}% %\PYG{n+nf}{LastTerm}%(log%\PYG{o}{[}%i%\PYG{o}{]}%),
                 entries %\PYG{o}{|->}% %\PYG{o}{<}%%\PYG{o}{<}%entry%\PYG{o}{>}%%\PYG{o}{>}%,
                 leaderCommit %\PYG{o}{|->}% commitIndex%\PYG{o}{[}%i%\PYG{o}{]}%%\PYG{o}{]}% : j %\PYG{o}{\symbol{92}in}% Server %\PYG{o}{\symbol{92} }%%\PYG{o}{\{}%i%\PYG{o}{\}}%%\PYG{o}{\}}%
    %\PYG{o}{/\symbol{92}}% %\PYG{k}{UNCHANGED}% %\PYG{o}{<}%%\PYG{o}{<}%state, currentTerm, votedFor, commitIndex, nextIndex, matchIndex, electionTimeout, leader%\PYG{o}{>}%%\PYG{o}{>}%

%\PYG{c+c1}{\symbol{92}* Handle append entries}%
%\PYG{n+nf}{HandleAppendEntries}%(i, m) %\PYG{o}{==}%
    %\PYG{o}{/\symbol{92}}% m.type %\PYG{o}{=}% MsgApp
    %\PYG{o}{/\symbol{92}}% m.to %\PYG{o}{=}% i
    %\PYG{o}{/\symbol{92}}% %\PYG{k}{IF}% m.term %\PYG{o}{>}% currentTerm%\PYG{o}{[}%i%\PYG{o}{]}%
       %\PYG{k}{THEN}% %\PYG{o}{/\symbol{92}}% currentTerm%\PYG{o}{'}% %\PYG{o}{=}% %\PYG{o}{[}%currentTerm %\PYG{k}{EXCEPT}% %\PYG{o}{![}%i%\PYG{o}{]}% %\PYG{o}{=}% m.term%\PYG{o}{]}%
            %\PYG{o}{/\symbol{92}}% state%\PYG{o}{'}% %\PYG{o}{=}% %\PYG{o}{[}%state %\PYG{k}{EXCEPT}% %\PYG{o}{![}%i%\PYG{o}{]}% %\PYG{o}{=}% Follower%\PYG{o}{]}%
            %\PYG{o}{/\symbol{92}}% votedFor%\PYG{o}{'}% %\PYG{o}{=}% %\PYG{o}{[}%votedFor %\PYG{k}{EXCEPT}% %\PYG{o}{![}%i%\PYG{o}{]}% %\PYG{o}{=}% Nil%\PYG{o}{]}%
            %\PYG{o}{/\symbol{92}}% leader%\PYG{o}{'}% %\PYG{o}{=}% %\PYG{o}{[}%leader %\PYG{k}{EXCEPT}% %\PYG{o}{![}%i%\PYG{o}{]}% %\PYG{o}{=}% m.from%\PYG{o}{]}%
       %\PYG{k}{ELSE}% %\PYG{o}{/\symbol{92}}% %\PYG{k}{UNCHANGED}% %\PYG{o}{<}%%\PYG{o}{<}%currentTerm, votedFor%\PYG{o}{>}%%\PYG{o}{>}%
            %\PYG{o}{/\symbol{92}}% %\PYG{k}{IF}% m.term %\PYG{o}{=}% currentTerm%\PYG{o}{[}%i%\PYG{o}{]}%
               %\PYG{k}{THEN}% %\PYG{o}{/\symbol{92}}% state%\PYG{o}{'}% %\PYG{o}{=}% %\PYG{o}{[}%state %\PYG{k}{EXCEPT}% %\PYG{o}{![}%i%\PYG{o}{]}% %\PYG{o}{=}% Follower%\PYG{o}{]}%
                    %\PYG{o}{/\symbol{92}}% leader%\PYG{o}{'}% %\PYG{o}{=}% %\PYG{o}{[}%leader %\PYG{k}{EXCEPT}% %\PYG{o}{![}%i%\PYG{o}{]}% %\PYG{o}{=}% m.from%\PYG{o}{]}%
               %\PYG{k}{ELSE}% %\PYG{k}{UNCHANGED}% %\PYG{o}{<}%%\PYG{o}{<}%state, leader%\PYG{o}{>}%%\PYG{o}{>}%
    %\PYG{o}{/\symbol{92}}% electionTimeout%\PYG{o}{'}% %\PYG{o}{=}% %\PYG{o}{[}%electionTimeout %\PYG{k}{EXCEPT}% %\PYG{o}{![}%i%\PYG{o}{]}% %\PYG{o}{=}% 0%\PYG{o}{]}%
    %\PYG{o}{/\symbol{92}}% %\PYG{k}{LET}% logOk %\PYG{o}{==}% %\PYG{o}{\symbol{92}/}% m.prevLogIndex %\PYG{o}{=}% 0
                    %\PYG{o}{\symbol{92}/}% %\PYG{o}{/\symbol{92}}% m.prevLogIndex %\PYG{o}{<}%%\PYG{o}{=}% %\PYG{n+nf}{Len}%(log%\PYG{o}{[}%i%\PYG{o}{]}%)
                       %\PYG{o}{/\symbol{92}}% log%\PYG{o}{[}%i%\PYG{o}{]}%%\PYG{o}{[}%m.prevLogIndex%\PYG{o}{]}%.term %\PYG{o}{=}% m.prevLogTerm
       %\PYG{k}{IN}% %\PYG{k}{IF}% logOk
          %\PYG{k}{THEN}% %\PYG{o}{/\symbol{92}}% log%\PYG{o}{'}% %\PYG{o}{=}% %\PYG{o}{[}%log %\PYG{k}{EXCEPT}% %\PYG{o}{![}%i%\PYG{o}{]}% %\PYG{o}{=}% %\PYG{n+nf}{SubSeq}%(@, 1, m.prevLogIndex) %\symbol{92}%o m.entries%\PYG{o}{]}%
               %\PYG{o}{/\symbol{92}}% commitIndex%\PYG{o}{'}% %\PYG{o}{=}% %\PYG{o}{[}%commitIndex %\PYG{k}{EXCEPT}% %\PYG{o}{![}%i%\PYG{o}{]}% %\PYG{o}{=}% %\PYG{k}{IF}% m.leaderCommit %\PYG{o}{>}% commitIndex%\PYG{o}{[}%i%\PYG{o}{]}%
                                                            %\PYG{k}{THEN}% %\PYG{n+nf}{Min}%(m.leaderCommit, %\PYG{n+nf}{Len}%(log%\PYG{o}{'}%%\PYG{o}{[}%i%\PYG{o}{]}%))
                                                            %\PYG{k}{ELSE}% commitIndex%\PYG{o}{[}%i%\PYG{o}{]}%%\PYG{o}{]}%
               %\PYG{o}{/\symbol{92}}% %\PYG{n+nf}{Send}%(%\PYG{o}{[}%type %\PYG{o}{|->}% MsgAppResp,
                        from %\PYG{o}{|->}% i,
                        to %\PYG{o}{|->}% m.from,
                        term %\PYG{o}{|->}% currentTerm%\PYG{o}{'}%%\PYG{o}{[}%i%\PYG{o}{]}%,
                        success %\PYG{o}{|->}% %\PYG{k}{TRUE}%,
                        matchIndex %\PYG{o}{|->}% m.prevLogIndex + %\PYG{n+nf}{Len}%(m.entries)%\PYG{o}{]}%)
          %\PYG{k}{ELSE}% %\PYG{o}{/\symbol{92}}% %\PYG{n+nf}{Send}%(%\PYG{o}{[}%type %\PYG{o}{|->}% MsgAppResp,
                        from %\PYG{o}{|->}% i,
                        to %\PYG{o}{|->}% m.from,
                        term %\PYG{o}{|->}% currentTerm%\PYG{o}{'}%%\PYG{o}{[}%i%\PYG{o}{]}%,
                        success %\PYG{o}{|->}% %\PYG{k}{FALSE}%,
                        matchIndex %\PYG{o}{|->}% 0%\PYG{o}{]}%)
               %\PYG{o}{/\symbol{92}}% %\PYG{k}{UNCHANGED}% %\PYG{o}{<}%%\PYG{o}{<}%log, commitIndex%\PYG{o}{>}%%\PYG{o}{>}%
    %\PYG{o}{/\symbol{92}}% messages%\PYG{o}{'}% %\PYG{o}{=}% messages %\PYG{o}{\symbol{92} }%%\PYG{o}{\{}%m%\PYG{o}{\}}%
    %\PYG{o}{/\symbol{92}}% %\PYG{k}{UNCHANGED}% %\PYG{o}{<}%%\PYG{o}{<}%nextIndex, matchIndex%\PYG{o}{>}%%\PYG{o}{>}%

%\PYG{c+c1}{\symbol{92}* Handle append response}%
%\PYG{n+nf}{HandleAppendResponse}%(i, m) %\PYG{o}{==}%
    %\PYG{o}{/\symbol{92}}% m.type %\PYG{o}{=}% MsgAppResp
    %\PYG{o}{/\symbol{92}}% m.to %\PYG{o}{=}% i
    %\PYG{o}{/\symbol{92}}% state%\PYG{o}{[}%i%\PYG{o}{]}% %\PYG{o}{=}% Leader
    %\PYG{o}{/\symbol{92}}% m.term %\PYG{o}{=}% currentTerm%\PYG{o}{[}%i%\PYG{o}{]}%
    %\PYG{o}{/\symbol{92}}% %\PYG{k}{IF}% m.success
       %\PYG{k}{THEN}% %\PYG{o}{/\symbol{92}}% matchIndex%\PYG{o}{'}% %\PYG{o}{=}% %\PYG{o}{[}%matchIndex %\PYG{k}{EXCEPT}% %\PYG{o}{![}%i%\PYG{o}{]}%%\PYG{o}{[}%m.from%\PYG{o}{]}% %\PYG{o}{=}% m.matchIndex%\PYG{o}{]}%
            %\PYG{o}{/\symbol{92}}% nextIndex%\PYG{o}{'}% %\PYG{o}{=}% %\PYG{o}{[}%nextIndex %\PYG{k}{EXCEPT}% %\PYG{o}{![}%i%\PYG{o}{]}%%\PYG{o}{[}%m.from%\PYG{o}{]}% %\PYG{o}{=}% m.matchIndex + 1%\PYG{o}{]}%
            %\PYG{o}{/\symbol{92}}% %\PYG{k}{LET}% %\PYG{n+nf}{Agree}%(idx) %\PYG{o}{==}% %\PYG{o}{\{}%i%\PYG{o}{\}}% %\PYG{o}{\symbol{92}cup}% %\PYG{o}{\{}%s %\PYG{o}{\symbol{92}in}% Server : matchIndex%\PYG{o}{'}%%\PYG{o}{[}%i%\PYG{o}{]}%%\PYG{o}{[}%s%\PYG{o}{]}% %\PYG{o}{>}%%\PYG{o}{=}% idx%\PYG{o}{\}}%
                   agreeIndexes %\PYG{o}{==}% %\PYG{o}{\{}%idx %\PYG{o}{\symbol{92}in}% 1%\PYG{o}{..}%%\PYG{n+nf}{Len}%(log%\PYG{o}{[}%i%\PYG{o}{]}%) :
                                     %\PYG{n+nf}{Cardinality}%(%\PYG{n+nf}{Agree}%(idx)) %\PYG{o}{>}% %\PYG{n+nf}{Cardinality}%(Server) %\PYG{o}{\symbol{92}div}% 2 %\PYG{o}{/\symbol{92}}%
                                     log%\PYG{o}{[}%i%\PYG{o}{]}%%\PYG{o}{[}%idx%\PYG{o}{]}%.term %\PYG{o}{=}% currentTerm%\PYG{o}{[}%i%\PYG{o}{]}%%\PYG{o}{\}}%
               %\PYG{k}{IN}% commitIndex%\PYG{o}{'}% %\PYG{o}{=}% %\PYG{o}{[}%commitIndex %\PYG{k}{EXCEPT}% %\PYG{o}{![}%i%\PYG{o}{]}% %\PYG{o}{=}% %\PYG{k}{IF}% agreeIndexes %\PYG{o}{/=}% %\PYG{o}{\{}%%\PYG{o}{\}}%
                                                            %\PYG{k}{THEN}% %\PYG{n+nf}{Max}%(commitIndex%\PYG{o}{[}%i%\PYG{o}{]}%, CHOOSE idx %\PYG{o}{\symbol{92}in}% agreeIndexes :
                                                                     %\PYG{o}{\symbol{92}A}% idx2 %\PYG{o}{\symbol{92}in}% agreeIndexes : idx %\PYG{o}{>}%%\PYG{o}{=}% idx2)
                                                            %\PYG{k}{ELSE}% commitIndex%\PYG{o}{[}%i%\PYG{o}{]}%%\PYG{o}{]}%
       %\PYG{k}{ELSE}% %\PYG{o}{/\symbol{92}}% nextIndex%\PYG{o}{'}% %\PYG{o}{=}% %\PYG{o}{[}%nextIndex %\PYG{k}{EXCEPT}% %\PYG{o}{![}%i%\PYG{o}{]}%%\PYG{o}{[}%m.from%\PYG{o}{]}% %\PYG{o}{=}% %\PYG{n+nf}{Max}%(1, nextIndex%\PYG{o}{[}%i%\PYG{o}{]}%%\PYG{o}{[}%m.from%\PYG{o}{]}% - 1)%\PYG{o}{]}%
            %\PYG{o}{/\symbol{92}}% %\PYG{k}{UNCHANGED}% %\PYG{o}{<}%%\PYG{o}{<}%matchIndex, commitIndex%\PYG{o}{>}%%\PYG{o}{>}%
    %\PYG{o}{/\symbol{92}}% messages%\PYG{o}{'}% %\PYG{o}{=}% messages %\PYG{o}{\symbol{92} }%%\PYG{o}{\{}%m%\PYG{o}{\}}%
    %\PYG{o}{/\symbol{92}}% %\PYG{k}{UNCHANGED}% %\PYG{o}{<}%%\PYG{o}{<}%state, currentTerm, votedFor, log, electionTimeout, leader%\PYG{o}{>}%%\PYG{o}{>}%

%\PYG{c+c1}{\symbol{92}* Send heartbeat}%
%\PYG{n+nf}{SendHeartbeat}%(i) %\PYG{o}{==}%
    %\PYG{o}{/\symbol{92}}% state%\PYG{o}{[}%i%\PYG{o}{]}% %\PYG{o}{=}% Leader
    %\PYG{o}{/\symbol{92}}% messages%\PYG{o}{'}% %\PYG{o}{=}% messages %\PYG{o}{\symbol{92}cup}%
         %\PYG{o}{\{}%%\PYG{o}{[}%type %\PYG{o}{|->}% MsgHeartbeat,
           from %\PYG{o}{|->}% i,
           to %\PYG{o}{|->}% j,
           term %\PYG{o}{|->}% currentTerm%\PYG{o}{[}%i%\PYG{o}{]}%,
           leaderCommit %\PYG{o}{|->}% commitIndex%\PYG{o}{[}%i%\PYG{o}{]}%%\PYG{o}{]}% : j %\PYG{o}{\symbol{92}in}% Server %\PYG{o}{\symbol{92} }%%\PYG{o}{\{}%i%\PYG{o}{\}}%%\PYG{o}{\}}%
    %\PYG{o}{/\symbol{92}}% %\PYG{k}{UNCHANGED}% %\PYG{o}{<}%%\PYG{o}{<}%state, currentTerm, votedFor, log, commitIndex, nextIndex, matchIndex, electionTimeout, leader%\PYG{o}{>}%%\PYG{o}{>}%

%\PYG{c+c1}{\symbol{92}* Handle heartbeat}%
%\PYG{n+nf}{HandleHeartbeat}%(i, m) %\PYG{o}{==}%
    %\PYG{o}{/\symbol{92}}% m.type %\PYG{o}{=}% MsgHeartbeat
    %\PYG{o}{/\symbol{92}}% m.to %\PYG{o}{=}% i
    %\PYG{o}{/\symbol{92}}% %\PYG{k}{IF}% m.term %\PYG{o}{>}%%\PYG{o}{=}% currentTerm%\PYG{o}{[}%i%\PYG{o}{]}%
       %\PYG{k}{THEN}% %\PYG{o}{/\symbol{92}}% currentTerm%\PYG{o}{'}% %\PYG{o}{=}% %\PYG{o}{[}%currentTerm %\PYG{k}{EXCEPT}% %\PYG{o}{![}%i%\PYG{o}{]}% %\PYG{o}{=}% m.term%\PYG{o}{]}%
            %\PYG{o}{/\symbol{92}}% state%\PYG{o}{'}% %\PYG{o}{=}% %\PYG{o}{[}%state %\PYG{k}{EXCEPT}% %\PYG{o}{![}%i%\PYG{o}{]}% %\PYG{o}{=}% Follower%\PYG{o}{]}%
            %\PYG{o}{/\symbol{92}}% votedFor%\PYG{o}{'}% %\PYG{o}{=}% %\PYG{o}{[}%votedFor %\PYG{k}{EXCEPT}% %\PYG{o}{![}%i%\PYG{o}{]}% %\PYG{o}{=}% Nil%\PYG{o}{]}%
            %\PYG{o}{/\symbol{92}}% leader%\PYG{o}{'}% %\PYG{o}{=}% %\PYG{o}{[}%leader %\PYG{k}{EXCEPT}% %\PYG{o}{![}%i%\PYG{o}{]}% %\PYG{o}{=}% m.from%\PYG{o}{]}%
            %\PYG{o}{/\symbol{92}}% electionTimeout%\PYG{o}{'}% %\PYG{o}{=}% %\PYG{o}{[}%electionTimeout %\PYG{k}{EXCEPT}% %\PYG{o}{![}%i%\PYG{o}{]}% %\PYG{o}{=}% 0%\PYG{o}{]}%
            %\PYG{o}{/\symbol{92}}% commitIndex%\PYG{o}{'}% %\PYG{o}{=}% %\PYG{o}{[}%commitIndex %\PYG{k}{EXCEPT}% %\PYG{o}{![}%i%\PYG{o}{]}% %\PYG{o}{=}% %\PYG{k}{IF}% m.leaderCommit %\PYG{o}{>}% commitIndex%\PYG{o}{[}%i%\PYG{o}{]}%
                                                         %\PYG{k}{THEN}% %\PYG{n+nf}{Min}%(m.leaderCommit, %\PYG{n+nf}{Len}%(log%\PYG{o}{[}%i%\PYG{o}{]}%))
                                                         %\PYG{k}{ELSE}% commitIndex%\PYG{o}{[}%i%\PYG{o}{]}%%\PYG{o}{]}%
       %\PYG{k}{ELSE}% %\PYG{k}{UNCHANGED}% %\PYG{o}{<}%%\PYG{o}{<}%currentTerm, state, votedFor, leader, electionTimeout, commitIndex%\PYG{o}{>}%%\PYG{o}{>}%
    %\PYG{o}{/\symbol{92}}% messages%\PYG{o}{'}% %\PYG{o}{=}% messages %\PYG{o}{\symbol{92} }%%\PYG{o}{\{}%m%\PYG{o}{\}}%
    %\PYG{o}{/\symbol{92}}% %\PYG{k}{UNCHANGED}% %\PYG{o}{<}%%\PYG{o}{<}%log, nextIndex, matchIndex%\PYG{o}{>}%%\PYG{o}{>}%

%\PYG{c+c1}{\symbol{92}* Advance election timeout}%
%\PYG{n+nf}{AdvanceElectionTimeout}%(i) %\PYG{o}{==}%
    %\PYG{o}{/\symbol{92}}% state%\PYG{o}{[}%i%\PYG{o}{]}% %\PYG{o}{\symbol{92}in}% %\PYG{o}{\{}%Follower, PreCandidate, Candidate%\PYG{o}{\}}%
    %\PYG{o}{/\symbol{92}}% electionTimeout%\PYG{o}{'}% %\PYG{o}{=}% %\PYG{o}{[}%electionTimeout %\PYG{k}{EXCEPT}% %\PYG{o}{![}%i%\PYG{o}{]}% %\PYG{o}{=}% @ + 1%\PYG{o}{]}%
    %\PYG{o}{/\symbol{92}}% %\PYG{k}{UNCHANGED}% %\PYG{o}{<}%%\PYG{o}{<}%state, currentTerm, votedFor, log, commitIndex, nextIndex, matchIndex, messages, leader%\PYG{o}{>}%%\PYG{o}{>}%

Next %\PYG{o}{==}%
    %\PYG{o}{\symbol{92}/}% %\PYG{o}{\symbol{92}E}% i %\PYG{o}{\symbol{92}in}% Server : %\PYG{n+nf}{Timeout}%(i)
    %\PYG{o}{\symbol{92}/}% %\PYG{o}{\symbol{92}E}% i %\PYG{o}{\symbol{92}in}% Server : %\PYG{n+nf}{StartPreVote}%(i)
    %\PYG{o}{\symbol{92}/}% %\PYG{o}{\symbol{92}E}% i %\PYG{o}{\symbol{92}in}% Server, m %\PYG{o}{\symbol{92}in}% messages : %\PYG{n+nf}{HandlePreVoteResponse}%(i, m)
    %\PYG{o}{\symbol{92}/}% %\PYG{o}{\symbol{92}E}% i %\PYG{o}{\symbol{92}in}% Server, m %\PYG{o}{\symbol{92}in}% messages : %\PYG{n+nf}{HandleVoteRequest}%(i, m)
    %\PYG{o}{\symbol{92}/}% %\PYG{o}{\symbol{92}E}% i %\PYG{o}{\symbol{92}in}% Server, m %\PYG{o}{\symbol{92}in}% messages : %\PYG{n+nf}{HandleVoteResponse}%(i, m)
    %\PYG{o}{\symbol{92}/}% %\PYG{o}{\symbol{92}E}% i %\PYG{o}{\symbol{92}in}% Server, v %\PYG{o}{\symbol{92}in}% Value : %\PYG{n+nf}{ClientRequest}%(i, v)
    %\PYG{o}{\symbol{92}/}% %\PYG{o}{\symbol{92}E}% i %\PYG{o}{\symbol{92}in}% Server, m %\PYG{o}{\symbol{92}in}% messages : %\PYG{n+nf}{HandleAppendEntries}%(i, m)
    %\PYG{o}{\symbol{92}/}% %\PYG{o}{\symbol{92}E}% i %\PYG{o}{\symbol{92}in}% Server, m %\PYG{o}{\symbol{92}in}% messages : %\PYG{n+nf}{HandleAppendResponse}%(i, m)
    %\PYG{o}{\symbol{92}/}% %\PYG{o}{\symbol{92}E}% i %\PYG{o}{\symbol{92}in}% Server : %\PYG{n+nf}{SendHeartbeat}%(i)
    %\PYG{o}{\symbol{92}/}% %\PYG{o}{\symbol{92}E}% i %\PYG{o}{\symbol{92}in}% Server, m %\PYG{o}{\symbol{92}in}% messages : %\PYG{n+nf}{HandleHeartbeat}%(i, m)
    %\PYG{o}{\symbol{92}/}% %\PYG{o}{\symbol{92}E}% i %\PYG{o}{\symbol{92}in}% Server : %\PYG{n+nf}{AdvanceElectionTimeout}%(i)

Fairness %\PYG{o}{==}%
    %\PYG{o}{/\symbol{92}}% %\PYG{o}{\symbol{92}A}% i %\PYG{o}{\symbol{92}in}% Server : %\PYG{n+nf}{WF\_vars}%(%\PYG{n+nf}{Timeout}%(i))
    %\PYG{o}{/\symbol{92}}% %\PYG{o}{\symbol{92}A}% i %\PYG{o}{\symbol{92}in}% Server : %\PYG{n+nf}{WF\_vars}%(%\PYG{n+nf}{StartPreVote}%(i))
    %\PYG{o}{/\symbol{92}}% %\PYG{o}{\symbol{92}A}% i %\PYG{o}{\symbol{92}in}% Server : %\PYG{n+nf}{WF\_vars}%(%\PYG{o}{\symbol{92}E}% m %\PYG{o}{\symbol{92}in}% messages : %\PYG{n+nf}{HandlePreVoteResponse}%(i, m))
    %\PYG{o}{/\symbol{92}}% %\PYG{o}{\symbol{92}A}% i %\PYG{o}{\symbol{92}in}% Server : %\PYG{n+nf}{WF\_vars}%(%\PYG{o}{\symbol{92}E}% m %\PYG{o}{\symbol{92}in}% messages : %\PYG{n+nf}{HandleVoteRequest}%(i, m))
    %\PYG{o}{/\symbol{92}}% %\PYG{o}{\symbol{92}A}% i %\PYG{o}{\symbol{92}in}% Server : %\PYG{n+nf}{WF\_vars}%(%\PYG{o}{\symbol{92}E}% m %\PYG{o}{\symbol{92}in}% messages : %\PYG{n+nf}{HandleVoteResponse}%(i, m))
    %\PYG{o}{/\symbol{92}}% %\PYG{o}{\symbol{92}A}% i %\PYG{o}{\symbol{92}in}% Server : %\PYG{n+nf}{WF\_vars}%(%\PYG{o}{\symbol{92}E}% m %\PYG{o}{\symbol{92}in}% messages : %\PYG{n+nf}{HandleAppendEntries}%(i, m))
    %\PYG{o}{/\symbol{92}}% %\PYG{o}{\symbol{92}A}% i %\PYG{o}{\symbol{92}in}% Server : %\PYG{n+nf}{WF\_vars}%(%\PYG{o}{\symbol{92}E}% m %\PYG{o}{\symbol{92}in}% messages : %\PYG{n+nf}{HandleAppendResponse}%(i, m))
    %\PYG{o}{/\symbol{92}}% %\PYG{o}{\symbol{92}A}% i %\PYG{o}{\symbol{92}in}% Server : %\PYG{n+nf}{WF\_vars}%(%\PYG{n+nf}{SendHeartbeat}%(i))
    %\PYG{o}{/\symbol{92}}% %\PYG{o}{\symbol{92}A}% i %\PYG{o}{\symbol{92}in}% Server : %\PYG{n+nf}{WF\_vars}%(%\PYG{o}{\symbol{92}E}% m %\PYG{o}{\symbol{92}in}% messages : %\PYG{n+nf}{HandleHeartbeat}%(i, m))
    %\PYG{o}{/\symbol{92}}% %\PYG{o}{\symbol{92}A}% i %\PYG{o}{\symbol{92}in}% Server : %\PYG{n+nf}{WF\_vars}%(%\PYG{n+nf}{AdvanceElectionTimeout}%(i))

Spec %\PYG{o}{==}% Init %\PYG{o}{/\symbol{92}}% %\PYG{o}{[}%%\PYG{o}{]}%%\PYG{o}{[}%Next%\PYG{o}{]}%_vars %\PYG{o}{/\symbol{92}}% Fairness

%\PYG{o}{==}%%\PYG{o}{==}%
\end{minted}

    \caption{Etcd Raft \tla{} model generated by the code translation agent with Claude-Sonnet-4.}
    \vspace{15pt}
    \label{fig:app-etcd-specula}
\end{longlisting}

\begin{longlisting}
    \begin{minted}[escapeinside=\%\%,breaklines,breakanywhere,breakautoindent]{text}
%\PYG{k}{SPECIFICATION}% Spec

%\PYG{k}{CONSTANTS}%
    Server %\PYG{o}{=}% %\PYG{o}{\{}%s1, s2, s3%\PYG{o}{\}}%
    Value %\PYG{o}{=}% %\PYG{o}{\{}%v1, v2%\PYG{o}{\}}%
    Nil %\PYG{o}{=}% %\PYG{l+s}{"Nil"}%
    Follower %\PYG{o}{=}% %\PYG{l+s}{"Follower"}%
    PreCandidate %\PYG{o}{=}% %\PYG{l+s}{"PreCandidate"}%
    Candidate %\PYG{o}{=}% %\PYG{l+s}{"Candidate"}%
    Leader %\PYG{o}{=}% %\PYG{l+s}{"Leader"}%
\end{minted}

    \caption{Etcd Raft TLC configuration generated by the code translation agent with Claude-Sonnet-4.}
    \vspace{5pt}
    \label{fig:app-etcd-specula-cfg}
\end{longlisting}

\section{PGo-compiled Systems}
\label{sec:pgo}

Table~\ref{tab:systems} lists all the system artifacts in \bench{}.
Unlike other open-source systems implemented mostly by human developers,
    PGo systems represent a special kind of compiler-generated systems.
PGo is a compiler converting distributed systems specifications written in a DSL of \tla{} into executable systems implementations in Go~\citep{hackett2023pgo}.

These systems reflect production use cases:
\begin{packed_itemize}
\item dqueue is a simple distributed queue with producers and consumers, which represents a common cloud computing mechanism. Similar distributed queues are available from many cloud platforms, like Amazon SQS, Cloudflare Queues, or Apache Kafka.
\item locksvc is a simple distributed locking system, which represents a common distributed systems concept. 
\item raftkvs is a verified distributed key-value store, with competitive performance.
For its consensus implementation, raftkvs specifies Raft~\citep{ongaro2014search}. 
\end{packed_itemize}

These systems are complex, each requiring several person-days of effort to specify. The raftkvs store is particularly complex, requiring almost a person-month of effort. 
While they are developed using a formal modeling language, these systems also account for practical coding concerns.
Each system compiles to usable, non-trivial Go code.
Notably, raftkvs outperforms other formally verified key-value stores, with 41\% higher throughput than the next-best formally verified store implementation, and similar latency but 21\% of the throughput achieved by Etcd.

A challenge unique to system modeling is that PGo-compiled systems contain machine-generated Go code, which includes unusual abstractions and coding patterns.
For instance, the generated code makes extensive use of abstractions from PGo's runtime support library, while containing many synthetically named variables.
These issues are representative of realistic engineering scenarios, such as generated code (macros, parser generators, state machines), or situations where the original source code is lost (decompilation artifacts).
This type of source code input currently leads to poor performance on our benchmarks.

\textbf{PGo Trace Validation.} %
For AI-generated system models,
    we must validate their behavior against gathered execution traces.
PGo's TraceLink feature provides a different trace validation method than for hand-written systems, 
    allowing for automatic implementation tracing and \tla{} glue generation.
As a result, no additional work is needed to gather traces.
For simplicity, we use traces taken from TraceLink's published artifact.
From these traces, TraceLink is able to generate its own binding \tla{}, mapping these logs precisely to a \tla{} state space.

\vspace{-4pt}
\section{\bench{} Evaluation Prompts}

During \bench{} evaluation, LLMs are invoked for conformance and invariant correctness evaluation to extract information, 
    map actions and variables, and concretize invariants based on invariant templates.

We show the complete prompts used for benchmark evaluation.
These prompts are templates with parameterized fields that are instantiated 
    by task-specific information.
For demonstration, we instantiate the fields in the task for modeling Etcd Raft, 
    with the instantiated parts marked in green.

\subsection{Conformance Evaluation Prompts}
Two prompts extract model information and map model action and variable names to code, both generating configuration files for script processing to support trace validation.

\para{Model component extraction.}
This prompt directs an LLM to extract \tla{} model components, such as constants, variables, and actions, which are used by a script to generate a \emph{trace specification}~\citep{cirstea2024trace}. A trace specification constrains state space exploration along the code trace path to verify whether a model state space path matches the code trace.
In the prompt, the \code{\{source\_code\}} field is instantiated with the \tla{} model.

\begin{promptboxwithheader}{Model Component Extraction Prompt}
    Generate a YAML configuration file from the provided TLA+ model (.tla) and configuration (.cfg) files. Extract information according to the following rules: \\
 \\
\#\# Task Description \\
Parse the TLA+ model and configuration files to create a structured YAML configuration that captures the model name, constants, variables, actions, and interactions. \\
 \\
\#\# Extraction Rules \\
 \\
\#\#\# spec\_name \\
Extract from the module declaration line: \textasciigrave{}\symbol{45}\symbol{45}\symbol{45}\symbol{45} MODULE \textless{}ModuleName\textgreater{} \symbol{45}\symbol{45}\symbol{45}\symbol{45}\textasciigrave{} \\
The spec\_name is the ModuleName between "\symbol{45}\symbol{45}\symbol{45}\symbol{45} MODULE" and "\symbol{45}\symbol{45}\symbol{45}\symbol{45}". \\
 \\
\#\#\# constants \\
Extract from the CONSTANTS section in the .cfg file. \\
\symbol{45} name: The constant identifier \\
\symbol{45} value: The assigned value, formatted as: \\
  \symbol{45} Sets: Wrap in single quotes, e.g., '\{s1, s2, s3\}' becomes '\{"s1", "s2", "s3"\}' \\
  \symbol{45} Strings: Wrap in single quotes with double quotes inside, e.g., Nil becomes '"Nil"' \\
  \symbol{45} Numbers: Wrap in single quotes as string, e.g., 5 becomes '5' \\
 \\
\#\#\# variables \\
Extract from the Init operator definition in the .tla file. \\
For each variable assignment in Init: \\
\symbol{45} name: The variable name \\
\symbol{45} default\_value: The initial value expression (preserve TLA+ syntax, escape backslashes) \\
 \\
\#\#\# actions \\
Extract from the Next operator definition. Include only direct action calls (not numbered interactions). \\
For each action: \\
\symbol{45} name: The action/operator name \\
\symbol{45} parameters: List of parameters with: \\
  \symbol{45} name: Parameter variable name \\
  \symbol{45} source: Where the parameter comes from (e.g., Server, messages) \\
\symbol{45} stmt: The complete statement as it appears in Next (including any conditions) \\
 \\
\#\#\# interactions \\
Extract from the Next operator definition. Include only numbered intermediate actions. \\
Just list the names (e.g., HandletickElection\_1, HandletickHeartbeat\_1) \\
 \\
\#\# Example \\
Given this TLA+ model: \\
 \\
\symbol{45}\symbol{45}\symbol{45}\symbol{45} MODULE SimpleSpec \symbol{45}\symbol{45}\symbol{45}\symbol{45} \\
... \\
Init == \\
/\symbol{92} x = 0 \\
/\symbol{92} y = [s \symbol{92}in Server |\symbol{45}\textgreater{} 0] \\
 \\
Next == \\
\symbol{92}/ \symbol{92}E s \symbol{92}in Server : Action1(s) \\
\symbol{92}/ \symbol{92}E m \symbol{92}in messages : Action2(m) \\
\symbol{92}/ IntermediateAction\_1 \\
 \\
And this configuration: \\
 \\
CONSTANTS \\
Server = \{s1, s2\} \\
MaxValue = 10 \\
 \\
Generate this YAML: \\
 \\
\textasciigrave{}\textasciigrave{}\textasciigrave{}yaml \\
spec\_name: SimpleSpec \\
constants: \\
\symbol{45} name: Server \\
  value: '\{"s1", "s2"\}' \\
\symbol{45} name: MaxValue \\
  value: '10' \\
variables: \\
\symbol{45} name: x \\
  default\_value: '0' \\
\symbol{45} name: y \\
  default\_value: '[s \symbol{92}\symbol{92}in Server |\symbol{45}\textgreater{} 0]' \\
actions: \\
\symbol{45} name: Action1 \\
  parameters: \\
  \symbol{45} name: s \\
    source: Server \\
  stmt: Action1(s) \\
\symbol{45} name: Action2 \\
  parameters: \\
  \symbol{45} name: m \\
    source: messages \\
  stmt: Action2(m) \\
interactions: \\
\symbol{45} name: IntermediateAction\_1 \\
\textasciigrave{}\textasciigrave{}\textasciigrave{} \\
 \\
\#\# Important Notes \\
1. Return ONLY the YAML content \symbol{45} no explanations, comments, or natural language \\
2. Preserve TLA+ syntax exactly in default\_value fields (escape backslashes) \\
3. For actions with conditions, include the full stmt as it appears in Next \\
4. Ignore variables that appear in Init but are not part of the main model (e.g., pc, info, stack) \\
5. Order matters: spec\_name, constants, variables, actions, interactions \\
 \\
Generate the YAML configuration based on the provided TLA+ files: \\
 \\
\{source\_code\}
\end{promptboxwithheader}

\para{Model and code component mapping.}
This prompt directs an LLM to map code to model variable naming for trace validation comparison.
The LLM generates a JSON file storing the mappings, which is further processed by a script to output a test harness that aligns code trace variable and action names with the model.
In the prompt, the \code{\{TLA\_SPEC\_CODE\_PLACEHOLDER\}} field is instantiated with the \tla{} model, and \code{\{IMPLEMENTATION\_CODE\_PLACEHOLDER\}} with the corresponding code.

\begin{promptboxwithheader}{Model and Code Component Mapping Prompt}
    You are tasked with generating a JSON mapping file that defines how to convert a concurrent or distributed system traces to TLA+ model format for trace validation. \\
 \\
\#\# System Overview \\
 \\
\promptgreen{}etcd Raft is a distributed consensus algorithm implementation that supports: \\
\symbol{45} Leader election with terms and prevoting/voting \\
\symbol{45} Log replication across multiple nodes \\
\symbol{45} State transitions between Follower, Candidate, and Leader roles \\
\symbol{45} Message passing between nodes \\
\bla{} \\
\#\# Code Analysis \\
Before generating the mapping, you need to analyze the relevant code to understand the system behavior: \\
 \\
**CRITICAL**: You MUST base your mapping on the actual TLA+ model content, NOT on the examples below. The examples are for format reference only. Always use the actual variables and actions defined in the provided model. \\
 \\
\#\#\# TLA+ Model Code \\
\textasciigrave{}\textasciigrave{}\textasciigrave{}tla+ \\
\{TLA\_SPEC\_CODE\_PLACEHOLDER\} \\
\textasciigrave{}\textasciigrave{}\textasciigrave{} \\
 \\
\#\#\# Implementation Code \\
\textasciigrave{}\textasciigrave{}\textasciigrave{}go \\
\{IMPLEMENTATION\_CODE\_PLACEHOLDER\} \\
\textasciigrave{}\textasciigrave{}\textasciigrave{} \\
 \\
\#\# Input: System Trace Format \\
 \\
\promptgreen{}System traces are in JSONL format with events like: \\
\textasciigrave{}\textasciigrave{}\textasciigrave{}json \\
\{"conf": [["1", "2", "3"], []], "log": 0, "name": "InitState", "nid": "1", "role": "StateFollower", "state": \{"commit": 0, "term": 0, "vote": "0"\}\} \\
\{"conf": [["1", "2", "3"], []], "log": 1, "name": "BecomeCandidate", "nid": "1", "role": "StateCandidate", "state": \{"commit": 0, "term": 1, "vote": "0"\}\} \\
\{"conf": [["1", "2", "3"], []], "log": 1, "name": "BecomeCandidate", "nid": "2", "role": "StateCandidate", "state": \{"commit": 0, "term": 1, "vote": "0"\}\} \\
\textasciigrave{}\textasciigrave{}\textasciigrave{} \\
 \\
Common actions in system traces: \\
\symbol{45} BecomeFollower: Transition to follower role \\
\symbol{45} BecomeCandidate: Transition to candidate role \\
\symbol{45} BecomeLeader: Transition to leader role \\
\symbol{45} Ready: Node is ready for operations \\
\symbol{45} PreVote/Vote: Cast prevote/vote during election \\
\symbol{45} AppendEntries: Replicate log entries \\
\symbol{45} Heartbeat: Send/receive heartbeat messages \\
\bla{} \\
\#\# Target: TLA+ Model Variables \\
 \\
\promptgreen{}The TLA+ model tracks these state variables: \\
\symbol{45} currentTerm: Current term number for each node \\
\symbol{45} state: Node role (Follower, Candidate, Leader) \\
\symbol{45} votedFor: Which candidate this node voted for in current term \\
\symbol{45} commitIndex: Index of highest log entry known to be committed \\
\symbol{45} nextIndex: For leaders, next log entry to send to each server \\
\symbol{45} matchIndex: For leaders, highest log entry known to be replicated on server \\
\bla{} \\
\#\# Required Mapping Structure \\
 \\
\promptgreen{}Generate a JSON file with this structure: \\
 \\
\textasciigrave{}\textasciigrave{}\textasciigrave{}json \\
\{ \\
  "config": \{ \\
    "Server": ["Server1", "Server2", "Server3"]  // List of node identifiers \\
  \}, \\
  "events": \{ \\
    // Map system actions to TLA+ events \\
    "InitState": "Init", \\
    "BecomeFollower": "BecomeFollower", \\
    "BecomeCandidate": "BecomeCandidate", \\
    "BecomeLeader": "BecomeLeader", \\
    "Ready": "Ready", \\
    "Vote": "Vote", \\
    "AppendEntries": "AppendEntries", \\
    "Heartbeat": "Heartbeat", \\
    // Add other mappings as needed based on code analysis \\
  \}, \\
  "node\_mapping": \{ \\
    // Map string node IDs to node names \\
    "1": "Node1", \\
    "2": "Node2", \\
    "3": "Node3", \\
    // Continue as needed \\
  \}, \\
  "role\_mapping": \{ \\
    // Map system roles to TLA+ states \\
    "StateFollower": "Follower", \\
    "StateCandidate": "Candidate", \\
    "StateLeader": "Leader" \\
  \} \\
\} \\
\textasciigrave{}\textasciigrave{}\textasciigrave{} \\
\bla{} \\
\#\# Implementation Notes \\
1. The mapping will be used by a state tracker that maintains complete system state \\
2. Server IDs in traces are numeric (0, 1, 2...) and must be mapped to "Server1", "Server2", etc. \\
3. The state tracker will automatically handle state transitions based on actions \\
4. Focus on correctly mapping actions and Server states \\
5. The config section should list all possible Server that might appear in traces \\
 \\
\#\# Your Task \\
Generate a complete mapping.json file that: \\
1. Maps all common actions to their TLA+ equivalents \\
2. Provides server ID mappings for all servers that appear in traces \\
3. Ensures compatibility with the state tracking implementation
\end{promptboxwithheader}

\subsection{Invariant Correctness Evaluation Prompt}

This prompt concretizes invariants from given invariant templates to model-specific forms.
It typically requires the LLM to map different names in an invariant template to the corresponding model elements.
The \code{\$tla\_model} field is instantiated with the \tla{} model, and the \code{\$invariant\_templates} field with the invariant templates defined in the system artifact (see \S\ref{sec:inv} for an example).

\begin{promptboxwithheader}{Invariant Concretization Prompt}
    You are a TLA+ expert specializing in \promptgreen{}distributed systems and Raft consensus\bla{}. Your task is to implement a set of expert\symbol{45}written invariants for the given \promptgreen{}etcd\bla{} TLA+ model. \\
 \\
\#\# Target Model \\
 \\
\$tla\_model \\
 \\
\#\# Invariants to Implement \\
 \\
\$invariant\_templates \\
 \\
\#\# Implementation Requirements \\
 \\
\promptgreen{}1. **Deep Analysis**: First, thoroughly understand both the invariant template's semantic intent and the model's modeling approach: \\
   \symbol{45} What distributed consensus property does each template aim to verify? \\
   \symbol{45} How does the model represent server states, logs, terms, and leadership? \\
   \symbol{45} What are the semantic equivalents between template concepts and model implementation? \\
 \\
2. **Semantic Mapping**: For each invariant, identify the conceptual mapping between template and model: \\
   \symbol{45} Template server state concepts \symbol{45}\textgreater{} Model's server state representation \\
   \symbol{45} Template log structure \symbol{45}\textgreater{} Model's log data structures and indexing \\
   \symbol{45} Template leadership concepts \symbol{45}\textgreater{} Model's leader election and term management \\
   \symbol{45} Template node/server sets \symbol{45}\textgreater{} Model's server constants and domains \\
\bla{} \\
3. **Creative Adaptation**: Translate the invariant while preserving its core safety/liveness meaning: \\
   \symbol{45} **DO NOT** simply replace variable names \symbol{45} understand the underlying distributed systems logic \\
   \symbol{45} **DO** redesign the predicate logic to fit the model's data structure granularity \\
   \symbol{45} **DO** use equivalent semantic concepts even if data representations differ \\
   \symbol{45} **PRESERVE** the original safety/liveness guarantees without weakening the property \\
 \\
4. **TLA+ Property Type Constraints**: \\
    \\
   **FOR SAFETY PROPERTIES** (type: "safety"): \\
   \symbol{45} **MUST** be STATE PREDICATES (describe single states only) \\
   \symbol{45} **NEVER** use primed variables (\textasciigrave{}currentTerm'\textasciigrave{}, \textasciigrave{}log'\textasciigrave{}) \\
   \symbol{45} **NEVER** use temporal operators (\textasciigrave{}[]\textasciigrave{}, \textasciigrave{}\textless{}\textgreater{}\textasciigrave{}, \textasciigrave{}\textasciitilde{}\textgreater{}\textasciigrave{}) \\
   \symbol{45} **NEVER** reference actions (like \textasciigrave{}RequestVote(s)\textasciigrave{}, \textasciigrave{}AppendEntries(s,t)\textasciigrave{}) \symbol{45} only use state variables \\
   \symbol{45} **ONLY** use unprimed variables (\textasciigrave{}currentTerm[s]\textasciigrave{}, \textasciigrave{}log[s]\textasciigrave{}) and constants \\
   \symbol{45} **CORRECT**: \textasciigrave{}LeaderUniqueness == \symbol{92}A term \symbol{92}in Terms : Cardinality(\{s \symbol{92}in Servers : state[s] = "leader" /\symbol{92} currentTerm[s] = term\}) \textless{}= 1\textasciigrave{}  \\
   \symbol{45} **INCORRECT**: \textasciigrave{}state[s] = "candidate" =\textgreater{} RequestVote(s)\textasciigrave{}  (references action RequestVote) \\
 \\
   **FOR LIVENESS PROPERTIES** (type: "liveness"):  \\
   \symbol{45} **MUST** be TEMPORAL FORMULAS (describe execution traces) \\
   \symbol{45} **MUST** use temporal operators (\textasciigrave{}\textless{}\textgreater{}\textasciigrave{}, \textasciigrave{}\textasciitilde{}\textgreater{}\textasciigrave{}) to express "eventually" or "leads\symbol{45}to" \\
   \symbol{45} **CORRECT**: \textasciigrave{}EventualLeaderElection == \textless{}\textgreater{}(\symbol{92}E s \symbol{92}in Servers : state[s] = "leader")\textasciigrave{}  \\
 \\
5. **Constraint Compliance**: \\
   \symbol{45} Use ONLY variables, constants, and operators that exist in the model \\
   \symbol{45} Generate complete, syntactically valid TLA+ invariant definitions \\
   \symbol{45} Maintain the exact invariant names from templates \\
 \\
6. **Output format**: Return a JSON object containing an array of complete TLA+ invariant definitions \\
 \\
7. **EXACT naming requirement**: You MUST use the exact invariant names specified in the templates above. Do not create your own names. \\
 \\
\#\# Example Output Format \\
 \\
\promptgreen{}\textasciigrave{}\textasciigrave{}\textasciigrave{}json \\
\{ \\
  "invariants": [ \\
    "LeaderUniqueness == \symbol{92}\symbol{92}A term \symbol{92}\symbol{92}in 1..MaxTerm : Cardinality(\{n \symbol{92}\symbol{92}in Servers : state[n].role = \symbol{92}"leader\symbol{92}" /\symbol{92}\symbol{92} state[n].currentTerm = term\}) \textless{}= 1", \\
    "LogConsistency == \symbol{92}\symbol{92}A n1, n2 \symbol{92}\symbol{92}in Servers : \symbol{92}\symbol{92}A i \symbol{92}\symbol{92}in DOMAIN log[n1] : (i \symbol{92}\symbol{92}in DOMAIN log[n2] /\symbol{92}\symbol{92} log[n1][i].term = log[n2][i].term) =\textgreater{} (\symbol{92}\symbol{92}A j \symbol{92}\symbol{92}in 1..i : log[n1][j] = log[n2][j])" \\
  ] \\
\} \\
\textasciigrave{}\textasciigrave{}\textasciigrave{} \\
\bla{} \\
**CRITICAL REQUIREMENTS**:  \\
\symbol{45} **SEMANTIC PRESERVATION**: Each translated invariant MUST verify the same property as the original template \\
\symbol{45} **CREATIVE ADAPTATION**: Do NOT simply omit invariants \symbol{45} find creative ways to express the same property using available model elements \\
\symbol{45} **COMPLETENESS**: Aim to translate ALL invariants by understanding their semantic intent, not just their syntactic form \\
\symbol{45} Use ONLY variables, constants, and operators that exist in the provided model \\
\symbol{45} Use EXACTLY the invariant names from the templates (preserve exact names for evaluation consistency) \\
\symbol{45} Return ONLY valid JSON, no explanatory text before or after \\
\symbol{45} Each array element must be a complete TLA+ invariant definition: "InvariantName == \textless{}expression\textgreater{}" \\
\symbol{45} For complex invariants, you may use multiline format within the JSON string (use actual line breaks) \\
\symbol{45} For simple invariants, single line format is preferred \\
\symbol{45} **LAST RESORT**: Only omit an invariant if its core concept is fundamentally incompatible with the model's design \\
\symbol{45} **CRITICAL JSON ESCAPING RULES**:  \\
  \symbol{45} TLA+ operators like \textasciigrave{}\symbol{92}A\textasciigrave{}, \textasciigrave{}\symbol{92}E\textasciigrave{}, \textasciigrave{}\symbol{92}in\textasciigrave{} contain ONE backslash in the final TLA+ code \\
  \symbol{45} In JSON strings, use EXACTLY ONE backslash escape: write \textasciigrave{}"\symbol{92}\symbol{92}A"\textasciigrave{} to get \textasciigrave{}\symbol{92}A\textasciigrave{} in TLA+ \\
  \symbol{45} **DO NOT double\symbol{45}escape**: \textasciigrave{}"\symbol{92}\symbol{92}\symbol{92}\symbol{92}A"\textasciigrave{} is WRONG and will produce \textasciigrave{}\symbol{92}\symbol{92}A\textasciigrave{} in TLA+ \\
  \symbol{45} **CORRECT**: \textasciigrave{}"LeaderUniqueness == \symbol{92}\symbol{92}A term \symbol{92}\symbol{92}in 1..MaxTerm : state[term] = \symbol{92}"leader\symbol{92}""\textasciigrave{} \\
  \symbol{45} **WRONG**: \textasciigrave{}"LeaderUniqueness == \symbol{92}\symbol{92}\symbol{92}\symbol{92}A term \symbol{92}\symbol{92}\symbol{92}\symbol{92}in 1..MaxTerm : state[term] = \symbol{92}"leader\symbol{92}""\textasciigrave{} \\
\symbol{45} Start your response immediately with the opening brace \{
\end{promptboxwithheader}

\section{Basic Modeling Agent}
\label{sec:model_agent}

The basic modeling agent operates in two steps for each system artifact:
    (1) generating the model, including both the \tla{} model and its TLC configuration,
    and (2) using a feedback loop that takes \bench{} evaluation results 
    to iteratively improve the generated \tla{} model.
We show the complete prompts of the basic modeling agent (\S\ref{sec:setup}) to provide its detailed implementation.

\subsection{Model Generation Prompts}
\label{app:basic-agent-prompt}

\para{\tla{} model generation.}
This prompt directs an LLM to generate the \tla{} model file, instantiated with the granularity definitions of the system artifact (see \S\ref{sec:task}).
The \code{\{file\_path\}} and \code{\{source\_code\}} fields are instantiated with the code file path in the repository and source code content, respectively.

\begin{promptboxwithheader}{\tla{} Model Generation Prompt}
    You are an expert in formal verification and TLA+ models with deep expertise in concurrent and distributed systems, particularly \promptgreen{}etcd and Raft consensus\bla{} \\
. \\
 \\
Convert the following source code to a comprehensive TLA+ model. \\
 \\
System: \promptgreen{}etcd distributed key\symbol{45}value store\bla{} \\
 \\
Source Code from \{file\_path\}: \\
\textasciigrave{}\textasciigrave{}\textasciigrave{}go \\
\{source\_code\} \\
\textasciigrave{}\textasciigrave{}\textasciigrave{} \\
 \\
System\symbol{45}specific modeling requirements: \\
 \\
\promptgreen{}MANDATORY CORE ACTIONS (must include all): \\
1. [Message Types] MsgHup (election timeout), MsgVote/MsgVoteResp (voting), MsgApp/MsgAppResp (log replication)   \\
2. [Node States] Four states: StateFollower, StateCandidate, StateLeader, StatePreCandidate (prevote enabled) \\
3. [Leader Election] Complete prevote + vote phases: PreCandidate $\rightarrow$ Candidate $\rightarrow$ Leader transitions \\
4. [Log Operations] Log entry appending, consistency checks, commitment with majority quorum \\
5. [Heartbeat/Timeout] Election timeouts triggering campaigns, heartbeat prevention of elections \\
6. [Client Proposals] MsgProp message handling and log entry creation by leaders \\
 \\
EXPLICITLY EXCLUDED (do not model): \\
\symbol{45} Configuration changes and joint consensus (ConfChange messages)   \\
\symbol{45} Log compaction and snapshots (MsgSnap) \\
\symbol{45} ReadIndex optimizations (MsgReadIndex)  \\
\symbol{45} Async storage operations (LocalAppendThread, LocalApplyThread) \\
\symbol{45} Advanced flow control and progress tracking details \\
 \\
REQUIRED BEHAVIORAL SCOPE: \\
\symbol{45} Prevote phase (StatePreCandidate) must be modeled as it's enabled by default in etcd \\
\symbol{45} State transition constraints: Follower $\rightarrow$ PreCandidate $\rightarrow$ Candidate $\rightarrow$ Leader (strict transitions) \\
\symbol{45} Message processing by state: only valid message types handled in each node state \\
\symbol{45} Term advancement rules: nodes advance term when receiving messages with higher term \\
\symbol{45} Voting restrictions: one vote per term, term must be current or newer \\
\symbol{45} Heartbeat mechanism: leaders send heartbeats, followers reset election timeout on receipt \\
\symbol{45} Log consistency checks: prevLogIndex/prevLogTerm validation in MsgApp processing \\
\symbol{45} Majority\symbol{45}based leader election and log commitment \\
\symbol{45} Basic network message delays and losses \\
\bla{} \\
Generate a TLA+ model that accurately models the system's behavior. \\
 \\
CRITICAL OUTPUT REQUIREMENTS: \\
1. The MODULE name must be exactly "\promptgreen{}etcdraft\bla{} \\
" (\symbol{45}\symbol{45}\symbol{45}\symbol{45} MODULE \promptgreen{}etcdraft\bla{} \symbol{45}\symbol{45}\symbol{45}\symbol{45}) \\
 \\
2. Return ONLY pure TLA+ model code \symbol{45} no markdown code blocks (no \textasciigrave{}\textasciigrave{}\textasciigrave{}tla or \textasciigrave{}\textasciigrave{}\textasciigrave{}) \\
3. Do not include any explanations, comments, or formatting markers \\
4. Start your response directly with: \symbol{45}\symbol{45}\symbol{45}\symbol{45} MODULE \promptgreen{}etcdraft\bla{} \\
 \symbol{45}\symbol{45}\symbol{45}\symbol{45} \\
5. End your response with the closing ==== \\
6. **DO NOT define invariants** (like MutualExclusion, Invariant, etc.) \symbol{45} focus on modeling the system behavior \\
7. **MUST include EXTENDS statement**: The model must extend at least these modules: TLC, Sequences, SequencesExt, Naturals, FiniteSets, Bags
\end{promptboxwithheader}

\para{TLC configuration generation.}
This prompt directs an LLM to generate a TLC configuration file.
The configuration file requires the LLM's understanding of the system to make the model executable, such as designating the initial predicate and next-state relations.
The \code{\$tla\_spec} field is instantiated with the \tla{} model generated in the previous step.

\begin{promptboxwithheader}{TLC Configuration Generation Prompt}
    You are a TLA+ expert. Generate a complete TLC configuration file (.cfg) for the \promptgreen{}etcd\bla{} model that can be directly saved and used for model checking. \\
 \\
\#\# Input Model: \\
 \\
\$tla\_spec \\
 \\
\#\# Requirements: \\
 \\
1. **Analyze the model** to identify the main model name and all declared constants \\
2. **Generate complete .cfg file content** with SPECIFICATION, CONSTANTS sections \\
3. **Use small values for constants** to ensure efficient model checking (2\symbol{45}3 servers, small integers) \\
4. **Output ONLY the raw .cfg file content** \symbol{45} no explanations, no markdown, no code blocks \\
 \\
\#\# Example Output Format: \\
 \\
SPECIFICATION SpecName \\
 \\
CONSTANTS \\
    ... \\
 \\
**CRITICAL: Your response must contain exactly ONE complete .cfg file. Do not repeat any sections. Start your response immediately with "SPECIFICATION" and include nothing else.**
\end{promptboxwithheader}

\subsection{Model Refinement Prompt}

This prompt provides guidance for the LLM to refine the previously generated model using syntax and runtime evaluation results from \bench{}.
The \code{\{current\_model\}} field contains the previous iteration's model, \code{\{current\_tlc\_cfg\}} contains the previous TLC configuration,
\code{\{syntax\_errors\}} contains the syntax errors reported by SANY, and \code{\{runtime\_errors\}} contains the runtime errors reported by TLC.

\begin{promptboxwithheader}{Model Refinement Prompt}
    You are an expert TLA+ model specialist with extensive experience in concurrent and distributed systems modeling. \\
 \\
I need you to fix errors in a TLA+ model for \promptgreen{}etcdraft\bla{} system. \\
 \\
\#\# Current TLA+ Model \\
\textasciigrave{}\textasciigrave{}\textasciigrave{}tla \\
\{current\_model\} \\
\textasciigrave{}\textasciigrave{}\textasciigrave{} \\
 \\
\#\# Current TLC Configuration \\
\textasciigrave{}\textasciigrave{}\textasciigrave{} \\
\{current\_tlc\_cfg\} \\
\textasciigrave{}\textasciigrave{}\textasciigrave{} \\
 \\
\#\# Errors Found \\
 \\
**Detailed Syntax Errors:** \\
\{syntax\_errors\} \\
 \\
**Detailed Runtime Errors:** \\
\{runtime\_errors\} \\
 \\
\#\# Correction Instructions \\
This is correction attempt \{attempt\_number\} of \{max\_attempts\}. \\
 \\
Please provide a corrected TLA+ model that fixes these errors. Your corrected model should: \\
 \\
1. **Fix all syntax errors**: Ensure proper TLA+ syntax, correct operator usage, and valid module structure \\
2. **Resolve runtime errors**: Define missing variables, operators, and ensure logical consistency \\
3. **Maintain original intent**: Keep the core distributed system logic and behavior from the source code \\
4. **Follow TLA+ best practices**: Use appropriate data structures, actions, and invariants \\
5. **Be complete and self\symbol{45}contained**: Include all necessary EXTENDS, CONSTANTS, VARIABLES, and operator definitions \\
 \\
Focus specifically on: \\
\symbol{45} Defining any missing variables or constants \\
\symbol{45} Implementing missing operators or functions \\
\symbol{45} Fixing syntax issues with operators, expressions, or module structure \\
\symbol{45} Ensuring proper action definitions and state transitions \\
\symbol{45} Maintaining consistency with \promptgreen{}etcdraft\bla{}'s system behavior \\
 \\
**CRITICAL OUTPUT REQUIREMENTS:** \\
\symbol{45} Return ONLY pure TLA+ model code \\
\symbol{45} NO markdown code blocks (no \textasciigrave{}\textasciigrave{}\textasciigrave{}tla or \textasciigrave{}\textasciigrave{}\textasciigrave{})   \\
\symbol{45} NO explanations, comments, or text outside the model \\
\symbol{45} NO formatting markers of any kind \\
\symbol{45} The MODULE name must be exactly "\promptgreen{}etcdraft\bla{}" \\
\symbol{45} Start directly with: \symbol{45}\symbol{45}\symbol{45}\symbol{45} MODULE \promptgreen{}etcdraft\bla{} \symbol{45}\symbol{45}\symbol{45}\symbol{45}
\end{promptboxwithheader}

\section{Trace Learning Agent}
\label{sec:trace_agent}

The trace learning agent does not use any code as input; instead, 
it relies on the distributed traces as context. 
Similar to the basic modeling agent, we provide an initial prompt analogous to the basic modeling agent's prompt (\S\ref{app:basic-agent-prompt}), but substituting the codebase context with trace information instead. If the first model generation fails to pass compilation, the model refinement loop will pass the errors back to the LLM to iteratively fix the model.

\textbf{Trace formats.} The trace-based method works with several types of traces and can be easily extended to additional systems. For each trace format, we provide a short custom prompt explaining the format. We currently support:
\begin{packed_itemize}
    \item \texttt{.ndjson} and \texttt{.jsonl} logs: Newline-delimited JSON, with coarse-grained logs defined by the specific system. One log file contains multiple nodes' execution logs.
    \item PGo-instrumented logs generated by TraceLink \citep{hackett2025tracelink}: Also newline-delimited JSON and contains PGo-specific concepts like archetype names and vector clocks. Variable updates are logged in fine-grained detail at each PGo-defined critical section. One log file is output per node; there are multiple log files per distributed execution.
\end{packed_itemize}

\textbf{Optimizations.} We anecdotally noticed that passing single execution traces results in overfitting by the model, with generated models closely reflecting the single executed path. Providing more traces improves context for the model.

One issue we encountered was fitting large traces into models' context windows. The JSON structure of traces is expensive in tokens, because each ``\texttt{[}'', ``\texttt{:}'', and other punctuation represents a separate token. Most of the models we used had a context window of about 200K tokens; a JSON trace of several megabytes, such as the Etcd Raft traces, simply could not fit. We solved this with three workarounds:
\begin{packed_itemize}
\item We support sampling for systems with large traces, randomly choosing a set of execution traces among all collected traces.
\item We convert the nested JSON structure into tab-separated values (TSV) format, which deduplicates JSON keys into the TSV header and uses only tabs as a separator to save tokens.
\item We abbreviate repeated state or action values (e.g., \texttt{ReceiveRequestVoteResponse}) to acronyms (e.g., \texttt{RRVR}) and provide a mapping in the prompt.
\end{packed_itemize}

The TSV and abbreviation optimizations significantly save tokens: with the Claude tokenizer, it reduces token use by 62\% for ten lines of Etcd Raft traces (from 645 to 262 tokens), and by 63\% for ten lines of mutex traces (from 866 to 318 tokens). This enabled us to fit multiple traces into the initial prompt, reducing the impact of overfitting. We did not apply this optimization to the other methods, since code is less structured and does not have obvious candidates for deduplication.

\section{Complete Evaluation Results}

\begin{plsreview}
\subsection{Liveness Violation Analysis}
\label{appendix:liveness-analysis}

We analyzed counterexamples from two representative systems (Asterinas SpinLock and Etcd Raft) 
    and categorized liveness violations into two main classes:
\begin{packed_itemize}
\item \emph{Fairness-related issues} that prevent progress due to missing fairness declarations, 
    overly narrow or overly broad constraints (e.g., defined as WF(Next));
\item \emph{Logical/structural issues} that block progress due to conflicting updates 
    or missing/incorrect logic in action definitions.
\end{packed_itemize}
Since the modeling task focuses on state/action models of the system implementation 
    and LLMs are not required to generate temporal operators (e.g., in liveness properties),
    our categorization does not include errors related to temporal operators.

Table~\ref{tab:liveness-breakdown} presents the detailed breakdown of violations by category and LLM.
For Asterinas SpinLock, fairness-related issues dominate the violations, 
    particularly ``too broad'' and ``too narrow'' constraints.
For instance, Claude-Sonnet-4, generated 26 out of 32 violations due to overly broad fairness assumptions.

For Etcd Raft, liveness violations are primarily caused by logical/structural issues.
The model's large state space causes these logical errors to block progress 
    before fairness-related issues can manifest.
Nevertheless, manual inspection confirms that fairness conditions are generally incorrect.

\begin{table}[h]
\plsreviewtable{}
\centering
\small
\caption{\plsreviewinline{Liveness violations by category in Asterinas SpinLock and Etcd Raft for the basic modeling agent.}}
\label{tab:liveness-breakdown}

\begin{subtable}{\textwidth}
\centering
\caption{\plsreviewinline{Asterinas SpinLock liveness violations by category}}
\label{tab:liveness-spinlock}
\begin{tabular}{lccccc}
\toprule
\textbf{LLM} & \textbf{Fairness} & \textbf{Fairness} & \textbf{Missing} & \textbf{Logical} & \textbf{Total} \\
 & \textbf{too broad} & \textbf{too narrow} & \textbf{fairness} & \textbf{errors} & \textbf{violations} \\
\midrule
Claude-Sonnet-4 & 26 & 2 & 4 & 0 & 32 \\
GPT-5 & 8 & 10 & 2 & 0 & 20 \\
Gemini-2.5-Pro & 4 & 6 & 0 & 0 & 10 \\
DeepSeek-R1 & 4 & 2 & 0 & 2 & 8 \\
\bottomrule
\end{tabular}
\end{subtable}

\vspace{10pt}

\begin{subtable}{\textwidth}
\centering
\caption{\plsreviewinline{Etcd Raft liveness violations by category}}
\label{tab:liveness-raft}
\begin{tabular}{lccc}
\toprule
\textbf{LLM} & \textbf{Logical missing/errors} & \textbf{Conflicting updates} & \textbf{Total violations} \\
\midrule
Claude-Sonnet-4 & 4 & 8 & 12 \\
GPT-5 & 2 & 8 & 20 \\
Gemini-2.5-Pro & 0 & 0 & 0 \\
DeepSeek-R1 & 4 & 4 & 8 \\
\bottomrule
\end{tabular}
\end{subtable}

\end{table}

\end{plsreview}

\begin{plsreview}
\subsection{Detailed Results by System}
\label{appendix:results}
\end{plsreview}

We present the complete evaluation results for all systems in our benchmark 
    using three AI agents: Basic Modeling, Code Translation, and Trace Learning
    in Tables~\ref{tab:spinlock-result-2}--\ref{tab:zookeeper-result}.
These tables follow the same evaluation setup as described in \S\ref{sec:setup}. %

\begin{table}[htbp]
    \centering
    \footnotesize
    \vspace{-4pt}
    \caption{Asterinas Spinlock}
    \label{tab:spinlock-result-2}
    \footnotesize
    \begin{tabular}{llcccc}
        \toprule
        \bf Agent & \bf LLM & \multicolumn{1}{l}{\bf Syntax} & \multicolumn{1}{l}{\bf Runtime} & \multicolumn{1}{l}{\bf Conformance} & \multicolumn{1}{l}{\bf Invariant} \\ \midrule
\multirow{4}{*}{\begin{tabular}[c]{@{}l@{}}Basic Modeling\end{tabular}} & Claude-Sonnet-4 & \excellentcell{}100.00\%\checkmarksymbol{} & \excellentcell{}100.00\%\checkmarksymbol{} & \excellentcell{}100.00\% & \excellentcell{}100.00\% \\
         & GPT-5 & \excellentcell{}100.00\%\checkmarksymbol{} & \excellentcell{}100.00\%\checkmarksymbol{} & \excellentcell{}80.00\% & \excellentcell{}100.00\% \\
         & Gemini-2.5-Pro & \excellentcell{}100.00\%\checkmarksymbol{} & \excellentcell{}100.00\%\checkmarksymbol{} & \excellentcell{}80.00\% & \excellentcell{}85.71\% \\
         & DeepSeek-R1 & \excellentcell{}100.00\%\checkmarksymbol{} & \excellentcell{}100.00\%\checkmarksymbol{} & \excellentcell{}80.00\% & \excellentcell{}100.00\% \\ \midrule
\multirow{4}{*}{Code Translation} & Claude-Sonnet-4 & \excellentcell{}100.00\%\checkmarksymbol{} & \excellentcell{}100.00\%\checkmarksymbol{} & \excellentcell{}100.00\% & \excellentcell{}100.00\% \\
         & GPT-5 & \excellentcell{}100.00\%\checkmarksymbol{} & \excellentcell{}100.00\%\checkmarksymbol{} & \excellentcell{}100.00\% & \excellentcell{}85.71\% \\
         & Gemini-2.5-Pro & \excellentcell{}100.00\%\checkmarksymbol{} & \excellentcell{}100.00\%\checkmarksymbol{} & \excellentcell{}100.00\% & \excellentcell{}100.00\% \\
         & DeepSeek-R1 & \excellentcell{}100.00\%\checkmarksymbol{} & \excellentcell{}100.00\%\checkmarksymbol{} & \excellentcell{}100.00\% & \excellentcell{}100.00\% \\ \midrule
\multirow{4}{*}{Trace Learning} & Claude-Sonnet-4 & \goodcell{}50.00\%\wrongsymbol{} & - & - & - \\
         & GPT-5 & \excellentcell{}100.00\%\checkmarksymbol{} & 0.00\%\wrongsymbol{} & - & - \\
         & Gemini-2.5-Pro & \excellentcell{}100.00\%\checkmarksymbol{} & 0.00\%\wrongsymbol{} & - & - \\
         & DeepSeek-R1 & \excellentcell{}100.00\%\checkmarksymbol{} & 0.00\%\wrongsymbol{} & - & - \\ \bottomrule
    \end{tabular}
    \begin{flushleft}
        {\vspace{5pt}
        \footnotesize
    The trace learning agent underperforms compared to the other two agents, 
        typically failing compilation and runtime checks. 
    We observe that it is more difficult for LLMs to process structured trace data,
        in comparison to source code. 
        Specifically, Claude-Sonnet-4 appears to be particularly weak in this regard, achieving the lowest syntax scores,
        despite its coding capabilities.
        This trend of the trace learning agent is consistent across all the evaluated system artifacts (Tables~\ref{tab:etcd-result-2}--\ref{tab:zookeeper-result}).}
    \end{flushleft}
\end{table}

\begin{table}[htbp]
    \centering
    \vspace{-4pt}
    \caption{Etcd Raft}
    \label{tab:etcd-result-2}
    \footnotesize
    \begin{tabular}{llcccc}
        \toprule
        \bf Agent & \bf LLM & \multicolumn{1}{l}{\bf Syntax} & \multicolumn{1}{l}{\bf Runtime} & \multicolumn{1}{l}{\bf Conformance} & \multicolumn{1}{l}{\bf Invariant} \\ \midrule
\multirow{4}{*}{\begin{tabular}[c]{@{}l@{}}Basic Modeling\end{tabular}} & Claude-Sonnet-4 & \excellentcell{}100.00\%\checkmarksymbol{} & \faircell{}25.00\%\checkmarksymbol{} & \faircell{}7.69\% & \goodcell{}69.23\% \\
         & GPT-5 & \faircell{}47.87\%\wrongsymbol{} & - & - & - \\
         & Gemini-2.5-Pro & \goodcell{}50.00\%\wrongsymbol{} & - & - & - \\
         & DeepSeek-R1 & \goodcell{}50.00\%\wrongsymbol{} & - & - & - \\ \midrule
\multirow{4}{*}{Code Translation} & Claude-Sonnet-4 & \excellentcell{}100.00\%\checkmarksymbol{} & \goodcell{}66.67\%\checkmarksymbol{} & \faircell{}15.38\% & \excellentcell{}92.31\% \\
         & GPT-5 & \excellentcell{}100.00\%\checkmarksymbol{} & \faircell{}20.00\%\wrongsymbol{} & - & - \\
         & Gemini-2.5-Pro & \faircell{}44.44\%\wrongsymbol{} & - & - & - \\
         & DeepSeek-R1 & \excellentcell{}100.00\%\checkmarksymbol{} & 0.00\%\wrongsymbol{} & - & - \\ \midrule
\multirow{4}{*}{Trace Learning} & Claude-Sonnet-4 & \goodcell{}50.00\%\wrongsymbol{} & - & - & - \\
         & GPT-5 & \faircell{}48.78\%\wrongsymbol{} & - & - & - \\
         & Gemini-2.5-Pro & \faircell{}42.31\%\wrongsymbol{} & - & - & - \\
         & DeepSeek-R1 & \faircell{}47.73\%\wrongsymbol{} & - & - & - \\ \bottomrule
    \end{tabular}
\end{table}

\begin{table}[htbp]
    \centering
    \caption{Asterinas Mutex}
    \label{tab:mutex-result}
    \footnotesize
    \begin{tabular}{llcccc}
        \toprule
        \bf Agent & \bf LLM & \multicolumn{1}{l}{\bf Syntax} & \multicolumn{1}{l}{\bf Runtime} & \multicolumn{1}{l}{\bf Conformance} & \multicolumn{1}{l}{\bf Invariant} \\ \midrule
\multirow{4}{*}{\begin{tabular}[c]{@{}l@{}}Basic Modeling\end{tabular}} & Claude-Sonnet-4 & \excellentcell{}100.00\%\checkmarksymbol{} & \excellentcell{}100.00\%\checkmarksymbol{} & \excellentcell{}100.00\% & \excellentcell{}100.00\% \\
         & GPT-5 & \excellentcell{}100.00\%\checkmarksymbol{} & \excellentcell{}100.00\%\checkmarksymbol{} & \excellentcell{}100.00\% & \excellentcell{}85.71\% \\
         & Gemini-2.5-Pro & \excellentcell{}100.00\%\checkmarksymbol{} & \excellentcell{}100.00\%\checkmarksymbol{} & \goodcell{}66.67\% & \excellentcell{}85.71\% \\
         & DeepSeek-R1 & \excellentcell{}100.00\%\checkmarksymbol{} & \excellentcell{}100.00\%\checkmarksymbol{} & \goodcell{}66.67\% & \excellentcell{}100.00\% \\ \midrule
\multirow{4}{*}{Code Translation} & Claude-Sonnet-4 & \excellentcell{}100.00\%\checkmarksymbol{} & \excellentcell{}100.00\%\checkmarksymbol{} & \excellentcell{}100.00\% & \excellentcell{}100.00\% \\
         & GPT-5 & \excellentcell{}100.00\%\checkmarksymbol{} & \excellentcell{}100.00\%\checkmarksymbol{} & \excellentcell{}100.00\% & \excellentcell{}100.00\% \\
         & Gemini-2.5-Pro & \excellentcell{}100.00\%\checkmarksymbol{} & \excellentcell{}100.00\%\checkmarksymbol{} & \excellentcell{}100.00\% & \excellentcell{}85.71\% \\
         & DeepSeek-R1 & \excellentcell{}100.00\%\checkmarksymbol{} & \excellentcell{}100.00\%\checkmarksymbol{} & \excellentcell{}100.00\% & \excellentcell{}85.71\% \\ \midrule
\multirow{4}{*}{Trace Learning} & Claude-Sonnet-4 & \goodcell{}50.00\%\wrongsymbol{} & - & - & - \\
         & GPT-5 & \goodcell{}50.00\%\wrongsymbol{} & - & - & - \\
         & Gemini-2.5-Pro & \excellentcell{}100.00\%\checkmarksymbol{} & 0.00\%\wrongsymbol{} & - & - \\
         & DeepSeek-R1 & \goodcell{}50.00\%\wrongsymbol{} & 0.00\%\wrongsymbol{} & - & - \\ \bottomrule
    \end{tabular}
\end{table}

\begin{table}[htbp]
    \centering
    \caption{Asterinas Rwmutex}
    \label{tab:rwmutex-result}
    \footnotesize
    \begin{tabular}{llcccc}
        \toprule
        \bf Agent & \bf LLM & \multicolumn{1}{l}{\bf Syntax} & \multicolumn{1}{l}{\bf Runtime} & \multicolumn{1}{l}{\bf Conformance} & \multicolumn{1}{l}{\bf Invariant} \\ \midrule
\multirow{4}{*}{\begin{tabular}[c]{@{}l@{}}Basic Modeling\end{tabular}} & Claude-Sonnet-4 & \excellentcell{}100.00\%\checkmarksymbol{} & \excellentcell{}100.00\%\checkmarksymbol{} & \excellentcell{}100.00\% & \excellentcell{}90.00\% \\
         & GPT-5 & \excellentcell{}100.00\%\checkmarksymbol{} & \excellentcell{}100.00\%\checkmarksymbol{} & \goodcell{}75.00\% & \excellentcell{}80.00\% \\
         & Gemini-2.5-Pro & \excellentcell{}100.00\%\checkmarksymbol{} & \excellentcell{}100.00\%\checkmarksymbol{} & 0.00\% & \excellentcell{}80.00\% \\
         & DeepSeek-R1 & \excellentcell{}100.00\%\checkmarksymbol{} & \excellentcell{}100.00\%\checkmarksymbol{} & \goodcell{}50.00\% & \excellentcell{}90.00\% \\ \midrule
\multirow{4}{*}{Code Translation} & Claude-Sonnet-4 & \excellentcell{}100.00\%\checkmarksymbol{} & \excellentcell{}100.00\%\checkmarksymbol{} & \excellentcell{}100.00\% & \excellentcell{}90.00\% \\
         & GPT-5 & \excellentcell{}100.00\%\checkmarksymbol{} & \excellentcell{}100.00\%\checkmarksymbol{} & \excellentcell{}100.00\% & \excellentcell{}90.00\% \\
         & Gemini-2.5-Pro & \excellentcell{}100.00\%\checkmarksymbol{} & \excellentcell{}100.00\%\checkmarksymbol{} & \excellentcell{}100.00\% & \excellentcell{}80.00\% \\
         & DeepSeek-R1 & \excellentcell{}100.00\%\checkmarksymbol{} & \excellentcell{}100.00\%\checkmarksymbol{} & \goodcell{}50.00\% & \excellentcell{}90.00\% \\ \midrule
\multirow{4}{*}{Trace Learning} & Claude-Sonnet-4 & \goodcell{}50.00\%\wrongsymbol{} & - & - & - \\
         & GPT-5 & \excellentcell{}100.00\%\checkmarksymbol{} & 0.00\%\wrongsymbol{} & - & - \\
         & Gemini-2.5-Pro & \excellentcell{}100.00\%\checkmarksymbol{} & 0.00\%\wrongsymbol{} & - & - \\
         & DeepSeek-R1 & \goodcell{}50.00\%\wrongsymbol{} & - & - & - \\ \bottomrule
    \end{tabular}
\end{table}

\begin{table}[htbp]
    \centering
    \caption{\plsreviewinline{Asterinas Ringbuffer}}
    \label{tab:ringbuffer-result}
    \footnotesize
    \plsreviewtable{}
    \begin{tabular}{llcccc}
        \toprule
        \bf Agent & \bf LLM & \multicolumn{1}{l}{\bf Syntax} & \multicolumn{1}{l}{\bf Runtime} & \multicolumn{1}{l}{\bf Conformance} & \multicolumn{1}{l}{\bf Invariant} \\ \midrule
\multirow{4}{*}{\begin{tabular}[c]{@{}l@{}}Basic Modeling\end{tabular}} & Claude-Sonnet-4 & \excellentcell{}100.00\%\checkmarksymbol{} & \excellentcell{}100.00\%\checkmarksymbol{} & \excellentcell{}100.00\% & \excellentcell{}100.00\% \\
         & GPT-5 & \excellentcell{}100.00\%\checkmarksymbol{} & 0.00\%\wrongsymbol{} & - & - \\
         & Gemini-2.5-Pro & \excellentcell{}100.00\%\checkmarksymbol{} & 0.00\%\wrongsymbol{} & - & - \\
         & DeepSeek-R1 & \excellentcell{}100.00\%\checkmarksymbol{} & 0.00\%\wrongsymbol{} & - & - \\ \midrule
\multirow{4}{*}{Code Translation} & Claude-Sonnet-4 & \excellentcell{}100.00\%\checkmarksymbol{} & \excellentcell{}100.00\%\checkmarksymbol{} & \excellentcell{}100.00\% & \excellentcell{}100.00\% \\
         & GPT-5 & \excellentcell{}100.00\%\checkmarksymbol{} & \excellentcell{}100.00\%\checkmarksymbol{} & \excellentcell{}100.00\% & \goodcell{}75.00\% \\
         & Gemini-2.5-Pro & \excellentcell{}100.00\%\checkmarksymbol{} & \excellentcell{}100.00\%\checkmarksymbol{} & \excellentcell{}100.00\% & \excellentcell{}100.00\% \\
         & DeepSeek-R1 & \excellentcell{}100.00\%\checkmarksymbol{} & 0.00\%\wrongsymbol{} & - & - \\ \midrule
\multirow{4}{*}{Trace Learning} & Claude-Sonnet-4 & \excellentcell{}100.00\%\checkmarksymbol{} & 0.00\%\wrongsymbol{} & - & - \\
         & GPT-5 & \excellentcell{}100.00\%\checkmarksymbol{} & 0.00\%\wrongsymbol{} & - & - \\
         & Gemini-2.5-Pro & \excellentcell{}100.00\%\checkmarksymbol{} & 0.00\%\wrongsymbol{} & - & - \\
         & DeepSeek-R1 & \excellentcell{}100.00\%\checkmarksymbol{} & 0.00\%\wrongsymbol{} & - & - \\ \bottomrule
    \end{tabular}
\end{table}

\begin{table}[htbp]
    \centering
    \caption{Redis Raft}
    \label{tab:redisraft-result}
    \footnotesize
    \begin{tabular}{llcccc}
        \toprule
        \bf Agent & \bf LLM & \multicolumn{1}{l}{\bf Syntax} & \multicolumn{1}{l}{\bf Runtime} & \multicolumn{1}{l}{\bf Conformance} & \multicolumn{1}{l}{\bf Invariant} \\ \midrule
\multirow{4}{*}{\begin{tabular}[c]{@{}l@{}}Basic Modeling\end{tabular}} & Claude-Sonnet-4 & \excellentcell{}100.00\%\checkmarksymbol{} & 0.00\%\wrongsymbol{} & - & - \\
         & GPT-5 & \excellentcell{}100.00\%\checkmarksymbol{} & 0.00\%\wrongsymbol{} & - & - \\
         & Gemini-2.5-Pro & \goodcell{}50.00\%\wrongsymbol{} & - & - & - \\
         & DeepSeek-R1 & \excellentcell{}100.00\%\checkmarksymbol{} & 0.00\%\wrongsymbol{} & - & - \\ \midrule
\multirow{4}{*}{Code Translation} & Claude-Sonnet-4 & \excellentcell{}100.00\%\checkmarksymbol{} & \faircell{}23.81\%\checkmarksymbol{} & \faircell{}9.09\% & \goodcell{}75.00\% \\
         & GPT-5 & \excellentcell{}100.00\%\checkmarksymbol{} & 0.00\%\wrongsymbol{} & - & - \\
         & Gemini-2.5-Pro & \goodcell{}50.00\%\wrongsymbol{} & - & - & - \\
         & DeepSeek-R1 & \excellentcell{}100.00\%\checkmarksymbol{} & \excellentcell{}100.00\%\checkmarksymbol{} & 0.00\% & \faircell{}25.00\% \\ \midrule
         \multirow{4}{*}{Trace Learning} & Claude-Sonnet-4 & \goodcell{}50.00\%\wrongsymbol{} & - & - & - \\
                  & GPT-5 & \faircell{}47.06\%\wrongsymbol{} & - & - & - \\
                  & Gemini-2.5-Pro & \excellentcell{}100.00\%\checkmarksymbol{} & 0.00\%\wrongsymbol{} & - & - \\
                  & DeepSeek-R1 & \faircell{}48.53\%\wrongsymbol{} & - & - & - \\ \bottomrule
    \end{tabular}
\begin{flushleft}
    {\vspace{5pt}
    \footnotesize
    The system model generated by DeepSeek-R1 is overly simplified; thus, it has high syntax and runtime correctness,
        but have low score on invariants (the protocol logic is incorrect) and 0\% on conformance.}
\end{flushleft}
\end{table}

\begin{table}[htbp]
    \centering
    \caption{Xline CURP}
    \label{tab:curp-result}
    \footnotesize
    \begin{tabular}{llcccc}
        \toprule
        \bf Agent & \bf LLM & \multicolumn{1}{l}{\bf Syntax} & \multicolumn{1}{l}{\bf Runtime} & \multicolumn{1}{l}{\bf Conformance} & \multicolumn{1}{l}{\bf Invariant} \\ \midrule
\multirow{4}{*}{\begin{tabular}[c]{@{}l@{}}Basic Modeling\end{tabular}} & Claude-Sonnet-4 & \excellentcell{}100.00\%\checkmarksymbol{} & 0.00\%\wrongsymbol{} & - & - \\
         & GPT-5 & \excellentcell{}100.00\%\checkmarksymbol{} & 0.00\%\wrongsymbol{} & - & - \\
         & Gemini-2.5-Pro & \excellentcell{}100.00\%\checkmarksymbol{} & 0.00\%\wrongsymbol{} & - & - \\
         & DeepSeek-R1 & \excellentcell{}100.00\%\checkmarksymbol{} & 0.00\%\wrongsymbol{} & - & - \\ \midrule
\multirow{4}{*}{Code Translation} & Claude-Sonnet-4 & \excellentcell{}100.00\%\checkmarksymbol{} & \excellentcell{}100.00\%\checkmarksymbol{} & \goodcell{}50.00\% & \excellentcell{}100.00\% \\
         & GPT-5 & \excellentcell{}100.00\%\checkmarksymbol{} & 0.00\%\wrongsymbol{} & - & - \\
         & Gemini-2.5-Pro & \excellentcell{}100.00\%\checkmarksymbol{} & \excellentcell{}100.00\%\checkmarksymbol{} & \goodcell{}66.67\% & \excellentcell{}100.00\% \\
         & DeepSeek-R1 & \excellentcell{}100.00\%\checkmarksymbol{} & 0.00\%\wrongsymbol{} & - & - \\ \midrule
\multirow{4}{*}{Trace Learning} & Claude-Sonnet-4 & \goodcell{}50.00\%\wrongsymbol{} & - & - & - \\
         & GPT-5 & \excellentcell{}100.00\%\checkmarksymbol{} & 0.00\%\wrongsymbol{} & - & - \\
         & Gemini-2.5-Pro & \faircell{}46.15\%\wrongsymbol{} & - & - & - \\
         & DeepSeek-R1 & \excellentcell{}100.00\%\checkmarksymbol{} & 0.00\%\wrongsymbol{} & - & - \\ \bottomrule
    \end{tabular}
\begin{flushleft}
    {\vspace{5pt}
    \footnotesize
    Xline CURP is one of the largest system artifacts in \bench{} (see Table~\ref{tab:systems}).
    We suspect that the system model generated by Gemini-2.5-Pro benefits from its 1M-token context window, 
        enabling effective summarization of the 4000+ line codebase into a concise \tla{} representation.}
\end{flushleft}
\end{table}

\begin{table}[htbp]
    {\centering
    \caption{PGo dqueue}
    \label{tab:dqueue-result}
    \footnotesize
    \begin{tabular}{llcccc}
        \toprule
        \bf Agent & \bf LLM & \multicolumn{1}{l}{\bf Syntax} & \multicolumn{1}{l}{\bf Runtime} & \multicolumn{1}{l}{\bf Conformance} & \multicolumn{1}{l}{\bf Invariant} \\ \midrule
\multirow{4}{*}{\begin{tabular}[c]{@{}l@{}}Basic Modeling\end{tabular}} & Claude-Sonnet-4 & \excellentcell{}100.00\%\checkmarksymbol{} & \faircell{}33.33\%\checkmarksymbol{} & \faircell{}33.33\% & 0.00\% \\
         & GPT-5 & \excellentcell{}100.00\%\checkmarksymbol{} & 0.00\%\wrongsymbol{} & - & - \\
         & Gemini-2.5-Pro & \excellentcell{}100.00\%\checkmarksymbol{} & 0.00\%\wrongsymbol{} & - & - \\
         & DeepSeek-R1 & \excellentcell{}100.00\%\checkmarksymbol{} & 0.00\%\wrongsymbol{} & - & - \\ \midrule
\multirow{4}{*}{Code Translation} & Claude-Sonnet-4 & \excellentcell{}100.00\%\checkmarksymbol{} & \excellentcell{}100.00\%\checkmarksymbol{} & 0.00\% & \excellentcell{}100.00\% \\
         & GPT-5 & \excellentcell{}100.00\%\checkmarksymbol{} & \excellentcell{}100.00\%\checkmarksymbol{} & 0.00\% & \excellentcell{}100.00\% \\
         & Gemini-2.5-Pro & \excellentcell{}100.00\%\checkmarksymbol{} & \excellentcell{}100.00\%\checkmarksymbol{} & 0.00\% & \excellentcell{}100.00\% \\
         & DeepSeek-R1 & \excellentcell{}100.00\%\checkmarksymbol{} & \excellentcell{}100.00\%\checkmarksymbol{} & 0.00\% & \excellentcell{}100.00\% \\ \midrule
\multirow{4}{*}{Trace Learning} & Claude-Sonnet-4 & \excellentcell{}100.00\%\checkmarksymbol{} & 0.00\%\wrongsymbol{} & - & - \\
         & GPT-5 & \excellentcell{}100.00\%\checkmarksymbol{} & 0.00\%\wrongsymbol{} & - & - \\
         & Gemini-2.5-Pro & \excellentcell{}100.00\%\checkmarksymbol{} & 0.00\%\wrongsymbol{} & - & - \\
         & DeepSeek-R1 & \excellentcell{}100.00\%\checkmarksymbol{} & 0.00\%\wrongsymbol{} & - & - \\ \bottomrule
    \end{tabular}}
\end{table}

\begin{table}[htbp]
    \centering
    \caption{PGo locksvc}
    \label{tab:locksvc-result}
    \footnotesize
    \begin{tabular}{llcccc}
        \toprule
        \bf Agent & \bf LLM & \multicolumn{1}{l}{\bf Syntax} & \multicolumn{1}{l}{\bf Runtime} & \multicolumn{1}{l}{\bf Conformance} & \multicolumn{1}{l}{\bf Invariant} \\ \midrule
\multirow{4}{*}{\begin{tabular}[c]{@{}l@{}}Basic Modeling\end{tabular}} & Claude-Sonnet-4 & \faircell{}44.45\%\wrongsymbol{} & - & - & - \\
         & GPT-5 & \excellentcell{}100.00\%\checkmarksymbol{} & 0.00\%\wrongsymbol{} & - & - \\
         & Gemini-2.5-Pro & \excellentcell{}100.00\%\checkmarksymbol{} & 0.00\%\wrongsymbol{} & - & - \\
         & DeepSeek-R1 & \excellentcell{}100.00\%\checkmarksymbol{} & 0.00\%\wrongsymbol{} & - & - \\ \midrule
\multirow{4}{*}{Code Translation} & Claude-Sonnet-4 & \excellentcell{}100.00\%\checkmarksymbol{} & \excellentcell{}100.00\%\checkmarksymbol{} & 0.00\% & \excellentcell{}83.33\% \\
         & GPT-5 & \excellentcell{}100.00\%\checkmarksymbol{} & \excellentcell{}100.00\%\checkmarksymbol{} & 0.00\% & \goodcell{}66.67\% \\
         & Gemini-2.5-Pro & \excellentcell{}100.00\%\checkmarksymbol{} & 0.00\%\wrongsymbol{} & - & - \\
         & DeepSeek-R1 & \excellentcell{}100.00\%\checkmarksymbol{} & \excellentcell{}100.00\%\checkmarksymbol{} & 0.00\% & \goodcell{}50.00\% \\ \midrule
\multirow{4}{*}{Trace Learning} & Claude-Sonnet-4 & \faircell{}42.31\%\wrongsymbol{} & - & - & - \\
         & GPT-5 & \excellentcell{}100.00\%\checkmarksymbol{} & 0.00\%\wrongsymbol{} & - & - \\
         & Gemini-2.5-Pro & \goodcell{}50.00\%\wrongsymbol{} & - & - & - \\
         & DeepSeek-R1 & \goodcell{}50.00\%\wrongsymbol{} & - & - & - \\ \bottomrule
    \end{tabular}
\end{table}

\begin{table}[htbp] %
    \centering
    \caption{PGo raftkvs}
    \label{tab:raftkvs-result}
    \footnotesize
    \begin{tabular}{llcccc}
        \toprule
        \bf Agent & \bf LLM & \multicolumn{1}{l}{\bf Syntax} & \multicolumn{1}{l}{\bf Runtime} & \multicolumn{1}{l}{\bf Conformance} & \multicolumn{1}{l}{\bf Invariant} \\ \midrule
\multirow{4}{*}{\begin{tabular}[c]{@{}l@{}}Basic Modeling\end{tabular}} & Claude-Sonnet-4 & \faircell{}44.57\%\wrongsymbol{} & - & - & - \\
         & GPT-5 & \excellentcell{}100.00\%\checkmarksymbol{} & 0.00\%\wrongsymbol{} & - & - \\
         & Gemini-2.5-Pro & \faircell{}45.84\%\wrongsymbol{} & - & - & - \\
         & DeepSeek-R1 & \faircell{}45.32\%\wrongsymbol{} & - & - & - \\ \midrule
\multirow{4}{*}{Code Translation} & Claude-Sonnet-4 & \excellentcell{}100.00\%\checkmarksymbol{} & \goodcell{}50.00\%\checkmarksymbol{} & 0.00\% & \excellentcell{}90.91\% \\
         & GPT-5 & \excellentcell{}100.00\%\checkmarksymbol{} & \excellentcell{}100.00\%\checkmarksymbol{} & 0.00\% & \goodcell{}72.73\% \\
         & Gemini-2.5-Pro & \faircell{}40.91\%\wrongsymbol{} & - & - & - \\
         & DeepSeek-R1 & \excellentcell{}100.00\%\checkmarksymbol{} & \faircell{}22.22\%\wrongsymbol{} & - & - \\ \midrule
\multirow{4}{*}{Trace Learning} & Claude-Sonnet-4 & \goodcell{}50.00\%\wrongsymbol{} & - & - & - \\
         & GPT-5 & \faircell{}46.55\%\wrongsymbol{} & - & - & - \\
         & Gemini-2.5-Pro & \faircell{}41.67\%\wrongsymbol{} & - & - & - \\
         & DeepSeek-R1 & \faircell{}47.83\%\wrongsymbol{} & - & - & - \\ \bottomrule
    \end{tabular}
\begin{flushleft}
        {\vspace{5pt}
    \footnotesize
    We observe that the characteristics of LLM performance on PGo-compiled systems are very different 
        from human-written systems as discussed in Section~\ref{sec:eval} and Appendix~\ref{sec:pgo}.
    We find that GPT-5 performs generally perform well on PGo systems, indicating its ability
        of understanding machine-generated code patterns.}
\end{flushleft}
\end{table}

\begin{table}[htbp]
    \centering
    \caption{\plsreviewinline{ZooKeeper Fast Leader Election (FLE)}}
    \label{tab:zookeeper-result}
    \footnotesize
    \plsreviewtable{}
    \begin{tabular}{llcccc}
        \toprule
        \bf Agent & \bf LLM & \multicolumn{1}{l}{\bf Syntax} & \multicolumn{1}{l}{\bf Runtime} & \multicolumn{1}{l}{\bf Conformance} & \multicolumn{1}{l}{\bf Invariant} \\ \midrule
\multirow{4}{*}{\begin{tabular}[c]{@{}l@{}}Basic Modeling\end{tabular}} & Claude-Sonnet-4 & \excellentcell{}100.00\%\checkmarksymbol{} & 0.00\%\wrongsymbol{} & - & - \\
         & GPT-5 & \excellentcell{}100.00\%\checkmarksymbol{} & 0.00\%\wrongsymbol{} & - & - \\
         & Gemini-2.5-Pro & \excellentcell{}100.00\%\checkmarksymbol{} & 0.00\%\wrongsymbol{} & - & - \\
         & DeepSeek-R1 & \excellentcell{}100.00\%\checkmarksymbol{} & 0.00\%\wrongsymbol{} & - & - \\ \midrule
\multirow{4}{*}{Code Translation} & Claude-Sonnet-4 & \excellentcell{}100.00\%\checkmarksymbol{} & 0.00\%\wrongsymbol{} & - & - \\
         & GPT-5 & \excellentcell{}100.00\%\checkmarksymbol{} & 0.00\%\wrongsymbol{} & - & - \\
         & Gemini-2.5-Pro & \excellentcell{}100.00\%\checkmarksymbol{} & 0.00\%\wrongsymbol{} & - & - \\
         & DeepSeek-R1 & \excellentcell{}100.00\%\checkmarksymbol{} & 0.00\%\wrongsymbol{} & - & - \\ \midrule
\multirow{4}{*}{Trace Learning} & Claude-Sonnet-4 & \faircell{}44.44\%\wrongsymbol{} & - & - & - \\
         & GPT-5 & \faircell{}47.92\%\wrongsymbol{} & - & - & - \\
         & Gemini-2.5-Pro & \excellentcell{}100.00\%\checkmarksymbol{} & 0.00\%\wrongsymbol{} & - & - \\
         & DeepSeek-R1 & \excellentcell{}100.00\%\checkmarksymbol{} & 0.00\%\wrongsymbol{} & - & - \\ \bottomrule
    \end{tabular}
\begin{flushleft}
    {\vspace{5pt}
    \footnotesize ZooKeeper FLE has the largest codebase and implements the complex ZAB protocol, 
        making it the most challenging system to model among all \numberofsystems{} artifacts.}
\end{flushleft}
\end{table}

\end{document}